\documentclass[10pt]{article} 
\usepackage[accepted]{tmlr}

\usepackage{hyperref}
\usepackage{url}

\usepackage{makecell}
\usepackage{adjustbox}
\usepackage{threeparttable}
\usepackage{array}
\usepackage{tabularx}

\usepackage[section]{placeins}

\newcolumntype{Y}{>{\raggedright\arraybackslash}X}
\newcolumntype{C}[1]{>{\centering\arraybackslash}p{#1}}
\newcolumntype{L}[1]{>{\raggedright\arraybackslash}p{#1}}

\title{Localized Operator Learning with Adaptive Partition-of-Unity Mixture-of-Expert Networks}

\author{
    \name \hspace{-0.15cm}Madison Cooley
    \email mcooley@sci.utah.edu \\
    \addr Scientific Computing Institute\\
    University of Utah \\
    Salt Lake City, UT
    \AND
    \name Ramansh Sharma
    \email ramansh@cs.utah.edu\\
    \addr Kahlert School of Computing\\
    University of Utah\\
    Salt Lake City, UT
    \AND
    \name Shandian Zhe
    \email shandian.zhe@utah.edu \\
    \addr Kahlert School of Computing\\
    University of Utah \\
    Salt Lake City, UT
    \AND
    \name Robert M. Kirby
    \email mike.kirby@utah.edu \\
    \addr Scientific Computing Institute\\
    University of Utah \\
    Salt Lake City, UT
    \AND
    \name Varun Shankar
    \email shankar@cs.utah.edu \\
    \addr Kahlert School of Computing\\
    University of Utah \\
    Salt Lake City, UT
}

\def\month{10}  
\def\year{2026} 
\def\openreview{\url{https://openreview.net/forum?id=ccEY1fZ5gA}} 

\newcommand{\done}[1]{\textcolor{black}{#1}}

\definecolor{darkgreen}{RGB}{0,100,0}

\newcommand{\roneedits}[1]{\textcolor{black}{#1}}
\newcommand{\rtwoedits}[1]{\textcolor{black}{#1}}
\newcommand{\rthreeedits}[1]{\textcolor{black}{#1}}
\newcommand{\generaledits}[1]{\textcolor{black}{#1}}      

\newcommand{\METHODNAME}{HiRefPOU}
\usepackage{stmaryrd} 
\usepackage{xcolor}     
\usepackage{url}
\usepackage{booktabs}       
\usepackage{wrapfig}     
\usepackage{amsmath}
\usepackage{subcaption}
\usepackage{graphicx}    
\usepackage{algorithm2e}
\usepackage{siunitx}
\usepackage{pgfplots}
\pgfplotsset{compat=1.18}
\usepackage{tikz}
\usetikzlibrary{3d}
\usetikzlibrary{positioning}
\usetikzlibrary{fit}
\usetikzlibrary{shapes.geometric, arrows.meta, calc}
\usetikzlibrary{positioning,3d,arrows.meta}

\usepackage[utf8]{inputenc} 
\usepackage[T1]{fontenc}    
\usepackage{amsfonts}       
\usepackage{nicefrac}       
\usepackage{microtype}      
\usepackage{lipsum}		    
\usepackage{doi}
\usepackage{cleveref}
\usepackage{todonotes}
\usepackage{lscape}

\usepackage[table]{xcolor}   
\usepackage{adjustbox}
\usepackage{multirow}
\usepackage{arydshln} 
\usepackage{colortbl}
\usepackage{ulem}

\newcommand{\best}[1]{\textbf{#1}}

\newcommand{\ben}{\begin{enumerate}}
\newcommand{\een}{\end{enumerate}}

\begin{document}

\maketitle

\begin{abstract}
Operator learning methods such as DeepONets and FNOs often struggle with PDE families featuring sharp interfaces, heterogeneous coefficients, and localized multiscale structures.
We introduce a partition-of-unity (POU) mixture-of-experts framework for localized operator learning, in which geometry-aware gating networks produce smooth spatial partitions  
\rthreeedits{ which blend the contributions of local expert networks.}
Our main contribution is HiRefPOU, a \roneedits{residual-style} hierarchical POU architecture for DeepONets that \roneedits{organizes localized representations through nested parent-child partitions} while preserving global continuity.
We also show that the same POU principle can be incorporated into Fourier Neural Operators to introduce spatial adaptivity without modifying the underlying spectral layers.
On heterogeneous Darcy and reaction--diffusion benchmarks, HiRefPOU achieves substantially lower error than global DeepONet and static POU-MoE baselines, while the broader operator-learning experiments show that the benefits of localization depend on the PDE structure and the chosen neural-operator backbone.
The learned partitions are interpretable and align with interfaces and regions of rapid solution variation.
These results show that explicit geometric localization can improve both accuracy and interpretability in neural operator learning.
\end{abstract}


\section{Introduction} \label{sec:intro}
The emergence of machine learning (ML)-based partial differential equation (PDE) solvers has enabled the efficient and data-driven modeling of complex physical systems. 
Among these are neural operator frameworks, such as Deep Operator Networks (DeepONets)~\citep{lu2021Learning} and Fourier Neural Operators (FNOs)~\citep{li2020fourier}, which learn mappings between function spaces and capture the input-output behavior of entire PDE families. 
These models have achieved notable success across a range of domains, including fluid dynamics, porous media flow, and materials science, where scalable operator-learning approaches are increasingly central to practical surrogate modeling in scientific computing. 

Neural operators, while designed to learn mappings between families of PDEs, can encounter limitations when approximating solutions with localized, multi-scale, or heterogeneous structures.
In practice, many neural operator architectures rely on globally supported components, such as dense neural networks defined across the full problem domain or spectral expansions, which tend to under-represent high-frequency or spatially localized features.
While such global formulations preserve smoothness and continuity, they limit the network’s ability to 
allocate \rthreeedits{representational capacity} to regions of higher solution complexity, often resulting in an insufficient representation of fine-scale structure, leading to reduced accuracy in these areas.

A natural remedy is to introduce locality through domain decomposition or mixture-of-experts formulations.
In the physics-informed setting, methods such as APINNs~\citep{hu2023AugmentedPINN}, XPINNs~\citep{jagtap2020extended}, and POU-PINNs~\citep{rodriguez2024Partition} decompose the domain into subregions, while recent ensemble DeepONet models~\citep{SharmaShankar2025ensembledeeponet} introduce partition-of-unity (POU) mixtures on the operator-learning side.
These approaches demonstrate the value of localization, but they are typically fixed, single-level, or only weakly structured geometrically, and therefore do not provide \roneedits{learned nested hierarchical partitioning} for neural operator learning.

On the operator-learning side, ensemble and mixture-based formulations have emerged for introducing locality while preserving a global input--output structure.
In particular, recent ensemble DeepONet architectures~\citep{SharmaShankar2025ensembledeeponet} decompose the trunk network (i.e., the evaluation-coordinate encoder) into a mixture of expert components satisfying a partition-of-unity (POU) constraint, and demonstrate that combining multiple operator experts improves expressivity relative to both single global models and comparably parameterized baselines.
These results provide strong motivation for incorporating geometrically structured ensembles into neural operator models.
However, these approaches remain limited to flat (single-level) decompositions and do not support \roneedits{learned nested parent–child spatial partitioning}.

POU constructions provide a natural mechanism for blending multiple localized models through smooth, overlapping weighting functions by introducing learnable partition functions that modulate the contributions of local subnetworks~\citep{lee2021Partition}.
Existing neural POU approaches have primarily been explored in the context of single-function approximation and have seen more limited development in operator-learning settings.
Moreover, these methods typically rely on flat partitioning schemes and do not jointly incorporate adaptive partitioning, geometric awareness, and expressive neural parameterizations within a hierarchical framework.

A complementary line of work introduces hierarchical refinement through residual network architectures, which improve accuracy in PDE problems with sharp gradients by incrementally correcting coarse global approximations.
Hybrid Residual PINNs (HyResPINNs)~\citep{cooley2025hyrespinns} exemplify this approach by refining a global solution using localized residual components.
While effective, such residual architectures rely on predefined combinations of neural and RBF-based components, leading to fixed, radially structured regions of influence and limiting the ability to form flexible, data-adaptive partitions or hierarchical mixture-of-experts decompositions.
Taken together, these observations motivate the extension of hierarchical decomposition principles from solution learning to operator learning.

We address this limitation with a hierarchical POU framework for operator learning that combines smooth learnable partitions, localized experts, and \roneedits{additive multi-level spatial representations}.
The resulting architecture preserves global continuity while allowing \roneedits{spatially varying contributions from localized experts}.

Although the proposed framework is also relevant to physics-informed settings, the main contributions of this paper are in localized neural operator learning.
Our architectural developments, mathematical formulation, and numerical studies focus on DeepONet- and FNO-type models, while PINN-related discussions are included primarily for motivation and conceptual contrast.

Specifically, this paper makes the following contributions:
\begin{itemize}
    \item \done{We propose \METHODNAME, a hierarchical partition-of-unity framework for localized neural operator learning.}
    \item We introduce a hierarchical POU architecture and study one-, two-, and three-level instantiations, with the two-level formulation serving as the primary architectural case and the three-level variant used to probe \rtwoedits{the effect of an additional nested partition level.}
    \item \done{We embed the proposed POU mechanism in DeepONet and FNO architectures and evaluate it on a range of operator-learning benchmarks, with a targeted PINN case study included in the appendix.}
    \item 
    \rtwoedits{Across these experiments, we show that learned POU localization improves accuracy relative to global and fixed-partition DeepONet baselines on several heterogeneous PDE benchmarks. Comparisons among the 1L, 2L, and 3L variants further show that the benefit of organizing these localized representations hierarchically is problem-dependent and does not increase monotonically with depth.}
\end{itemize}

\rtwoedits{
\textbf{Novelty scope.} Partition-of-unity approximation, mixture-of-experts weighting, learned domain decomposition, additive global–local representations, hierarchical spatial partitioning, POD augmentation, and DeepONet/FNO backbones are established ideas and are not individually introduced in this work. 
}
\rthreeedits{
Accordingly, we position HiRefPOU as a localized neural-operator architecture
rather than as a fundamentally new operator-learning paradigm.
}
\rtwoedits{
The contribution of HiRefPOU is their specific integration for localized neural operator learning: a geometry-aware hybrid gate learns smooth coordinate-dependent partitions; child weights are constrained through parent POU weights; the resulting global and level-wise local trunk components are combined additively; and the same geometric POU principle is studied across DeepONet- and FNO-type backbones. 
The empirical contribution is a systematic evaluation of when this localization bias is useful and when additional hierarchy or POU augmentation provides limited benefit.
}

The remainder of this paper is organized as follows. 
Section~\ref{sec:related} reviews related work in physics-informed and operator-learning architectures.
Section~\ref{sec:prelims} introduces the relevant background related to PINNs, operator learning, and hierarchical POU models and their mathematical formulation.
Section~\ref{sec:methods} presents the hierarchical POU architecture and geometry-aware gating formulation.
Section~\ref{sec:methods_noinstants} describes its instantiation in DeepONet and FNO architectures and analyzes computational complexity.
Section~\ref{sec:expdetails} outlines the numerical experiments, baseline methods and relevant implementation details, while Section~\ref{sec:experiments1} discusses results and Section~\ref{sec:conclusion} concludes.

\section{Related Works} \label{sec:related}
This section reviews prior work most relevant to the proposed hierarchical partition-of-unity (POU) framework: physics-informed neural networks, neural operators, domain decomposition and POU methods, mixture-of-experts architectures, and hierarchical neural representations.
Our emphasis is on localized operator learning, which is the primary methodological and empirical focus of this paper.

\subsection{Physics-Informed Neural Networks (PINNs)}
In the context of PDE solving, PINNs represent the most direct attempt to approximate physical laws through global neural surrogates~\citep{raissi2017physicsI,raissi2017physicsII} and approximate the solutions of PDEs by embedding physical laws directly into the neural network’s loss function.
They offer a flexible, mesh-free framework in which expressivity arises from the architecture, activation functions, and optimization process~\citep{hornik1989multilayer,cybenko1989approximation}.
While universal approximation theorems establish that sufficiently large networks can represent any continuous function, realizing this theoretical expressivity in practice remains challenging~\citep{devore2020neuralnetworkapproximation,adcock2021gaptheorypracticefunction,marwah2021parametriccomplexityboundsapproximating}.
In particular, standard PINNs employ globally coupled representations, where parameters are updated according to a loss aggregated over the entire domain.
This implicit, global adaptivity contrasts sharply with the localized refinement mechanisms used in classical numerical methods and limits the network’s ability to capture sharp gradients, discontinuities, or multiscale behaviors~\citep{wang2022ntk}.

\subsection{Neural Operator Learning}
Neural operators learn mappings between function spaces and provide data-driven surrogates for PDE families.
Two prominent examples are DeepONets~\citep{lu2021Learning}, which combine branch and trunk networks to approximate nonlinear operators, and Fourier Neural Operators~\citep{li2020fourier}, which learn operators through spectral convolutions.
Despite their success, standard neural operators remain largely global representations.
DeepONets rely on global trunk bases, while FNOs rely on globally supported Fourier modes; both can struggle with localized, heterogeneous, or multiscale structure.
Recent extensions introduce multiresolution kernels, adaptive discretizations, or localized spectral filters, but these approaches often remain computationally expensive or lack clear geometric interpretability.
This motivates architectures that combine global operator structure with adaptive spatial localization.

\subsection{Domain Decomposition and Partition of Unity Methods}
Recent extensions of PINNs mitigate optimization and scalability challenges through domain decomposition, which divides the computational domain into smaller subregions for localized approximation and
\rthreeedits{can support} parallelization.
Methods such as XPINNs~\citep{jagtap2020extended} and DT-PINNs~\citep{DTPINNs} still rely largely on fixed or manually designed partitions, while soft decomposition approaches use adaptive weighting schemes to reduce hand-crafted partitioning~\citep{barbara2024Adaptive,hu2023AugmentedPINN}.
These results highlight the value of learnable localization, but most existing approaches remain non-hierarchical and use standard MLP-based gating.

Partition-of-unity (POU) methods provide a classical way to build globally continuous approximations from overlapping local components.
In finite element and meshfree settings, the POU constraint enables local adaptivity while preserving global consistency, making it a natural foundation for machine-learning-based PDE solvers.
Unlike xPINN-style methods, which train separate models on disjoint subdomains and enforce consistency through interface penalties, our POU formulation constructs a single global approximation as a weighted sum of local experts.
The weights are learned jointly with the solution, remain continuous, and sum to one over the domain, so continuity is enforced implicitly while \rtwoedits{the learned partitions can align with spatial structure that recurs across the training distribution.}

\subsection{Mixture-of-Experts and Adaptive Gating Architectures}
Mixture-of-experts (MoE) models provide an adaptive counterpart to domain decomposition by representing a global approximation as a weighted combination of specialized subnetworks.
In scientific machine learning, this allows different experts to focus on distinct spatial regions, physical regimes, or modes of variation while remaining coupled through a shared training objective.
Related ideas have appeared in both PINN and operator-learning settings, where learnable gating is used to decompose complex mappings into simpler local components~\citep{hu2023AugmentedPINN,wang2025operatormixture,SharmaShankar2025ensembledeeponet}.
In particular, the ensemble DeepONet framework of~\citet{SharmaShankar2025ensembledeeponet} demonstrates that partitioning the trunk representation into localized experts can improve expressivity and accuracy relative to purely global operator networks, providing direct motivation for localized operator learning.

Despite this promise, standard MoE formulations remain limited for PDE and operator-learning problems.
Most gating networks are implemented as generic neural modules with little explicit geometric structure, making the learned partitions difficult to interpret and not necessarily aligned with meaningful physical subregions~\citep{lee2021Partition}.
Moreover, gating is often performed in latent feature space rather than directly in the spatial coordinates, which can hinder spatial coherence and reduce control over how experts divide the domain.
Even structured alternatives such as RBF-based gating typically impose purely radial supports, which can be too restrictive for anisotropic or irregular localized behavior.
These limitations suggest that adaptive specialization alone is not sufficient: for PDE and operator-learning problems, the gating mechanism should also be geometry-aware and spatially interpretable.


\rtwoedits{
The closest operator-learning predecessor is the POU-MoE DeepONet of ~\citet{SharmaShankar2025ensembledeeponet}, which also constructs a localized trunk by blending independent local trunk networks through POU weights. 
Their reported formulation uses a flat collection of overlapping patches with prescribed centers and radii and normalized compactly supported kernel weights. Compact support permits local expert evaluation because each coordinate lies in only a subset of patches. HiRefPOU retains the localized-trunk interpretation but differs in four principal ways. First, its partition parameters are learned jointly through a hybrid center-distance and neural gate rather than specified as fixed kernel patches. Second, its learned weights may deform beyond purely radial compact supports. Third, it supports nested parent–child POU weights and additive level-wise trunk contributions. Fourth, the current HiRefPOU implementation evaluates all experts densely, rather than exploiting compact support for sparse local execution. Thus, the principal extension is not the introduction of a POU-MoE trunk itself, but the learned geometric gating and nested organization studied within neural-operator architectures.
}

\subsection{Multi-Fidelity and Hierarchical Neural Architectures}
Hierarchical and multi-fidelity architectures address a complementary limitation of global neural PDE models: many problems exhibit structure across multiple spatial or temporal scales and therefore benefit from coarse-to-fine refinement.
This principle is classical in numerical analysis, where multigrid, wavelet, and adaptive finite element methods improve approximation quality by combining coarse global structure with targeted fine-scale correction~\citep{gerhard2022wavelet,restrepo1995wavelet,mitchell1991adaptiverefinement}.
Analogous ideas have emerged in scientific machine learning through multi-level PINNs, multi-fidelity surrogates, and hierarchical neural operators, where progressively refined subnetworks or feature maps improve accuracy and stability~\citep{howard2023Multifidelity}.

Among the most closely related ideas are residual hierarchical architectures, which refine a coarse approximation through additional localized correction terms.
For example, HyResPINNs~\citep{cooley2025hyrespinns} combine global and local components in a residual hierarchy to improve approximation of PDE solutions with sharp gradients and heterogeneous structure.
This provides strong evidence that hierarchical refinement can be beneficial in scientific machine learning.
However, such approaches are primarily designed for solution learning rather than operator learning, and their local regions of influence are typically induced by predefined architectural choices rather than by an adaptive partition-of-unity decomposition learned jointly from data.

More broadly, most existing hierarchical neural architectures operate over fixed scales or latent feature hierarchies rather than over learned geometric partitions of the physical domain.
As a result, they provide multiscale refinement without explicitly learning where localized specialization should occur.
This is the main distinction we emphasize in the present work.
Our framework combines \roneedits{additive level-wise representations} with geometry-aware partition-of-unity gating, yielding a model in which 
\roneedits{localized representations are organized through learned nested spatial partitions rather than through fixed scales or latent feature hierarchies.}
Viewed this way, the proposed method bridges two strands of prior work: localized expert decompositions for operator learning~\citep{SharmaShankar2025ensembledeeponet} and residual hierarchical refinement in scientific machine learning~\citep{cooley2025hyrespinns}.

\paragraph{Summary.} 
\begin{table*}[t]
\color{black}
\centering
\caption{
\rtwoedits{
Comparison of HiRefPOU with closely related localization and domain-decomposition
approaches. 
Execution refers to the implementation described in the cited work.
}
}
\label{tab:related_work_comparison}
\renewcommand{\arraystretch}{1.15}
\small
\begin{adjustbox}{max width=\textwidth}
\begin{tabular}{lccccc}
\toprule
\textbf{Method}
&
\textbf{Primary setting}
&
\makecell{\textbf{Partition /}\\\textbf{localization}}
&
\textbf{Organization}
&
\makecell{\textbf{Operator}\\\textbf{learning}}
&
\makecell{\textbf{Expert / local}\\\textbf{execution}}
\\
\midrule

POUnet~\cite{lee2021Partition}
&
Function regression
&
\makecell{Learned POU\\partition}
&
Flat
&
No
&
\makecell{Dense weighted\\local basis}
\\


XPINN~\cite{jagtap2020extended}
&
PINN solution learning
&
Prescribed subdomains
&
Flat
&
No
&
\makecell{Subdomain-local\\networks}
\\

APINN~\cite{hu2023AugmentedPINN}
&
PINN solution learning
&
Learned soft gate
&
Flat
&
No
&
\makecell{Dense weighted\\subnetworks}
\\

HyResPINN~\cite{cooley2025hyrespinns}
&
PINN solution learning
&
\makecell{Neural/RBF\\representation}
&
Residual blocks
&
No
&
Dense
\\

\midrule

POU-MoE DeepONet~\cite{SharmaShankar2025ensembledeeponet}
&
Operator learning
&
\makecell{Prescribed compact-\\support POU patches}
&
Flat
&
Yes
&
\makecell{Sparse local\\patches}
\\

GeoFNO~\cite{li2023fourier}
&
Operator learning
&
\makecell{Learned geometric\\deformation}
&
Global
&
Yes
&
Dense
\\

LocalFNO~\cite{liu2024neural}
&
Operator learning
&
\makecell{Localized integral /\\differential kernels}
&
\makecell{Layer-level\\locality}
&
Yes
&
\makecell{Dense local\\kernels}
\\

\textbf{HiRefPOU (ours)}
&
Operator learning
&
\makecell{Learned hybrid\\geometric POU}
&
\makecell{Flat or nested\\parent--child}
&
Yes
&
\makecell{Dense all-expert\\blending}
\\

\bottomrule
\end{tabular}
\end{adjustbox}
\end{table*}

Existing approaches typically provide either adaptive expert specialization without strong geometric structure, or hierarchical refinement without learned spatial partitioning.
In contrast, the proposed hierarchical POU framework combines geometry-aware, learnable partitions with \roneedits{additive level-wise representations organized through nested spatial partitions} in a single operator-learning architecture.
This yields a model that is globally continuous, spatially interpretable, and adaptively localized across \roneedits{nested spatial regions}, linking localized expert decompositions in operator learning~\citep{SharmaShankar2025ensembledeeponet} with residual hierarchical architectures in scientific machine learning~\citep{cooley2025hyrespinns}.
\rthreeedits{
From this perspective, HiRefPOU is best viewed as a localized neural-operator
architecture that introduces a particular spatial inductive bias into established
operator-learning backbones.
}

\section{Preliminaries}\label{sec:prelims}

\subsection{Problem Setup}
\done{
We consider PDE solution approximation and, more generally, operator learning between function spaces.
Let $\mathcal{D} = [0,T]\times\Omega \subset \mathbb{R}^{1+d}$, where $\Omega \subset \mathbb{R}^d$ is a bounded spatial domain with boundary $\partial \Omega$.
A general time-dependent PDE takes the form
\begin{align}
u_t + \mathcal{F}(u,\nabla u,\dots) = f,
\end{align}
subject to initial and boundary conditions $u(0,\mathbf{x}) = g(\mathbf{x})$ for $\mathbf{x}\in\Omega$ and $\mathcal{B}[u(t,\mathbf{x})]=0$ for $t\in[0,T]$ and $\mathbf{x}\in\partial\Omega$.
}

\done{
Operator learning seeks to approximate a mapping $\mathcal{G}:\mathcal{U}\to\mathcal{V}$ from admissible inputs $a\in\mathcal{U}$ to solution functions $s\in\mathcal{V}$.
We write the governing problem abstractly as
\begin{displaymath}
\mathcal{N}(a,s)=0, \qquad \mathcal{B}(a,s)=0,
\end{displaymath}
and assume that each admissible $a$ defines a unique solution $s=s(a)$, inducing the solution operator $\mathcal{G}: a \mapsto s$.
}

\done{
Given paired data $\{a^{(i)}, s^{(i)}\}_{i=1}^N$, the goal is to learn $\mathcal{G}_\theta \approx \mathcal{G}$ so that unseen inputs can be mapped to their corresponding solutions without solving the PDE from scratch.
DeepONets and FNOs implement this idea through different parameterizations of the input-function representation and output evaluation map.
}

\subsection{Physics-Informed Neural Networks (PINNs)}
\done{
The general formulation of PINNs~\citep{raissi2018jmlr} aims to approximate the unknown solution $u(t, \mathbf{x})$ using a parameterized model $u_{\theta}(t, \mathbf{x})$, such that $u_{\theta}(t, \mathbf{x}) \approx u(t, \mathbf{x})$ and $\theta$ denotes the set of all trainable parameters of the network. 
The parameters $\theta$ are optimized to minimize the composite loss function:
\begin{align}
    L_{\roneedits{pinn}}(\theta) = \lambda_{ic}L_{ic} + \lambda_{bc}L_{bc} + \lambda_{r}L_{r}, \label{eq:full_loss}
\end{align}
where $L_{ic}$, $L_{bc}$, and $L_{r}$ represent the loss components associated with the initial conditions, boundary conditions, and the residual of the PDE, respectively; and $\lambda_{ic}$, $\lambda_{bc}$, and $\lambda_r$ are corresponding weighting coefficients that balance the contribution of each loss term. 
These terms can be set as static weights or adapted during training using various techniques such as NTK weighting~\citep{wang2021eigenvector} or gradient annealing~\citep{wang2021siam}, among others. 
}

\done{
Each loss term in Equation~\eqref{eq:full_loss} is computed as the mean squared error of the initial condition, boundary, and PDE residuals. 
Specifically, each loss term is defined as: 
\begin{align}
    L_{ic} &= \frac{1}{N_{ic}} \sum_{i=1}^{N_{ic}}
\left| u_\theta(0, \mathbf{x}_{ic}^i) - g(\mathbf{x}_{ic}^i) \right|^2, \\
    L_{bc} &= \frac{1}{N_{bc}} \sum_{i=1}^{N_{bc}} 
\left| \mathcal{B}[u_\theta]( t_{bc}^i, \mathbf{x}_{bc}^i) \right|^2, \\
    L_r &= \frac{1}{N_r} \sum_{i=1}^{N_r} 
\left| \frac{\partial u_{\mathbf{\theta}}}{\partial t}(t^i_r, \mathbf{x}^i_r) + \mathcal{F}[u_{\mathbf{\theta}}](t^i_r, \mathbf{x}^i_r)- f(t^i_r, \mathbf{x}^i_r) \right|^2.
\end{align}
The training data points $\{ \mathbf{x}_{ic}^i \}_{i=1}^{ N_{ic}}$, $\{t_{bc}^i, \mathbf{x}_{bc}^i\}_{i=1}^{N_{bc}}$ and $\{t_{r}^i, \mathbf{x}_{r}^i\}_{i=1}^{N_{r}}$ can be drawn from a fixed set or resampled randomly at each iteration of a gradient descent algorithm. 
}

\subsection{Deep Operator Networks (DeepONets)}
\done{
Deep Operator Networks (DeepONets) are designed to approximate nonlinear operators $\mathcal{G} : \mathcal{U} \to \mathcal{V}$ by decomposing the mapping into two subnetworks: the branch network and the trunk network. 
The branch network $B_\theta$ takes as input a finite-dimensional representation of the function $a \in \mathcal{U}$, for example
\begin{displaymath}
\mathbf{a} = (a(x_1),\dots,a(x_m))
\end{displaymath}
evaluated at fixed sensor locations $\{x_i\}_{i=1}^m$, and produces a latent coefficient vector
\begin{displaymath}
  \mathbf{b}(a) = \big( b_1(\mathbf{a}), b_2(\mathbf{a}), \dots, b_q(\mathbf{a}) \big)^\top \in \mathbb{R}^q.
\end{displaymath}
The trunk network $T_\phi$ takes as input the evaluation coordinate $y \in \mathbb{R}^p$ and outputs a latent basis vector,
\begin{displaymath}
  \mathbf{t}(y) = \big( t_1(y), t_2(y), \dots, t_q(y) \big)^\top \in \mathbb{R}^q.
\end{displaymath}
The output of the DeepONet for a given $a$ and $y$ is obtained by merging the branch and trunk outputs through an inner product:
\begin{displaymath}
G_{\theta}(a)(y) = \sum_{k=1}^q b_k\big(a(x_1), \dots, a(x_m)\big) \, t_k(y) + c,
\end{displaymath}
where $c \in \mathbb{R}$ is an optional bias term.
For vector-valued outputs of dimension $n$, the latent dimension $q$ is partitioned as $0 = q_0 < q_1 < \dots < q_n = q$, and each component is computed as
\begin{displaymath}
  G_{\theta}^{(i)}(a)(y) = \sum_{k=q_{i-1}+1}^{q_i} b_k\big(a(x_1), \dots, a(x_m)\big) \, t_k(y) + c, \quad i = 1, \dots, n.
\end{displaymath}
This formulation enables DeepONets to flexibly approximate both scalar- and vector-valued operators.
}

To quantify the discrepancy between the DeepONet prediction and the true PDE solution, we define the pointwise mean-squared error for a given input $a$ as
\begin{displaymath}
   \mathcal{L}(a, \theta) = \frac{1}{P} \sum_{j=1}^{P} 
    \left| G_{\theta}(a)(y_j) - s(y_j) \right|^2,
\end{displaymath}
where $\{y_j\}_{j=1}^P$ are evaluation locations in the domain of $G(a)$, and $s(y_j)$ denotes the corresponding PDE solution values obtained from the reference model. 
The total training objective is then obtained by averaging over all $N$ samples in the dataset:
\begin{align}
    \mathcal{L}_{\roneedits{ol}}(\theta) = \frac{1}{N} \sum_{i=1}^N \mathcal{L}\big(a^{(i)}, \theta\big). \label{eq:ol_loss}
\end{align}

\subsection{Fourier Neural Operators}
\done{
Fourier Neural Operators (FNOs) are another class of operator networks. 
The central building block of an FNO is the Fourier layer, which replaces traditional 
local convolutions with multiplications in the frequency domain. 
Given an intermediate feature representation $v(x)$ defined on a discretized domain, 
the Fourier layer computes
\begin{equation}
  v'(x) \;=\; \mathcal{F}^{-1}\!\big( R_\theta(\hat v(\xi)) \big)(x) \;+\; W v(x),
\end{equation}
where $\hat v(\xi) = \mathcal{F}[v](\xi)$ is the Fourier transform of $v$, 
$R_\theta$ denotes a learnable spectral multiplier acting on a truncated set of modes, 
and $W$ is a trainable pointwise linear transformation. 
The inverse Fourier transform $\mathcal{F}^{-1}$ maps the modified representation back to the spatial domain.
An FNO consists of a sequence of Fourier layers interleaved with pointwise nonlinearities. 
The network is typically structured as
\[
  v^{(0)} = P(a), \quad 
  v^{(\ell+1)} = \sigma\!\left( \mathcal{F}^{-1}\big( R_\theta(\hat v^{(\ell)}) \big) + W v^{(\ell)} \right),
\]
where $P$ is a lifting operator that maps the input $a$ into a higher-dimensional latent space, 
and $\sigma$ is a nonlinearity such as $\tanh$ or ReLU. 
A final projection operator $Q$ maps the latent representation back to the output space:
\[
  \mathcal{G}_\theta(a) = Q(v^{(L)}).
\]
By operating directly in the Fourier domain, FNOs efficiently capture 
long-range dependencies and global interactions, which are challenging for purely local architectures. 
However, standard FNOs rely on globally supported Fourier modes and lack explicit mechanisms for localized or adaptive refinement, which can degrade performance near sharp gradients or localized features.
}

\paragraph{Unified perspective.}
Despite their different training settings, PINNs, DeepONets, and FNOs share a common limitation: they rely primarily on global parameterizations.
This can hinder accuracy when the target solution or operator contains fine-scale or spatially localized structure.
Partition-of-unity models address this issue by blending local components while preserving global continuity.
The approach developed here extends that idea by learning \roneedits{geometry-aware partitions organized through nested parent--child relationships and combining their spatially weighted expert contributions additively across levels.}

\subsection{Partition-of-Unity and Mixture-of-Experts Formulations}\label{sec:pou_formulation}
\done{
Partition-of-unity (POU) methods provide a general mathematical framework for constructing globally continuous approximations from a collection of local approximants.  
Originally developed in the context of finite element and mesh-free methods
, the POU principle ensures global smoothness by blending overlapping local basis functions through nonnegative weighting functions that sum to one over the domain.
Formally, let $\{\phi_j(\mathbf{x})\}_{j=1}^{N_p}$ denote a set of partition functions satisfying
\begin{align}
  \phi_j(\mathbf{x}) \ge 0, 
  \qquad \text{and} \qquad 
  \sum_{j=1}^{N_p} \phi_j(\mathbf{x}) = 1, 
  \quad \forall\, \mathbf{x} \in \Omega.
  \label{eq:pouproperty}
\end{align}
Each $\phi_j$ is supported primarily on a subregion $\Omega_j \subset \Omega$,
with overlapping supports across neighboring partitions.
}

\done{
Mixture-of-Experts (MoE) models provide a closely related perspective by representing a global function as a weighted combination of local approximants.
Given a collection of expert functions $\{f_j(\mathbf{x})\}_{j=1}^{N_c}$, the global approximation is expressed as
\begin{align}
  u(\mathbf{x}) = \sum_{j=1}^{N_c} \phi_j(\mathbf{x}) \, f_j(\mathbf{x}),
  \label{eq:pou_classic}
\end{align}
where the weighting functions $\phi_j$ act as a gating mechanism.
When the weights satisfy the partition-of-unity conditions~\eqref{eq:pouproperty}, the resulting model inherits the stability and smoothness properties of classical POU formulations.
}

\done{
In this context, the POU framework can be interpreted as a structured MoE model in which the gating functions are explicitly constrained to be nonnegative and to sum to one.
The experts $\{f_j\}$ may be linear models, radial basis functions, multilayer perceptrons, or other specialized subnetworks.
When the experts are nonlinear subnetworks, the model becomes a locally adaptive ensemble in which each expert specializes in a region of the input space, while smooth blending across regions is guaranteed by the partition-of-unity property.
}

\done{
In practice, the gating functions $\{\phi_j(\mathbf{x})\}$ may be parameterized through normalized radial basis functions, distance-based kernels, or other geometry-aware constructions.
In this work, we define the gating functions using learnable, spatially structured mechanisms, which combine RBF activations with box initialization-based partitions---which are described in detail in Section~\ref{sec:methods}.
This design preserves the interpretability and continuity guarantees of classical POU methods while enabling data-driven, adaptive partitioning within a neural network framework.
}

\begin{figure*}[ht]
\centering
\resizebox{0.90\textwidth}{!}{  
\begin{tikzpicture}[
    >=latex,
    node distance=1.6cm and 3.0cm,
    block/.style={draw, thick, minimum width=3cm, minimum height=1.1cm, align=center},
    smallblock/.style={draw, thick, minimum size=0.9cm, align=center},
    sum/.style={draw, circle, thick, minimum size=0.9cm, inner sep=0pt},
    teal/.style={draw=teal, -{Latex[length=2mm]}},
    blackarrow/.style={->, thick},
    every node/.style={font=\small}
]

\node[smallblock] (embed) at (0,0) {$e$};
\node[smallblock, above=1.2cm of embed] (vnode) {$v$};
\node[smallblock, below=1.2cm of embed] (unode) {$u$};

\node[draw=orange, very thick, rounded corners,
      fit=(vnode) (embed) (unode),
      inner sep=6pt] (Abox) {};

\node[circle, draw, left=2cm of embed] (xin) {$x$};


\draw[blackarrow] (xin) -- (embed.west);
\draw[blackarrow] (embed) -- (vnode);
\draw[blackarrow] (embed) -- (unode);

\node[block, above right=1.0cm and 4.0cm of embed] (global) {Global};
\node[block, below right=0.5cm and 4.0cm of embed] (coarse) {Coarse\\POU--MoE};
\node[block, right=1.0cm of coarse] (fine)  {Fine\\POU--MoE};

\node[draw=green, very thick, rounded corners,
      fit=(coarse) (fine),
      inner sep=6pt] (Bbox) {};

\node[circle, draw, below left=1.0cm and 1.25cm of coarse] (xin2) {$x$};
\draw[blackarrow] (xin2) -- (coarse.west);



\draw[teal] (vnode.east)  to[out=0,in=120] (global.west);
\draw[teal] (vnode.east)  to[out=0,in=120] (coarse.west);
\draw[teal] (vnode.east)  to[out=0,in=120] (fine.west);

\draw[teal] (unode.east)  to[out=0,in=-120] (global.west);
\draw[teal] (unode.east)  to[out=0,in=-120] (coarse.west);
\draw[teal] (unode.east)  to[out=0,in=-120] (fine.west);

\draw[blackarrow] (embed.east) -- (coarse.west);
\draw[blackarrow] (embed.east) -- (global.west);


\node[sum,    right=13.0cm of embed] (sum) {};       
\node[circle, draw, right=1.0cm of sum] (uhat) {$\hat u$};

\draw[blackarrow] (global.east) -- (sum.west);
\draw[blackarrow] (coarse.east) -- (fine.west);
\draw[blackarrow] (fine.east)   -- (sum.west);

\draw[blackarrow] (sum) -- (uhat);

\end{tikzpicture}
 }
\caption{\small 
\roneedits{Residual-style} hierarchical architecture used in \METHODNAME.
The input $x$ is embedded and passed through a global backbone to obtain $u_{\mathrm{g}}(x)$.  
A coarse Mixture-of-Experts (MoE) layer predicts $u_{c}(x)$.
A second MoE layer predicts the fine-\roneedits{level} contributions $u_{f}(x)$ \roneedits{through nested parent--child partitioning.} 
The final output is the sum of all three components, $u(x) = u_{\mathrm{g}}(x) + u_{c}(x) + u_{f}(x)$.
}
\label{fig:full_arch}
\end{figure*}
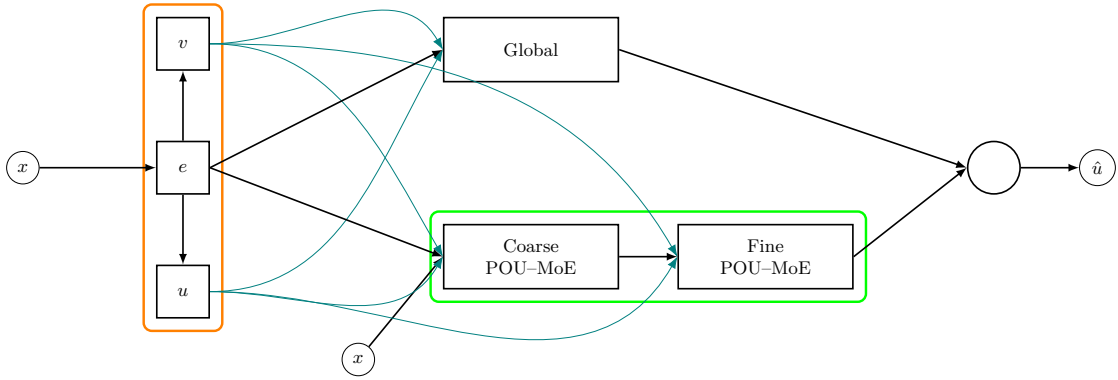

\section{Hierarchical Partition-of-Unity Models for Adaptive Operator Learning}
\label{sec:methods}
We introduce \METHODNAME, a hierarchical \roneedits{POU} architecture that combines a global neural approximator with multi-level Partition-of-Unity Mixture-of-Experts (POU--MoE) subnetworks.
Figure~\ref{fig:full_arch} shows our full network architecture diagram. 
\roneedits{
Because several terms used below have distinct architectural meanings, Table~\ref{tab:terminology} summarizes the terminology used throughout this work.
The hierarchy is defined through the organization of the partition functions: each child partition is modulated by and locally normalized within a parent partition. 
This construction permits additional spatial subdivision within the regions emphasized by a parent gate while preserving smooth global aggregation.
The level outputs are combined through additive residual-style updates. 
However, all components are learned jointly from the final training objective, and the architecture does not require any level to approximate a detached residual left by the preceding levels. 
Thus, HiRefPOU should be interpreted as a nested spatial representation rather than as a strict spectral or orthogonal multiresolution decomposition.
}

\roneedits{
Throughout this work, the term `hierarchical' refers to the nested parent–child organization of the POU weights, and residual refers to the additive composition of the level-wise outputs. 
The construction guarantees nested spatial weighting, smooth aggregation, and a globally continuous representation. 
It does not explicitly impose frequency separation, orthogonality, or stage-wise residual fitting between levels. 
Accordingly, the terms `coarse' and `fine' identify levels of the spatial partition hierarchy rather than predetermined frequency bands or uniquely identifiable components of the target operator.
}

\begin{table}[t]
\centering
\caption{
\roneedits{
Terminology used throughout this work. These definitions describe
the specific architectural meaning of each term in HiRefPOU and are not
intended as general definitions for all hierarchical or mixture-of-experts
models.}}
\label{tab:terminology}
\renewcommand{\arraystretch}{1.15}
\small
\begin{tabularx}{\linewidth}{l X}
\toprule
\textbf{Term} & \textbf{Meaning in this work} \\
\midrule

\rtwoedits{Adaptive}
&
\rtwoedits{The parameters defining the POU gates, including centers, scales, neural
gate parameters, and mixing coefficients, are learned jointly from the
training objective rather than prescribed a priori. In operator-learning
experiments, the gates depend only on evaluation coordinates and are shared
across input samples; they are therefore not instance-conditioned.}
\\

Hierarchical
&
The POU weights are organized through nested parent--child relationships.
A child partition is spatially weighted by its parent partition, producing
a nested POU structure across levels.
\\

Residual-style
&
Successive POU levels contribute additively to the model representation.
The term describes the additive architecture and does not imply stage-wise
training against detached residual targets.
\\

Coarse / fine level
&
Labels identifying relative levels of the spatial partition hierarchy.
They do not denote prescribed low- and high-frequency components or an
orthogonal multiresolution decomposition.
\\

Localization
&
Spatially varying POU weights assign different relative contributions to
local expert representations across the evaluation domain.
\\

Expert specialization
&
An expert may contribute more strongly in regions where its learned POU
weight is large. Specialization is an empirical property of the learned
representation rather than an imposed functional decomposition.
\\

Gating
&
When used in this work, spatial gating refers to soft weighting and blending
of expert outputs. The current implementation evaluates all experts densely
and does not perform sparse conditional routing or top-$k$ expert selection.
\\

\bottomrule
\end{tabularx}
\end{table}

\subsection{\roneedits{Residual-Style Hierarchical POU Representation}}
Formally, given an input $x \in \mathbb{R}^d$, the \roneedits{two-level HiRefPOU} model produces the \roneedits{representation}
\begin{equation}
    u_{\mathbf{\theta}}(x) 
    = u_{\mathrm{g}}(x; \theta_g) 
    + u_{c}(x; \theta_c) 
    + u_{f}(x; \theta_f), \label{eq:hirefpou_output}
\end{equation}
where $u_{\mathrm{g}}$ \roneedits{denotes the global component}, 
$u_{c}$ \roneedits{denotes the coarse-level POU component}, 
and $u_{f}$ \roneedits{denotes the fine-level POU component}. 
\roneedits{The parameter sets $\theta_g$, $\theta_c$, and $\theta_f$ collectively contain the trainable parameters associated with the corresponding global network, local experts, and gating networks.}

\roneedits{
We refer to this construction as residual-style because the POU levels are combined additively with the global component.
This terminology describes the architecture and does not imply stage-wise training against residual targets. 
In particular, all components are optimized jointly from the final training objective, and no frequency separation, orthogonality, or level-wise identifiability is imposed.
}

\roneedits{
The term hierarchical instead refers to the organization of the POU weights. 
Fine-level partitions are nested within coarse-level partitions through the parent--child construction introduced in Section~\ref{sec:hierarchical_pou_refinement}. 
Accordingly, the labels coarse and fine refer to relative levels of the spatial partition hierarchy rather than prescribed low- and high-frequency components of the target function.
}

\subsection{Global Approximator Network.}
The global approximator ($u_g(x; \theta_g)$) is implemented as a modified residual network that incorporates shared conditioning features to stabilize optimization and provides \roneedits{a globally supported coordinate representation}~\citep{wang2021siam,wang2024piratenets,rodriguez2024Partition}.
Let $e(x)\in\mathbb{R}^m$ denote the shared embedding of the input coordinate.
Starting from $g^{(0)}(x)=e(x)$, the global network applies a sequence of $L$ gated bottleneck blocks of the form
\begin{equation}
  g^{(\ell+1)}(x)
  = \mathcal{B}\!\left(g^{(\ell)}(x),\, h_u(x),\, h_v(x)\right),
  \qquad \ell=0,\dots,L-1,
\end{equation}
where $h_u(x)$ and $h_v(x)$ are shared conditioning vectors derived from the embedding.
Each bottleneck block consists of a stack of fully connected layers interleaved with nonlinear activations, followed by element-wise gating using the conditioning features.
Specifically, intermediate activations are modulated according to
\begin{equation}
  z \;\mapsto\; z \odot h_u(x) + \bigl(1 - z\bigr) \odot h_v(x),
\end{equation}
which enables the network to interpolate between two learned feature representations and improves stability across heterogeneous regions of the domain~\citep{wang2021siam}.

To promote stable training \roneedits{while allowing the transformed representation to be introduced gradually during optimization, each block is equipped with a learnable residual connection.}
Let $x_{\mathrm{in}}$ denote the block input and $x_{\mathrm{out}}$ its transformed output.
The final block output is given by
\begin{equation}
  \mathcal{B}(x_{\mathrm{in}}, h_u, h_v)
  = \alpha\, x_{\mathrm{out}} + (1-\alpha)\, x_{\mathrm{in}},
\end{equation}
where $\alpha$ is a scalar parameter initialized near zero and learned during training.
This design allows the network to begin close to the identity map and gradually increase expressivity as optimization proceeds~\citep{cooley2025hyrespinns,wang2024piratenets}.
A final linear projection produces the global prediction
\begin{equation}
  u_{\mathrm g}(x) = W_{\mathrm g} \, g^{(L)}(x).
\end{equation}
The resulting global component provides a smooth, \roneedits{globally supported baseline representation that is combined additively with the localized POU components.}

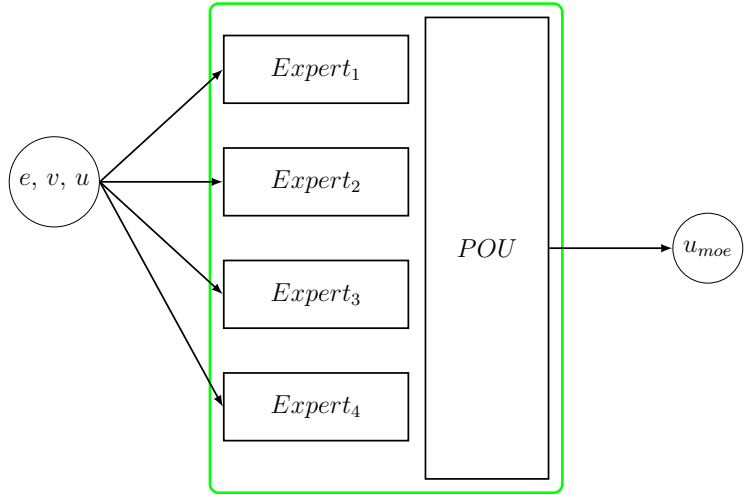
\begin{figure*}[ht]
\centering
\resizebox{0.60\textwidth}{!}{  
\begin{tikzpicture}[
    >=latex,
    node distance=1.6cm and 3.0cm,
    block/.style={draw, thick, minimum width=3cm, minimum height=1.1cm, align=center},
    largeblock/.style={draw, thick, minimum width=2cm, minimum height=7.5cm, align=center},
    smallblock/.style={draw, thick, minimum size=0.9cm, align=center},
    sum/.style={draw, circle, thick, minimum size=0.9cm, inner sep=0pt},
    teal/.style={draw=teal, -{Latex[length=2mm]}},
    blackarrow/.style={->, thick},
    every node/.style={font=\large}
]

\node[block] at (0,0) (exp2) {$Expert_2$};

\node[circle, draw, left=2cm of exp2] (exin) {$e$, $v$, $u$};

\node[block, above=0.7cm of exp2] (exp1) {$Expert_1$};
\node[block, below=0.7cm of exp2] (exp3) {$Expert_3$};
\node[block, below=0.7cm of exp3] (exp4) {$Expert_4$};

\node[largeblock, below right=-3.25cm and 0.25cm of exp2] (pou) {$POU$};

\node[draw=green, very thick, rounded corners, 
fit=(exp1) (exp2) (exp3) (exp4) (pou), inner sep=6pt] (Abox) {};

\node[circle, draw, right=2cm of pou] (out) {$u_{moe}$};

\draw[blackarrow] (exin.east) -- (exp1.west);
\draw[blackarrow] (exin.east) -- (exp2.west);
\draw[blackarrow] (exin.east) -- (exp3.west);
\draw[blackarrow] (exin.east) -- (exp4.west);

\draw[blackarrow] (pou.east) -- (out.west);

\end{tikzpicture}
 }
\caption{\small 
 A POU-Mixture-of-Experts (MoE) layer predicts $u_{p}(x)$.
}
\label{fig:poublock_arch}
\end{figure*}

\subsection{Geometry-Aware Gating Networks for Adaptive Domain Decomposition} \label{sec:gating}
A central contribution of this work is a geometry-aware gating network that combines  center-based distance partitions with box-initialized residual gating.
This design generalizes classical radial basis function (RBF) partitions while retaining the flexibility of learned, anisotropic domain decompositions.
Each partition is associated with a learnable center point $c_j \in \mathbb{R}^d$, which acts as an explicit geometric anchor, allowing for easier partition initialization, regularization, and interpretation. 
We then compute the Euclidean distances as,
\begin{equation}
  d_j^2(x) = \| x - c_j \|^2,
\end{equation}
which serve as intrinsic geometric features that define the proximity of each point to the available partitions.
These distances help define the center-based gating logits,
\begin{equation}
  z^{\mathrm{cntr}}_j(x)
  = -\frac{d_j^2(x)}{2 \alpha_j^2},
\end{equation}
where $\alpha_j>0$ is a learnable, per-center scaling parameter that controls the sharpness of the partition.
This formulation preserves the interpretability and smoothness of RBF-like partitions while allowing the effective support of each region 
\rtwoedits{to be learned and updated during training.}

To further increase the flexibility of the isotropic radial supports, we combine the center-based gating with a box-initialized residual gating network. 
This network consists of a shallow multilayer perceptron with residual connections, initialized such that its decision boundaries form approximately axis-aligned, non-overlapping hyperrectangular regions across the domain.
During training, these initial box-like partitions are progressively deformed through learned nonlinear transformations, enabling the gating to capture anisotropic and nonconvex structures.

Let $z^{\mathrm{box}}(x)$ denote the logits produced by the box-based gating network and $z^{\mathrm{cntr}}(x)$ those produced by the center-distance gate.
The final gating logits are obtained through a learnable convex combination
\begin{equation}
  z(x) = w\, z^{\mathrm{box}}(x) + (1 - w)\, z^{\mathrm{cntr}}(x),
\end{equation}
where $w = \text{sigmoid}(m)$ is a scalar mixing coefficient obtained by applying a sigmoid to a trainable parameter $m$ such that $w \in (0,1)$.
This \rtwoedits{learned} interpolation allows the model to transition smoothly between geometry-dominated and \rtwoedits{neural} gating regimes.

Finally, a softmax is applied to the combined logits to obtain smooth POU weights,
\begin{equation}
  \phi_j(x) = \frac{\exp(z_j(x))}{\sum_k \exp(z_k(x))},
\end{equation}
which satisfy $\sum_j \phi_j(x)=1$ for all $x$.
These weights define continuous domain partitions that are jointly optimized with the expert networks during training.

\rtwoedits{
Throughout this work, the term `adaptive' refers to the learned model parameters associated with the gating networks, including the learned center points $c_j$, scaling parameters $\alpha_j$, gate networks, mixing parameters associated with the POU weighting networks. 
These parameters are learned jointly with the other model parameters rather than being prescribed, or fixed prior to training. 
In the operator learning settings, the gating network does not receive the input functions or their branch-network representation, coefficient fields, or any other sample-specific information. 
Consequently, the learned partitions are shared across input functions, and represent dataset-level geometric adaptation, rather than instance-wise adaptation. 
This construction is well-suited for spatial structures that occur across consistent locations across the data distribution, but it cannot reposition the partitions to follow sample-dependent moving interfaces or features. 
This would be an interesting area for future development. 
}

Further, we emphasize that the gating construction above is written in the coordinate representation provided to the model.
In Euclidean domains, this corresponds directly to the physical coordinates.
For the three-dimensional reaction--diffusion benchmark considered here, the domain is the ball, not a manifold-valued surface.
Thus, the gating network is applied directly to the Euclidean coordinates used by the operator model, and we do not introduce any intrinsic surface parameterization, geodesic distance, or manifold-specific gating constraints.
Any locality is therefore learned in this ambient coordinate representation.

\subsection{Partition-of-Unity Mixture-of-Experts Ensembles}
Each POU level ($u_c$ and $u_f$ in Equation~\eqref{eq:hirefpou_output}) is implemented as a MOE ensemble composed of independently parameterized residual multilayer perceptrons.
Let $\{\phi_j(x)\}_{j=1}^{N_c}$ denote the partition weights produced by the gating network at a given level.
The corresponding expert ensemble produces local expert outputs $\{f_j(x)\}_{j=1}^{N_c}$, which are combined through a weighted sum.

Similar to the global approximator network ($u_g$), each expert $f_j$ is implemented as a residual modified MLP conditioned on the shared features $(h_u(x), h_v(x))$.
Given an input feature vector $x_0$, the expert applies a sequence of fully connected layers,
\begin{equation}
  x_{k+1}
  = \sigma\!\left(W_k x_k\right),
  \qquad k=0,\dots,D-1,
\end{equation}
where $\sigma$ denotes a nonlinear activation function.
As in the global network, each intermediate activation is modulated through the shared conditioning,
\begin{equation}
  x_k \;\mapsto\; x_k \odot h_u(x) + \bigl(1 - x_k\bigr) \odot h_v(x),
\end{equation}
which enables coordinated feature sharing across experts while preserving independent parameters.

To improve optimization and prevent overfitting in early training, each expert incorporates a learnable residual connection.
Let $x_{\mathrm{in}}$ denote the expert input and $x_{\mathrm{out}}$ the transformed output.
The residual update is given by
\begin{equation}
  \tilde{x} = \alpha\, x_{\mathrm{out}} + (1-\alpha)\, x_{\mathrm{in}},
\end{equation}
where $\alpha$ is a learnable scalar initialized near zero.
A final linear projection maps $\tilde{x}$ to a scalar expert output.

All experts at a given \rthreeedits{POU level contain independent parameters and are evaluated at every query coordinate. 
Each expert evaluation is executed in parallel due to batch vectorization.}
The ensemble output is then formed as,
\begin{equation}
  u_{\mathrm{POU}}(x)
  = \sum_{j=1}^{N_c} \phi_j(x)\, f_j(x),
\end{equation}
which guarantees smooth blending across partitions and preserves global continuity through the POU constraint.
This design enables each expert to specialize to localized features   
\rthreeedits{in regions with large POU weights} while the residual structure and shared conditioning promote coherent behavior across the ensemble and global network.
\rthreeedits{While parallelization of experts improves hardware utilization compare to serial execution, the arithmetic cost increases with additional experts. 
This implementation level parallelism does not constitute sparse routing, which is a direction for future work.}

\subsection{Geometric Regularization for Learned Partitions} \label{sec:geo_regularizers}
To encourage well-distributed domain partitions, we regularize the gating network and its learnable centers using lightweight geometric losses.
\roneedits{These regularizers are applied jointly with the full model, and do not require access to solution data.}

Let $X=\{x_i\}_{i=1}^N \subset \Omega$ denote a set of sample points and $C=\{c_j\}_{j=1}^{N_p}$ the corresponding set of learnable partition centers. 
For $\Omega \subset \mathbb{R}^d$, we define a generic coverage regularizer by comparing a geometric measure of the spread of the centers to the measure of the domain. 
In one dimension, this reduces to the length of the interval spanned by the centers; in two dimensions, to the area of their convex hull; and in three dimensions, to the volume of their convex hull. Denoting this measure generically by $M(C)$ and the corresponding domain measure by $M_\Omega$, we write
\begin{align}
L_{\mathrm{cov}} = \lambda_{\mathrm{cov}} \frac{|M(C)-M_\Omega|}{M_\Omega+\varepsilon}, \label{eq:cov_loss}
\end{align}
which discourages collapsed or poorly distributed center configurations.

For example, when $\Omega=[0,1]$, we may take
\begin{align}
M(C)= \max_j c_j - \min_j c_j, \qquad M_\Omega = 1.
\end{align}
When $\Omega=[0,1]^2$, $M(C)$ is the area of the convex hull of the centers, recovering the two-dimensional form used in this work. 
When $\Omega=[0,1]^3$, $M(C)$ is the volume of the convex hull of the centers.

We also apply a soft containment penalty to discourage centers from drifting outside \rtwoedits{a prescribed bounding box containing the computational domain.}
\rtwoedits{Let
\[
\mathcal{B}
=
\prod_{k=1}^{d}[a_k,b_k]
\]
denote an axis-aligned bounding box satisfying $\Omega \subseteq \mathcal{B}$.
We define the containment regularizer as
\begin{equation}
    L_{\mathrm{cont}}
    =
    \lambda_{\mathrm{cont}}
    \sum_{j=1}^{N_p}
    \sum_{k=1}^{d}
    \left[
        \max(0,a_k-c_{j,k})
        +
        \max(0,c_{j,k}-b_k)
    \right],
    \label{eq:cont_loss}
\end{equation}
where $c_{j,k}$ denotes the $k$th coordinate of center $\mathbf{c}_j$.
This penalty is zero when a center lies within the prescribed bounding box and
increases linearly when it moves outside the admissible coordinate range.
}

\rtwoedits{For rectangular domains, the bounding box coincides with the computational
domain. For nonrectangular domains, such as the three-dimensional unit ball
considered in this work, we use the corresponding axis-aligned enclosing box.
Thus, the containment term constrains the centers to remain within the
coordinate extent of the problem domain, but does not impose the exact geometry
of a nonrectangular boundary.}

\subsection{Hierarchical Partition-of-Unity Refinement} \label{sec:hierarchical_pou_refinement}
While a single-level POU
\roneedits{
model provides one collection of spatially weighted local experts, the hierarchical extension organizes additional POU levels through explicit parent--child relationships. 
The purpose of the hierarchy is to permit further spatial subdivision within the support
of a parent partition; it does not prescribe a frequency decomposition of the represented function.
}

Let $\{\phi^{c}_j(x)\}_{j=1}^{N_c}$ denote the coarse-level POU weights, produced by a gating network as described in the previous section.
These weights satisfy
\[
\sum_{j=1}^{N_c} \phi^{c}_j(x) = 1, \qquad \forall x \in \Omega.
\]
The coarse\roneedits{-level POU contribution} in Equation~\eqref{eq:hirefpou_output} is defined as
\begin{equation}
  u_{c}(x)
  =
  \sum_{j=1}^{N_c}
  \phi^{c}_j(x)\, f^{c}_j(x),
\end{equation}
where $f^{c}_j(x)$ denotes the expert associated with the $j$-th coarse partition.

To introduce \roneedits{a second level of nested spatial partitioning}, each coarse partition $\phi^{c}_j(x)$ is then subdivided by a second gating network that produces child weights $\{\psi_{j,k}(x)\}_{k=1}^{N_f}$, locally normalized within each parent partition.
The resulting fine-level POU weights are defined by
\[
\phi^f_{j,k}(x)
=
\phi^c_j(x)\, \psi_{j,k}(x),
\qquad
\sum_{k=1}^{N_f}\psi_{j,k}(x)=1,
\]
which guarantees the nested POU property
$\sum_{j=1}^{N_c}\sum_{k=1}^{N_f}\phi^f_{j,k}(x)=1$.
\roneedits{
The resulting parent--child construction is the sense in which the model is hierarchical: each fine-level weight inherits the support and spatial weighting of its parent through the multiplicative factor $\phi^c_j(x)$. 
The terms ``coarse'' and ``fine'' therefore refer to positions in this spatial partition tree.
}

In practice, this constraint is satisfied exactly by defining each fine-level weight as the product of its parent coarse weight and a locally normalized softmax over the children of that parent.
Each fine partition $\phi^{f}_{j,k}(x)$ is paired with its own local expert $f^{f}_{j,k}(x)$, yielding the fine-\roneedits{level POU contribution} from Equation~\eqref{eq:hirefpou_output} as, 
\begin{equation}
  u_{f}(x)
  =
  \sum_{j=1}^{N_c}
  \sum_{k=1}^{N_f}
  \phi^{f}_{j,k}(x)\, f^{f}_{j,k}(x).
  \label{eq:childparts}
\end{equation}

This structure mirrors the \roneedits{additive} skip-connection paradigm of residual networks, with the key distinction that \roneedits{the additional contributions are spatially weighted through nested POU partitions rather than added as unstructured features.}
The coarse\roneedits{ and fine levels provide first- and second-level localized contributions, respectively.}
Although we focus on a two-level hierarchy in this work, the construction extends naturally to deeper \roneedits{partition} trees by recursively subdividing partitions and \roneedits{adding corresponding POU level contributions.}
The full hierarchical POU model is trained end-to-end by minimizing the appropriate physics-informed or data-driven loss for the target application.

\section{Neural Operator Instantiations of the POU Framework} 
\label{sec:methods_noinstants}
The Partition-of-Unity (POU) framework extends naturally to neural operator architectures.

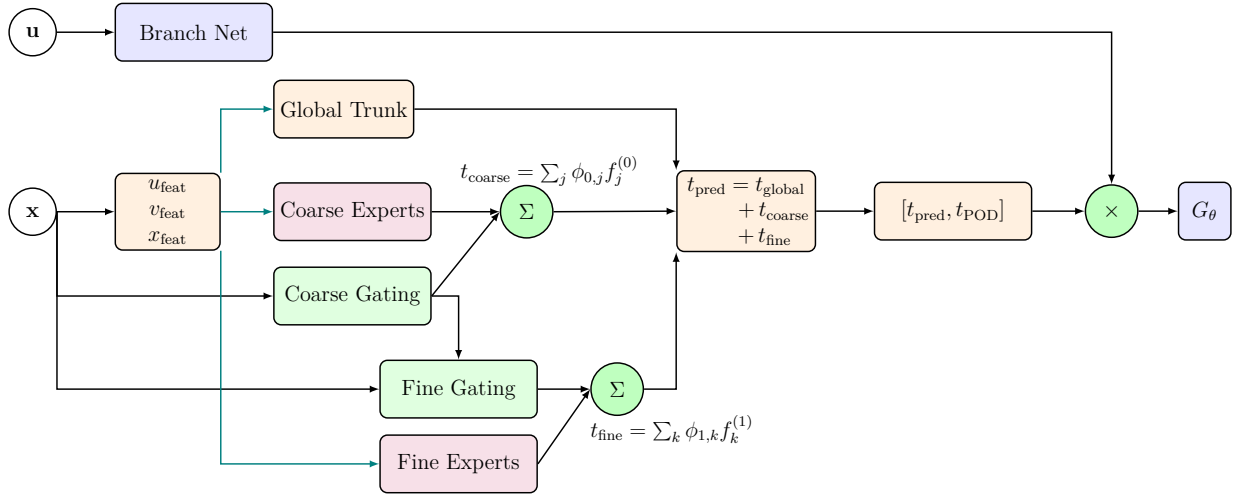
\begin{figure*}[ht]
\centering
\resizebox{0.99\textwidth}{!}{  
\begin{tikzpicture}[
    scale=0.9,
    every node/.style={font=\large},
    net/.style={draw, fill=blue!10, rounded corners, minimum height=1.1cm, minimum width=3.0cm, thick},
    trunknet/.style={draw, fill=orange!12, rounded corners, minimum height=1.1cm, minimum width=2.0cm, thick},
    gatenet/.style={draw, fill=green!12, rounded corners, minimum height=1.1cm, minimum width=3.0cm, thick},
    expertnet/.style={draw, fill=purple!12, rounded corners, minimum height=1.1cm, minimum width=3.0cm, thick},
    op/.style={circle, draw, fill=green!25, minimum size=1.0cm, thick},
    dottedbox/.style={rounded corners, dashed, draw=black!75, thick},
    arrow/.style={-latex, thick}
]

\node[draw, circle, minimum size=0.9cm, thick] (u) {$\mathbf{u}$};
\node[draw, circle, minimum size=0.9cm, thick, below=2.5cm of u] (x) {$\mathbf{x}$};

\node[net, right=1.1cm of u] (branchCNN) {
Branch Net 
};
\draw[arrow] (u) -- (branchCNN);

\node[trunknet, right=1.1cm of x, align=center] (condFeat) 
{
$u_{\text{feat}}$\\
$v_{\text{feat}}$\\
$x_{\text{feat}}$
};

\draw[arrow] (x) -- (condFeat);

\node[trunknet, above right=0.65cm and 1.0cm of condFeat] (globalTrunk) {
Global Trunk
};
\draw[arrow, teal] (condFeat.north east) |- node[above,pos=0.75,align=center] {} (globalTrunk.west);


\node[gatenet, below right=0.3cm and 1.0cm of condFeat] (coarseGate) {
Coarse Gating
};
\draw[arrow] (x.east) |- (coarseGate.west);

\node[expertnet, right=1.0cm of condFeat] (coarseExperts) {
Coarse Experts
};
\draw[arrow, teal] (condFeat.east) -- node[above,pos=0.55] {} (coarseExperts.west);

\node[op, right=1.3cm of coarseExperts] (sumCoarse) {$\Sigma$};
\node[above right=0.0cm and -1.75cm of sumCoarse] { $t_\text{coarse}=\sum_j\phi_{0,j} f^{(0)}_j$};

\draw[arrow] (coarseExperts.east) -- node[above,pos=0.5] {} (sumCoarse.west);
\draw[arrow] (coarseGate.east) -- node[right,pos=0.4] {} (sumCoarse.west);


\node[gatenet, below right=0.65cm and -1.0cm of coarseGate] (fineGate) {
Fine Gating 
};
\draw[arrow] (x.east) |- (fineGate.west);

\node[expertnet, below=0.3cm of fineGate] (fineExperts) {
Fine Experts
};

\draw[arrow, teal] (condFeat.south east) |- node[above,pos=0.70] {} (fineExperts.west);

\node[op, right=1.0cm of fineGate] (sumFine) {$\Sigma$};
\node[below right=0.0cm and -1.0cm of sumFine] {$t_\text{fine}=\sum_k\phi_{1,k} f^{(1)}_k$};

\draw[arrow] (fineExperts.east) -- node[above,pos=0.5] {} (sumFine.west);
\draw[arrow] (fineGate.east) -- node[below,pos=0.4] {} (sumFine.west);


\node[trunknet, 
      below right=0.6cm and 5.0cm of globalTrunk, 
      align=center] (combine) 
{
$t_{\text{pred}} = t_{\text{global}}$\\
$\;\;\;\;\;\;\;\;\;+\,t_{\text{coarse}}$\\
$\;\;\;\;\;\;+\,t_{\text{fine}}$
};

\draw[arrow] (globalTrunk.east) -| (combine.north west);
\draw[arrow] (sumCoarse.east) -- (combine.west);
\draw[arrow] (sumFine.east) -| (combine.south west);

\draw[arrow] (coarseGate.east) -| (fineGate.north);


\node[trunknet, right=1.1cm of combine, minimum width=3.0cm] (podTrunk) {
$[t_\text{pred}, t_\text{POD}]$
};
\draw[arrow] (combine.east) -- (podTrunk.west);


\node[op, right=1.0cm of podTrunk] (dotprod) {$\times$};

\draw[arrow] (branchCNN.east) -| node[above,pos=0.2] {} (dotprod.north);
\draw[arrow] (podTrunk.east) -- node[above,pos=0.5] {} (dotprod.west);


\node[net, right=0.75cm of dotprod, minimum width=1.0cm] (pred) {
$G_{\theta}$
};
\draw[arrow] (dotprod) -- (pred);



\end{tikzpicture}
 }
\caption{\small 
\roneedits{Residual-style} hierarchical architecture used in \METHODNAME-DeepONet.
The input $x$ is embedded and passed through a global backbone to obtain $t_{\mathrm{g}}(x)$.  
A coarse Mixture-of-Experts (MoE) layer predicts $t_{c}(x)$.
A second MoE layer further \roneedits{produces fine-level} contributions $t_{f}(x)$ \roneedits{through nested parent--child partitioning.}  
The final output is the sum of all three components, $t(x) = t_{\mathrm{g}}(x) + t_{c}(x) + t_{f}(x)$.
}
\label{fig:full_hierpou_deeponet_arch}
\end{figure*}

\subsection{Hierarchical POU Deep Operator Networks (\METHODNAME-DeepONets)} \label{sec:methods_hirefpou_deeponet}
\done{
We extend the POU framework to operator learning by introducing \METHODNAME-DeepONets, which incorporate spatially localized expert structure into the trunk representation of a Deep Operator Network (DeepONet)~\citep{lu2021Learning}.
As reviewed in Section~\ref{sec:prelims}, a standard DeepONet represents an operator through an inner product between
(i) a branch network encoding an input function and
(ii) a trunk network encoding evaluation coordinates.
}

The proposed construction mirrors the residual\roneedits{-style additive} POU \roneedits{representation} used in HiRefPOU for PDE approximation, but applies it instead to the coordinate-dependent component of the operator representation.
In our formulation, the branch network remains global and maps sampled observations of the input function $a$ to a coefficient vector
$b(a)\in\mathbb{R}^{p+p_{\mathrm{pod}}}$.   
The trunk representation is augmented with a POU mixture-of-experts defined over the evaluation coordinate $y$.

\done{
Specifically, we define a global trunk feature $t_{\mathrm g}(y)\in\mathbb{R}^{p}$ together with a collection of $N_c$ coarse local trunk experts
$\{t_j(y)\in\mathbb{R}^{p}\}_{j=1}^{N_c}$.
These experts are blended using smooth partition-of-unity weights
$\{\phi_j(y)\}_{j=1}^{N_c}$ satisfying $\sum_{j=1}^{N_c}\phi_j(y)=1$, yielding the POU-enhanced trunk feature
\begin{equation}
  t_{\mathrm{c}}(y)
  =
  t_{\mathrm g}(y)
  +
  \sum_{j=1}^{N_c}\phi_j(y)\, t_j(y),
  \qquad
  t_{\mathrm{c}}(y)\in\mathbb{R}^{p}. \label{eq:pou_deeponet_1l}
\end{equation} 
}

\done{
To incorporate problem-specific low-rank structure, we concatenate $t_{\mathrm{pod}}(y)$ with a fixed POD basis vector
$t_{\mathrm{c}}(y)\in\mathbb{R}^{p_{\mathrm{pod}}}$, forming the final trunk feature
\begin{equation}
  t(y)
  =
  \big[t_{\mathrm{c}}(y),\; t_{\mathrm{pod}}(y)\big]
  \in \mathbb{R}^{p+p_{\mathrm{pod}}}.
\end{equation}
The operator prediction is then obtained via the standard DeepONet contraction
\begin{equation}
  \hat{G}(a)(y) = \langle b(a),\, t(y)\rangle + b,
\end{equation}
where $b\in\mathbb{R}$ is a learned bias.
This formulation preserves the global expressivity of DeepONets while enabling spatially localized specialization through the POU-weighted trunk experts.
}

We further introduce a hierarchical two-level POU structure on the trunk side to \roneedits{organize localized operator representations through nested parent--child spatial partitions.}
At the coarse level, a gating network produces weights
$\{\phi^c_j(y)\}_{j=1}^{N_c}$ satisfying $\sum_{j=1}^{N_c}\phi^c_j(y)=1$, which blend a set of coarse trunk experts.
Each coarse partition is then subdivided by a second gating network that produces child weights
$\{\psi_{j,k}(y)\}_{k=1}^{N_f}$, locally normalized within each parent partition.
The resulting fine-level POU weights are defined by
\[
\phi^f_{j,k}(y)
=
\phi^c_j(y)\, \psi_{j,k}(y),
\qquad
\sum_{k=1}^{N_f}\psi_{j,k}(y)=1,
\]
which guarantees the nested POU property
$\sum_{j=1}^{N_c}\sum_{k=1}^{N_f}\phi^f_{j,k}(y)=1$.
The full hierarchical POU trunk representation is therefore given by
\begin{equation}
t_{\mathrm{pou}}(y)
=
t_{\mathrm g}(y)
+
\sum_{j=1}^{N_c}\phi^c_j(y)\, t^c_j(y)
+
\sum_{j=1}^{N_c} 
\sum_{k=1}^{N_f}
\phi^f_{j,k}(y)\, t^f_{j,k}(y), \label{eq:pou_deeponet_2l}
\end{equation}
where $t^c_j$ and $t^f_{j,k}$ denote coarse- and fine-level trunk experts, respectively.
This hierarchical construction \roneedits{provides nested spatial localization of the operator representation while preserving global consistency through the POU structure.}

\subsection{Partition-of-Unity Fourier Neural Operators (POU-FNO)}
\label{sec:methods_poufno}
In their standard form, FNOs act globally on the entire spatial domain, which can limit their ability to capture strongly localized features.
We extend the partition-of-unity (POU) framework to FNOs by introducing a geometry-aware gating mechanism that blends multiple operator components using smooth, spatially localized weights.
Importantly, the POU-FNO reuses the same geometry-aware gating network introduced in Section~\ref{sec:gating}, with no changes to its functional form or training procedure.
The resulting architecture combines global spectral operator learning with spatially localized POU weighting, enabling spatial adaptivity and localized feature learning.
We emphasize that the POU-FNO is intended as a simple architectural extension illustrating the generality of the POU principle, rather than as a primary hierarchical contribution of this work.

For each spatial coordinate $y_\ell$, a shared gating network produces logits $\ell(y_\ell)\in\mathbb{R}^{N_p}$, which are converted into partition weights as described in Section~\ref{sec:gating}.
The gate depends only on spatial coordinates and is shared across the batch, ensuring geometry-consistent partitions.
Because the POU weights are applied pointwise in physical space, this construction preserves the global FFT structure of the FNO and avoids introducing artificial subdomain boundaries.

A single 1D FNO layer updates hidden features $h(y) \in \mathbb{R}^{C}$ via
\begin{equation}
    \tilde{h}(y)
    = \mathcal{F}^{-1}\!\big(
        \sum_{|k|\leq K} \hat{W}_k \, \mathcal{F}[h]_k
      \big)
      + W_1 h(y),
    \label{eq:fno_update}
\end{equation}
where $\mathcal{F}$ and $\mathcal{F}^{-1}$ denote the forward and inverse discrete Fourier transforms, $\hat{W}_k \in \mathbb{C}^{C\times C}$ are learnable complex-valued spectral coefficients applied to the lowest $K$ modes, and $W_1 \in \mathbb{R}^{C\times C}$ is a learned pointwise linear transformation.
Equation~\eqref{eq:fno_update} is followed by a nonlinearity such as ReLU or GELU, and stacked for $D$ layers to form an FNO block, which may be either shared across partitions or instantiated per partition depending on the architectural variant.

Given $N_p$ expert predictions $\{\hat v_p(y)\}_{p=1}^{N_p}$, the final output is formed by a POU-weighted sum,
\begin{equation}
\hat v(y_\ell)
=
\sum_{p=1}^{N_p}\phi_p(y_\ell)\,\hat v_p(y_\ell),
\label{eq:pou_aggregate}
\end{equation}
which ensures smooth transitions between partitions.
For $N_p=1$, the model reduces exactly to a standard global FNO.

We consider two variations of this POU-FNO construction.
In the first (POU-ExpertFNO), each partition is associated with an independent local FNO, resulting in a fully localized spectral mixture.
In the second (POU-SharedFNO), which we use for most experiments, a single global FNO backbone is shared across all partitions, and locality is introduced through partition-specific expert heads applied to the shared spectral features.
Both variants satisfy the POU aggregation in~\eqref{eq:pou_aggregate} and differ only in the degree to which spectral parameters are shared.
\rthreeedits{In both variants, the partition-specific outputs are evaluated densely and subsequently combined using the POU weights. 
The gates therefore provide spatially varying output weighting rather than conditional selection of FNO experts or heads.}

\subsection{Computational Complexity of HiRefPOU-DeepONets}
We summarize the computational complexity of the HiRefPOU-DeepONet architectures in terms of the branch, trunk, expert, and gating components.
Let \(B\) denote the batch size and \(N_y\) the number of evaluation coordinates per input function.
Let \(p\) denote the learned trunk latent dimension and \(p_{\mathrm{pod}}\) the number of appended POD modes.
We assume that the global trunk network and each local trunk expert have comparable hidden width \(w_t\) and depth \(L_t\), while the gating networks have hidden width \(w_g\) and depth \(L_g\).
We denote the cost of evaluating the branch network on a batch of \(B\) input functions by \(C_{\mathrm{branch}}(B)\).

Because the trunk and gating networks depend only on the evaluation coordinate \(y\), their cost depends on whether all batch elements share the same evaluation grid.
We define
\[
\eta_y =
\begin{cases}
N_y, & \text{if all batch elements share the same evaluation coordinates},\\
B N_y, & \text{if evaluation coordinates differ across batch elements}.
\end{cases}
\]
The final DeepONet contraction is evaluated for every sample-coordinate pair and therefore scales as
\[
\mathcal{O}\!\left(BN_y(p+p_{\mathrm{pod}})\right).
\]
The fixed POD basis does not introduce trainable parameters and contributes only through this final concatenation and contraction.

For HiRefPOU-1L with \(M\) local trunk experts, the trunk representation is
\[
t_{\mathrm{pou}}(y)
=
t_g(y)
+
\sum_{j=1}^{M}\phi_j(y)t_j(y).
\]
Under the current dense implementation, all experts are evaluated at all query coordinates.
The dominant forward-pass cost is therefore
\[
C_{1L}
=
\mathcal{O}\!\left(
C_{\mathrm{branch}}(B)
+
\eta_y(1+M)L_t w_t^2
+
C_{\phi}(M)
+
BN_y(p+p_{\mathrm{pod}})
\right),
\]
where \(C_{\phi}(M)\) is the cost of evaluating the gate with \(M\) outputs.
For a dense MLP gate,
\[
C_{\phi}(M)
=
\mathcal{O}\!\left(
\eta_y(L_gw_g^2+w_gM+Md)
\right),
\]
where the \(Md\) term accounts for center-distance features in \(d\) spatial dimensions.

For HiRefPOU-2L, let \(M_c\) denote the number of coarse experts and \(M_f\) the total number of fine experts across all coarse parents.
The hierarchical trunk representation contains \(1+M_c+M_f\) densely evaluated trunk networks, so the dominant forward-pass cost is
\[
C_{2L}
=
\mathcal{O}\!\left(
C_{\mathrm{branch}}(B)
+
\eta_y(1+M_c+M_f)L_t w_t^2
+
C_{\phi_c}(M_c)
+
C_{\psi}(M_f)
+
BN_y(p+p_{\mathrm{pod}})
\right).
\]
If each coarse parent has \(K_f\) children, then \(M_f=M_cK_f\), recovering the balanced-tree scaling.

Ignoring lower-order input and output projections, the corresponding trainable parameter counts scale as
\[
P_{1L}
=
\mathcal{O}\!\left(
P_{\mathrm{branch}}
+
(1+M)L_t w_t^2
+
L_gw_g^2+w_gM
\right),
\]
and
\[
P_{2L}
=
\mathcal{O}\!\left(
P_{\mathrm{branch}}
+
(1+M_c+M_f)L_t w_t^2
+
2L_gw_g^2
+
w_gM_c
+
w_gM_f
\right).
\]

More generally, for a hierarchy with \(L_h\) POU refinement levels and \(M_\ell\) experts at level \(\ell\), the dense expert-evaluation cost scales linearly with the total number of experts:
\[
C_{\mathrm{trunk}}
=
\mathcal{O}\!\left(
\eta_y
\left(1+\sum_{\ell=1}^{L_h}M_\ell\right)
L_t w_t^2
\right).
\]
Thus, the practical cost of the current implementation grows with the total number of trunk experts across levels.
Since all experts are evaluated densely at all query coordinates, the current implementation does not obtain computational savings from sparse routing.
An ideal routed implementation could reduce the constant factor by evaluating only the active experts associated with each coordinate, but this is not used in the experiments reported here.
\section{Benchmark Problems and Implementation Details} \label{sec:expdetails}

\subsection{Baseline Comparison Methods}
\label{sec:baselines}
To evaluate the proposed \METHODNAME~framework, we primarily benchmark against established neural operator (NO) architectures.
Our focus is on operator learning problems, where the objective is to approximate mappings between infinite-dimensional function spaces rather than individual PDE solutions.
Accordingly, the majority of our numerical experiments compare \METHODNAME~to representative operator-learning baselines under matched training protocols and, where possible, comparable DeepONet-family partition or parameter budgets.

\paragraph{Neural Operator Baselines.}
We compare against the following operator-learning methods:
\begin{itemize}
    \item \textbf{Vanilla DeepONet}: the standard DeepONet architecture with a single global trunk network and no explicit spatial localization.
    \item \textbf{(P+1)-Vanilla DeepONet}: an overparameterized baseline consisting of $P+1$ independent global trunk networks, where $P$ matches the number of partitions used in POU-based models. 
    \rtwoedits{This provides a parameter-enlarged global baseline for assessing whether increased global model capacity alone reproduces the gains obtained from spatial localization. Exact parameter counts and remaining capacity differences are reported in Table~\ref{tab:parameter_counts}}
    \item \textbf{(POD)}: a DeepONet augmented with a fixed Proper Orthogonal Decomposition (POD) basis in the trunk, providing a global low-rank representation of the solution space.
    \item \textbf{(Vanilla, POU)}~\citep{SharmaShankar2025ensembledeeponet}: a DeepONet with a partition-of-unity trunk using fixed (non-adaptive) spatial partitions, isolating the effect of localization without adaptive gating.
    \item \textbf{(POD, POU)}~\citep{SharmaShankar2025ensembledeeponet}: a localized DeepONet combining global POD modes with fixed POU-based spatial experts.
    \item \textbf{(Vanilla, POD, POU)}~\citep{SharmaShankar2025ensembledeeponet}: a hybrid DeepONet that includes global vanilla, global POD, and localized POU expert components, testing whether additional global capacity improves upon POD–POU alone.
    \item \textbf{Geo-Fourier Neural Operator (GeoFNO)}~\citep{li2023fourier}: which extends the Fourier Neural Operator to irregular geometries and performs global spectral convolution in Fourier space. GeoFNO is particularly well-suited for learning solution operators of PDEs posed on complex domains.
    \item \textbf{LocalFNO}~\citep{liu2024neural}: a Fourier neural operator augmented with localized integral and/or differential kernel layers. 
    Unlike standard FNOs, which rely entirely on global spectral convolutions, this architecture introduces explicit local receptive fields. 
    We include LocalFNO as a strong locality-aware spectral baseline for comparison with our POU-based localized operator models.

\end{itemize}
These models provide strong baselines for assessing \rtwoedits{predictive accuracy and measured computational cost in operator-learning settings.}

\paragraph{Motivating Physics-Informed Case Study.}
In addition to operator-learning benchmarks, we include a limited set of physics-informed neural network (PINN) experiments as a motivating case study.
Specifically, we consider a two-dimensional Darcy flow problem with discontinuous coefficients, which highlights known limitations of globally smooth neural approximations when applied to interface-dominated PDEs.
These experiments are not intended as an exhaustive comparison with PINN or domain-decomposition variants such as XPINNs, but rather to illustrate why localized representations are advantageous for problems with sharp material interfaces.
For completeness, we compare against representative PINN-based methods in this setting, including:
the standard PINN formulation~\citep{raissi2018jmlr},
Fourier-feature PINNs~\citep{wang2021eigenvector},
and PirateNets~\citep{wang2024piratenets}.
Additional PINN experiments and implementation details are provided in Appendix~\ref{sec:appendix_pinn_darcy}.

\subsection{Training Setup and Evaluation Metrics}
To assess predictive accuracy, we report the relative $\ell_{2}$ error, which quantifies the discrepancy between the network prediction and the ground truth solution. 
Specifically, the error is defined as
\begin{equation}
    \ell_{2} = 
    \frac{\sqrt{\sum_{i=1}^{N} \left| u(x_i, t_i) - \hat{u}(x_i, t_i) \right|^2}}
         {\sqrt{\sum_{i=1}^{N} \left| u(x_i, t_i) \right|^2}}, 
    \label{eq:l2error}
\end{equation}
where $u$ denotes the exact solution and $\hat{u}$ corresponds to the approximation produced by the neural network. 
This metric provides a relative measure of the error by normalizing with respect to the magnitude of the reference solution.
For the operator-learning benchmarks, we report relative \(\ell_2\) and, in the detailed tables, \(\ell_\infty\) errors together with training and inference times.
For the physics-informed Darcy case study, residual and boundary losses are used during training, while the reported metrics focus on solution and flux errors.

\rtwoedits{
Timing measurements were collected after model compilation and warm-up, with device execution synchronized before recording wall-clock time.
For the full-batch DeepONet experiments, the reported training time corresponds to a single optimizer update evaluated over the entire training set.
For the FNO experiments, the reported training time corresponds to one complete pass through the training set (i.e., one training epoch) and therefore includes multiple optimizer updates.
In both cases, the timed training region includes the forward model evaluation, loss evaluation, reverse-mode differentiation, and parameter update, while excluding periodic metric evaluation and checkpoint writing.
The reported inference time denotes the wall-clock time required to evaluate the complete test set.
}

\rtwoedits{
Unless otherwise specified, benchmark-specific optimization settings are reported in Appendix Table~\ref{tab:training_protocol}.
Each DeepONet-family model is trained for 150,000 gradient updates without early stopping, and the reported results use the final training checkpoint.
Results are reported as the mean and standard deviation across five independent runs.
The fixed train/test datasets listed in Table~\ref{tab:dataset_details} were held constant across all compared methods, and no separate validation split was used.
The test set was not used to select or tune the primary model configurations reported in the main benchmark comparisons.
The additional gate, partition-count, residual-scaling, and child-allocation ablations reported in Appendices~\ref{app:burgers_additional_results} and~\ref{app:cavity_additional_ablations} were conducted post hoc, after the primary model configurations had been fixed, and are included only as retrospective sensitivity analyses.
}

\rtwoedits{
For the Vanilla DeepONet, $(P+1)$-Vanilla DeepONet, POD--POU, and related DeepONet baselines, we followed the architectures used in the corresponding prior publication.
The HiRefPOU-1L and HiRefPOU-2L architectures were selected to approximately match the trainable parameter counts of these competing DeepONet models where possible, although exact matching was not always feasible because of differences in model structure and partition configuration.
GeoFNO and LocalFNO were reimplemented for the present study, with architectures selected to provide parameter counts comparable to the other models where feasible.
The experiments were not designed as an exhaustive equal-budget hyperparameter or architecture search; rather, baseline configurations were inherited from prior work when available, while the proposed HiRefPOU and reimplemented FNO configurations were selected with approximate parameter comparability as a primary consideration.
Accordingly, we interpret the comparisons as applying to the reported model configurations rather than as the outcome of an equal-budget neural architecture search.
}

Training was conducted on an NVIDIA H100 GPU under JAX with single-precision arithmetic.
All code was implemented in Python 3.10.18 using JAX 0.6.2, with DeepMind Haiku 0.0.14 for neural network modules and Optax 0.2.5 for optimization\footnote{
\rtwoedits{
The code, configuration files, and scripts used to reproduce the reported results will be released publicly upon publication.
}}.
\rtwoedits{
Complete architecture, training, dataset, and solver specifications are provided in Appendix~\ref{sec:apx_full_exp_details}.
}

\subsection{\roneedits{Training Objectives}}
\label{sec:training_objectives}
\roneedits{ 
We next summarize the training objectives used for the motivating PINN example in Section~\ref{results:darcyflow_fiveslab} and the operator-learning experiments considered throughout the remainder of this section.
For the Darcy flow experiment, the baseline PINN models are trained using the standard physics-informed objective defined in Eq.~\eqref{eq:full_loss}.
For the HiRefPOU variants, this objective is augmented with the geometric POU regularizers introduced in Section~\ref{sec:geo_regularizers}:
\begin{equation}
    L =
    L_{\mathrm{PINN}}
    +
    L_{\mathrm{cov}}
    +
    L_{\mathrm{cont}},
    \label{eq:hirefpou_pinn_loss}
\end{equation}
where $L_{\mathrm{cov}}$ and $L_{\mathrm{cont}}$ are defined in Eqs.~\eqref{eq:cov_loss} and~\eqref{eq:cont_loss}, respectively.
The weighting coefficients $\lambda_{\mathrm{cov}}$ and $\lambda_{\mathrm{cont}}$ are included directly in the definitions of these regularization terms.
}

\roneedits{
The operator-learning models are trained using the data-driven objective introduced in Eq~\eqref{eq:ol_loss}.
The HiRefPOU variants are similarly combined with the geometric POU regularization described in Section~\ref{sec:geo_regularizers} yielding the complete training objective:
\begin{equation}
    L =
    L_{\mathrm{op}}
    +
    L_{\mathrm{cov}}
    +
    L_{\mathrm{cont}}.
    \label{eq:hirefpou_ol_loss}
\end{equation}
}
\roneedits{
The hierarchical POU construction does not introduce separate loss terms for the individual refinement levels. 
In particular, the coarse- and fine-level contributions are not trained against independently defined residual targets. Instead, the loss is evaluated on the final model prediction, and the global network, local experts, and gating parameters
at all hierarchy levels are optimized jointly through the corresponding objective above. 
Thus, the term residual-style refers to the additive composition of the architecture described in Section~\ref{sec:methods}, rather than to a separate hierarchical residual loss.
No staged training, parameter freezing, detached residual targets, or
stop-gradient operations are used. All global-network, local-expert, and
gating parameters are optimized simultaneously through the complete objective,
and every hierarchy level contributes to the final prediction and receives
gradients from the corresponding prediction loss throughout training.
}

\subsection{\METHODNAME~Model Variants}
For each problem, we compare the following variants of the proposed hierarchical POU architecture, 
\rtwoedits{which differ in the number and organization of learned POU refinement levels. 
In each variant, the gate parameters are optimized jointly with the remaining model parameters. 
In the operator-learning examples, the resulting coordinate-dependent partitions are shared across samples. 
}
\begin{itemize}
\item \textbf{\METHODNAME-1L}: \roneedits{a single-level POU model 
    ($u(x) = u_{\mathrm{g}}(x) + u_{0}(x)$) in which the coarse gating 
    network is co-trained jointly with the global network and local experts.}
\item \textbf{\METHODNAME-2L}: the full hierarchical POU model 
    \roneedits{($u(x) = u_{\mathrm{g}}(x) + u_{0}(x) + u_{1}(x)$) in which
    the coarse and fine gates are co-trained along all other model parameters 
    throughout PDE/operator training.}
\item \textbf{HiRefPOU-3L}: 
    \roneedits{a three-level POU model 
    (($u(x) = u_{\mathrm{g}}(x) + u_{0}(x) + u_{1}(x) + u_2(x)$) that tests whether 
    an additional level of nested spatial partitioning provides additional accuracy beyond the 2L hierarchy.
    }
\end{itemize}

Further, to demonstrate the generality of the proposed POU framework, we additionally embed the POU mechanism within neural operator architectures as described in Section~\ref{sec:methods_poufno}.
These extensions are not hierarchical, but instead aim to illustrate how POU-based localization can improve operator learning in FNO models.
Specifically, we consider the following POU-FNO realizations:
\begin{itemize}
    \item \textbf{POU-SharedFNO}: \rthreeedits{a FNO with a shared global spectral backbone and partition-specific expert heads.}
    \item \textbf{POU-ExpertFNO}: a more expressive but costly variant in which each partition is associated with an independent local FNO, yielding a fully localized spectral mixture.
\end{itemize}

\section{Numerical Results} \label{sec:experiments1}

\begin{table}[t]
\centering
\renewcommand{\arraystretch}{1.15}
\setlength{\tabcolsep}{6pt}
\caption{\small
Solution and flux errors for the 2D rough Darcy flow benchmark using
HiRefPOU-1L/-2L, APINN, and PirateNet.
\generaledits{
For each column, the best value is shown in \textbf{bold}, and the second-best value is underlined.
Rows corresponding to the proposed adaptive POU-based models are highlighted.
}
}
\vspace{0.05cm}

\footnotesize
\begin{adjustbox}{width=\linewidth}
\begin{tabular}{l c c c c}
\toprule
\textbf{Model}
& $\mathbf{(N_c, N_f)}$
& \textbf{Rel.} $\mathbf{\ell_2}$ $\mathbf{u}$ \textbf{Error}
& \textbf{Rel.} $\mathbf{\ell_2}$ \textbf{flux-}$\mathbf{x}$ \textbf{Error}
& \textbf{Abs. flux-}$\mathbf{y}$ \textbf{Error}
\\
\midrule

PirateNet
& $(-,-)$
& $\num{5.1e-05} \pm \num{2.9e-05}$
& $\num{2.8e-05} \pm \num{9.0e-06}$
& $\num{8.0e-02} \pm \num{5.5e-02}$
\\

APINN
& $(2,-)$
& $\num{3.7e-05} \pm \num{1.8e-05}$
& $\num{2.7e-05} \pm \num{1.4e-05}$
& $\num{4.1e-02} \pm \num{1.9e-02}$
\\

\midrule

\cellcolor{gray!10} HiRefPOU-1L
& \cellcolor{gray!10} $(2,-)$
& \cellcolor{gray!10} \best{$\num{1.5e-07} \pm \num{4.6e-08}$}
& \cellcolor{gray!10} \underline{$\num{1.6e-07} \pm \num{2.9e-08}$}
& \cellcolor{gray!10} \best{$\num{1.3e-04} \pm \num{1.2e-05}$}
\\

\cellcolor{gray!10} HiRefPOU-2L
& \cellcolor{gray!10} $(2,3)$
& \cellcolor{gray!10} \underline{$\num{1.8e-07} \pm \num{3.8e-07}$}
& \cellcolor{gray!10} \best{$\num{1.2e-07} \pm \num{1.9e-07}$}
& \cellcolor{gray!10} \underline{$\num{1.4e-04} \pm \num{2.2e-02}$}
\\

\bottomrule
\end{tabular}
\end{adjustbox}

\label{tab:df_rough_overview}
\end{table}

\subsection{Motivating PINN Example: Darcy Flow with Discontinuous Coefficients}
\label{results:darcyflow_fiveslab}

We begin with a two-dimensional Darcy flow problem with discontinuous coefficients, included as a motivating example to illustrate the limitations of globally smooth PINNs on interface-dominated PDEs.
Darcy flow serves as a canonical elliptic benchmark in heterogeneous media, where accurate flux recovery across material interfaces is essential.
The Darcy experiment is included as a motivating physics-informed case study, while the remaining experiments evaluate the proposed localization mechanisms in the operator-learning setting.

We consider the steady Darcy problem
\begin{align}
  -\nabla\!\cdot\!\big(\mu(x)\,\nabla \phi(x)\big) = f(x),
  \qquad x \in \Omega=[0,1]^2,
\end{align}
with mixed Dirichlet--Neumann boundary conditions.
The permeability field $\mu(x)$ is piecewise constant and discontinuous, inducing sharp material interfaces.

\paragraph{Five-strip benchmark.}
We employ the standard five-strip manufactured-solution test
\citep{trask2017fivestrip,nakshatrala2006stabilized},
in which the domain is partitioned into horizontal layers with distinct permeability values.
The exact solution is chosen such that the pressure and normal flux remain continuous across interfaces, while tangential flux components exhibit discontinuities.

\generaledits{\textit{Results.}} Figure~\ref{fig:df_rough_preds} and Table~\ref{tab:df_rough_overview} compare solution and flux predictions for representative PINN-based methods and the proposed \METHODNAME.
\rtwoedits{
Both HiRefPOU variants substantially reduce the solution and flux errors relative to PirateNets and APINNs, demonstrating} 
the benefit of localized, partition-based representations for PDEs with discontinuous coefficients.
\rtwoedits{However, the additional hierarchical level added in the HiRefPOU-2L variant does not provide a consistent advantage over the -1L variant across the reported metrics. }
Further technical details, interface conditions, and additional PINN results are provided in Appendix~\ref{sec:appendix_pinn_darcy}.


\begin{table}[!htpb]
\centering
\renewcommand{\arraystretch}{1.15}
\setlength{\tabcolsep}{6pt}

\caption{
\small
Testing relative $l_2$ errors for the \textbf{Burgers}, \textbf{cavity flow}, \textbf{Full-Field Burgers}, \textbf{Kuramoto--Sivashinsky}, \textbf{2D reaction--diffusion}, and \textbf{3D reaction--diffusion} problems.
RD stands for reaction--diffusion.
For each problem, the best result across all models is shown in \textbf{bold}, and the second-best result is underlined.
Rows corresponding to the proposed adaptive PoU-based models are highlighted.
}
\vspace{0.05cm}

\small
\begin{adjustbox}{width=0.80\linewidth}
\begin{tabular}{clcc}
\toprule

& \textbf{Model} & \multicolumn{1}{c}{\textbf{Burgers}} & \multicolumn{1}{c}{\textbf{Cavity flow}} \\ 
\midrule
\multirow{8}{*}{\rotatebox{90}{\textbf{DeepONets}}}
& Vanilla  
& $\num{2.24e-01} \pm \num{3.0e-03}$
& $\num{4.86e-02} \pm \num{6.2e-03}$ \\
& (P+1)-Vanilla 
& $\num{2.25e-01} \pm \num{6.5e-03}$
& $\num{2.05e-02} \pm \num{3.3e-03}$ \\
& (POD, PoU) 
& $\num{1.95e-01} \pm \num{1.4e-03}$
& $\num{2.28e-03} \pm \num{1.5e-04}$ \\
& (Vanilla, PoU)  
& $\num{2.43e-01} \pm \num{1.1e-02}$
& $\num{1.73e-02} \pm \num{4.7e-03}$ \\
& (Vanilla, POD, PoU) 
& $\num{1.92e-01} \pm \num{2.4e-03}$
& $\num{2.16e-03} \pm \num{1.7e-04}$ \\
& \cellcolor{gray!10} \METHODNAME-1L (POD) 
& \cellcolor{gray!10} $\num{1.75e-01} \pm \num{1.1e-03}$
& \cellcolor{gray!10} \best{$\num{1.55e-03} \pm \num{1.3e-04}$} \\
& \cellcolor{gray!10} \METHODNAME-2L (POD) 
& \cellcolor{gray!10} $\num{1.74e-01} \pm \num{1.2e-03}$
& \cellcolor{gray!10} \underline{$\num{1.69e-03} \pm \num{1.2e-04}$} \\
& \cellcolor{gray!10} \METHODNAME-3L (POD) 
& \cellcolor{gray!10} $\num{1.74e-01} \pm \num{2.1e-03}$
& \cellcolor{gray!10} $\num{1.79e-03} \pm \num{9.6e-05}$ \\
\midrule
\multirow{4}{*}{\rotatebox{90}{\textbf{FNOs}}}
& GeoFNO               
& $\num{3.56e-01} \pm \num{3.0e-02}$
& $\num{5.56e-02} \pm \num{1.2e-02}$ \\
& LocalFNO               
& \best{$\num{1.16e-01} \pm \num{4.4e-02}$}
& $\num{1.38e-02} \pm \num{1.2e-03}$ \\
& \cellcolor{gray!10} POU-SharedFNO
& \cellcolor{gray!10} $\num{1.74e-01} \pm \num{3.7e-02}$ 
& \cellcolor{gray!10} $\num{4.15e-02} \pm \num{1.5e-02}$ \\
& \cellcolor{gray!10} POU-ExpertFNO
& \cellcolor{gray!10} \underline{$\num{1.24e-01} \pm \num{1.3e-02}$} 
& \cellcolor{gray!10} $\num{8.07e-03} \pm \num{4.4e-05}$ \\
\midrule
\midrule
& \textbf{Model} & \multicolumn{1}{c}{\textbf{Full-Field Burgers}} & \multicolumn{1}{c}{\textbf{Kuramoto-Sivashinsky}} \\
\midrule
\multirow{8}{*}{\rotatebox{90}{\textbf{DeepONets}}}
& Vanilla
& $\num{1.02e-01} \pm \num{2.0e-03}$
& $\num{1.70} \pm \num{1.3e-01}$ \\
& (P+1)-Vanilla
& $\num{1.27e-01} \pm \num{2.8e-03}$
& $\num{1.68} \pm \num{1.3e-01}$ \\
& (POD, PoU)
& $\num{8.84e-02} \pm \num{2.3e-03}$
& $\num{1.59} \pm \num{2.2e-02}$ \\
& (Vanilla, PoU)
& $\num{1.01e-01} \pm \num{1.6e-03}$
& $\num{1.58} \pm \num{1.2e-02}$ \\
& (Vanilla, POD, PoU)
& $\num{9.06e-02} \pm \num{1.0e-03}$
& $\num{1.50} \pm \num{1.4e-02}$ \\
& \cellcolor{gray!10} \METHODNAME-1L (POD)
& \cellcolor{gray!10} $\num{8.76e-02} \pm \num{1.5e-03}$
& \cellcolor{gray!10} $\num{1.25} \pm \num{6.3e-02}$ \\
& \cellcolor{gray!10} \METHODNAME-2L (POD)
& \cellcolor{gray!10} $\num{8.62e-02} \pm \num{2.2e-03}$
& \cellcolor{gray!10} $\num{1.33} \pm \num{2.3e-02}$ \\
& \cellcolor{gray!10} \METHODNAME-3L (POD) 
& \cellcolor{gray!10} $\num{8.68e-02} \pm \num{1.1e-03}$
& \cellcolor{gray!10} $\num{1.39} \pm \num{2.2e-02}$ \\
\midrule
\multirow{4}{*}{\rotatebox{90}{\textbf{FNOs}}}
& GeoFNO
& $\num{4.21e-02} \pm \num{3.5e-03}$
& $\num{9.03e-01} \pm \num{4.5e-03}$ \\
& LocalFNO
& $\num{3.17e-02} \pm \num{4.3e-03}$  
& $\num{7.85e-01} \pm \num{9.9e-03}$ \\ 
& \cellcolor{gray!10} POU-SharedFNO
& \cellcolor{gray!10} \best{$\num{2.49e-02} \pm \num{3.4e-03}$}
& \cellcolor{gray!10} \best{$\num{7.62e-01} \pm \num{1.5e-02}$} \\ 
& \cellcolor{gray!10} POU-ExpertFNO
& \cellcolor{gray!10} \underline{$\num{2.73e-02} \pm \num{2.3e-03}$}
& \cellcolor{gray!10} \underline{$\num{7.80e-01} \pm \num{1.9e-02}$} \\ 
\midrule
\midrule
& \textbf{Model} & \multicolumn{1}{c}{\textbf{2D RD}} & \multicolumn{1}{c}{\textbf{3D RD}} \\ 
\midrule 
\multirow{8}{*}{\rotatebox{90}{\textbf{DeepONets}}}
& Vanilla  
& $\num{1.61e-03} \pm \num{4.4e-04}$
& $\num{1.20e-03} \pm \num{3.6e-04}$ \\
& (P+1)-Vanilla 
& $\num{5.77e-04} \pm \num{2.8e-04}$
& $\num{1.30e-01} \pm \num{1.4e-01}$ \\
& (POD, PoU) 
& $\num{4.74e-04} \pm \num{9.8e-05}$
& $\num{4.46e-04} \pm \num{2.0e-04}$ \\
& (Vanilla, PoU) 
& $\num{7.81e-04} \pm \num{2.1e-04}$
& $\num{5.57e-04} \pm \num{3.1e-04}$ \\
& (Vanilla, POD, PoU) 
& $\num{7.74e-04} \pm \num{1.4e-04}$
& $\num{4.22e-04} \pm \num{1.6e-04}$ \\
& \cellcolor{gray!10} \METHODNAME-1L (POD) 
& \cellcolor{gray!10} $\num{4.48e-04} \pm \num{2.4e-04}$
& \cellcolor{gray!10} $\num{3.68e-04} \pm \num{2.2e-04}$ \\
& \cellcolor{gray!10} \METHODNAME-2L (POD) 
& \cellcolor{gray!10} \underline{$\num{3.68e-04} \pm \num{1.3e-04}$}
& \cellcolor{gray!10} \underline{$\num{3.12e-04} \pm \num{1.6e-04}$} \\
& \cellcolor{gray!10} \METHODNAME-3L (POD)
& \cellcolor{gray!10} \best{$\num{3.56e-04} \pm \num{6.2e-05}$}
& \cellcolor{gray!10} \best{$\num{2.55e-04} \pm \num{3.0e-05}$} \\
\midrule 
\multirow{4}{*}{\rotatebox{90}{\textbf{FNOs}}}
& GeoFNO               
& $\num{4.84e-2} \pm \num{1.1e-1}$
& $\num{8.56e-1} \pm \num{5.2e-1}$ \\
& LocalFNO               
& $\num{6.31e-3} \pm \num{4.9e-6}$
& $\num{9.30e-3} \pm \num{2.3e-5}$ \\
& \cellcolor{gray!10} POU-SharedFNO
& \cellcolor{gray!10} $\num{7.47e-03} \pm \num{2.1e-04}$
& \cellcolor{gray!10} $\num{1.02e-0} \pm \num{1.2e-0}$ \\
& \cellcolor{gray!10} POU-ExpertFNO
& \cellcolor{gray!10} $\num{4.70e-03} \pm \num{1.1e-04}$
& \cellcolor{gray!10} $\num{9.98e-01} \pm \num{4.1e-02}$ \\
\bottomrule
\end{tabular} 
\end{adjustbox}
\label{tab:main_ol_results}
\end{table}

\subsection{2D Lid-Driven Cavity Flow}
\textit{Problem Setup.}
We next consider the two-dimensional lid-driven cavity flow problem, a classical benchmark for incompressible fluid dynamics in a confined domain.
The flow is governed by the incompressible Navier--Stokes equations
\begin{align}
    \frac{\partial \mathbf{u}}{\partial t}
    + (\mathbf{u} \cdot \nabla)\mathbf{u}
    &= -\nabla p + \nu \Delta \mathbf{u},
    \quad \nabla \cdot \mathbf{u} = 0,
    \quad \mathbf{x} \in \Omega, \; t \in T, \label{eq:ns}\\[4pt]
    \mathbf{u} &= \mathbf{u}_b, \label{eq:bc}
\end{align}
where $\mathbf{u} = (u,v)$ is the velocity field, $p$ is the pressure, and $\nu$ denotes the kinematic viscosity.
The Dirichlet boundary condition $\mathbf{u}=\mathbf{u}_b$ enforces the prescribed lid motion.

\begin{figure}[!htpb]
    \centering
    \includegraphics[width=0.60\textwidth]{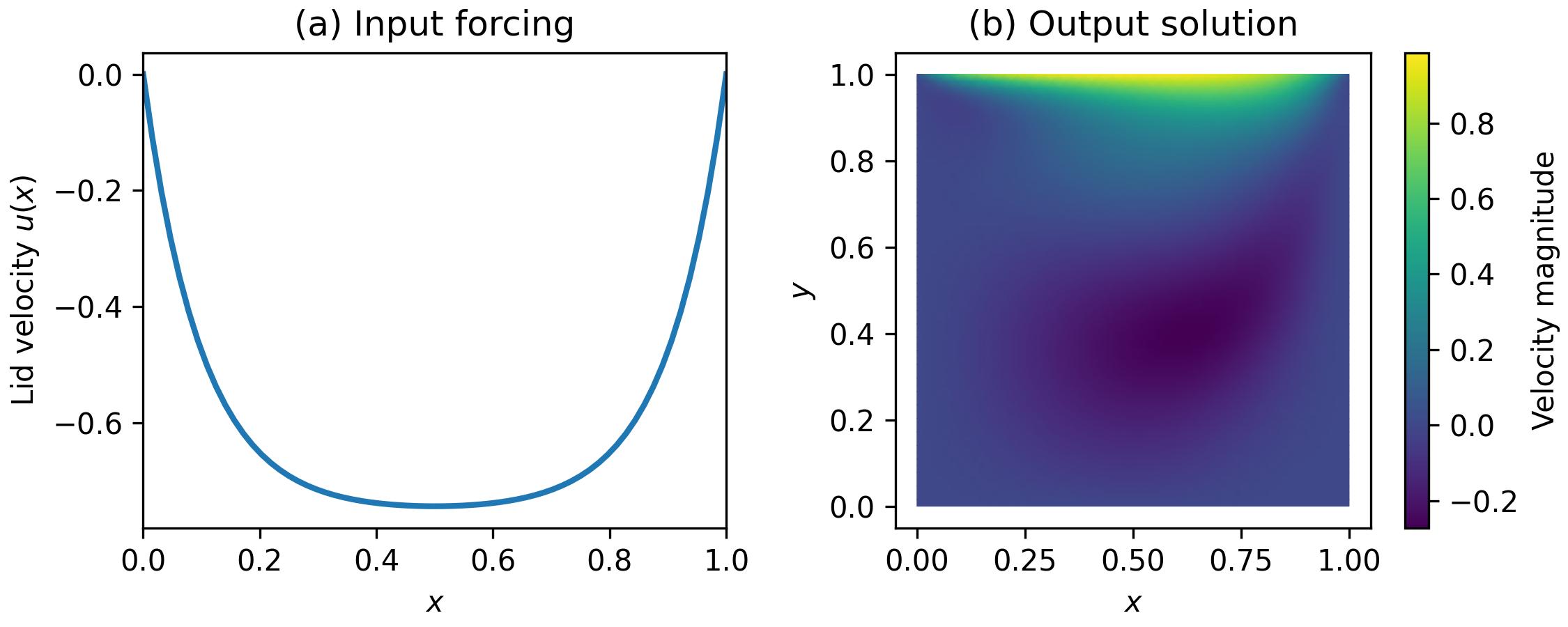}
\caption{
\small
Example input–output pair from the lid-driven cavity flow experiment.
(a) One-dimensional lid velocity profile $u(x)$ prescribed along the top boundary of the cavity and used as the model input.
(b) Corresponding two-dimensional cavity flow solution for a representative test instance.
}\label{fig:cavity_examples}
\end{figure}

We learn an operator mapping from the one-dimensional lid velocity profile prescribed along the top boundary to the resulting steady-state two-dimensional velocity field:
\[
\mathcal{G} : u_b(x) \;\mapsto\; \mathbf{u}(x,y).
\]
An example input--output pair is shown in Figure~\ref{fig:cavity_examples}.
We consider the steady-state regime on the unit square domain $\Omega=[0,1]^2$, following the dataset specification of~\citep{lu2022comprehensive}.
The lid velocity profile is defined as
\begin{align}
    u_b(x) = U\!\left(1 -
    \frac{\cosh\!\big(r(x - \tfrac{1}{2})\big)}
         {\cosh\!\big(\tfrac{r}{2}\big)}\right),
    \quad v_b = 0, \label{eq:lid}
\end{align}
with $r=10$.
All remaining boundaries are stationary.
Steady-state solutions are generated using a lattice Boltzmann method (LBM) solver.
The dataset consists of $100$ training and $10$ test input--output pairs, following~\citep{lu2022comprehensive,SharmaShankar2025ensembledeeponet}.

\textit{Results.}
Quantitative results for the lid-driven cavity flow problem are reported in Tables~\ref{tab:main_ol_results} and~\ref{tab:cavity_results}.
Among DeepONet-based models, the HiRefPOU variants substantially reduce both relative $\ell_2$ and $\ell_\infty$ errors compared with the vanilla and static-partition baselines.
HiRefPOU-1L achieves the lowest relative $\ell_2$ and $\ell_\infty$ errors among the DeepONet-family models.
\rtwoedits{Among the HiRefPOU variants, however, increasing hierarchy depth does not improve the aggregate error metrics. HiRefPOU-1L achieves the lowest mean relative $\ell_2$ and $\ell_\infty$ errors, while the 2L model remains comparable and the 3L model is less accurate. The differences between 1L and 2L are also modest relative to the reported variability across seeds. Thus, the cavity experiment supports adaptive localization, but does not provide evidence that a deeper hierarchy is necessary for this operator.}
The FNO-based results are mixed: LocalFNO and POU-ExpertFNO are competitive in relative $\ell_2$, but the HiRefPOU-DeepONet variants achieve the best errors.

\begin{figure}[!htpb]
    \centering
    \includegraphics[width=0.75\textwidth]{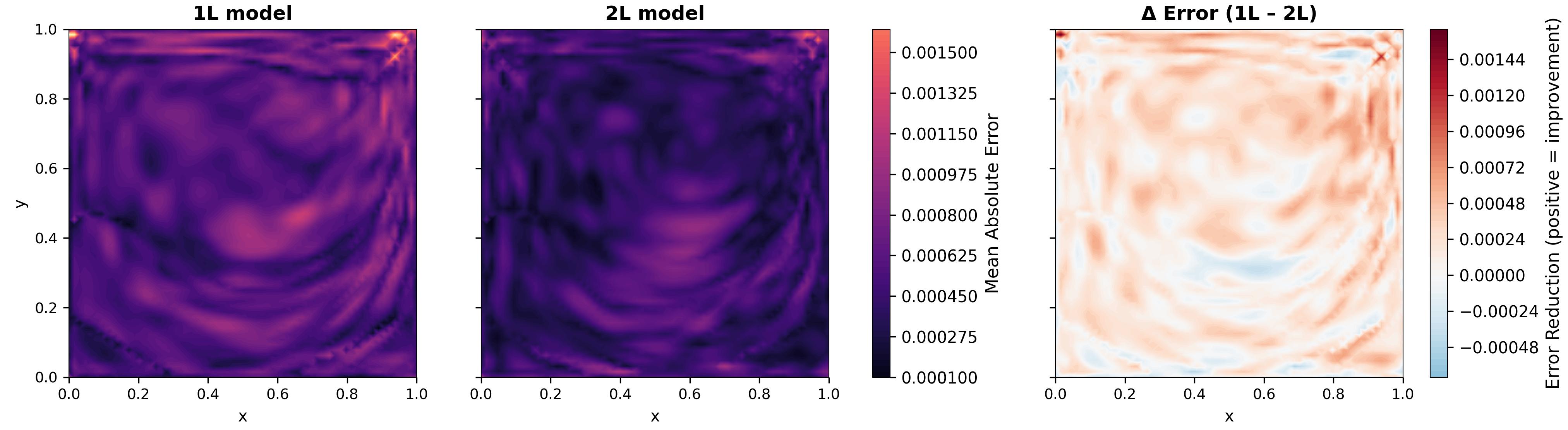}
\caption{\small
Comparison of single-level (1L) and hierarchical two-level (2L) \METHODNAME~DeepONet models for the cavity flow problem.
Both models employ six total partitions initialized at identical center locations.
The left and center panels show the mean absolute error (MAE) fields for the 1L ($e_{1L}$) and 2L ($e_{2L}$) architectures, respectively.
The rightmost panel displays the local error difference $\delta e = e_{1L} - e_{2L}$, where positive values indicate regions in which the hierarchical model improves upon the single-level model.
}
\label{fig:cavity_error_fields}
\end{figure}

\begin{figure}[!htpb]
    \centering
    \includegraphics[width=0.80\textwidth]{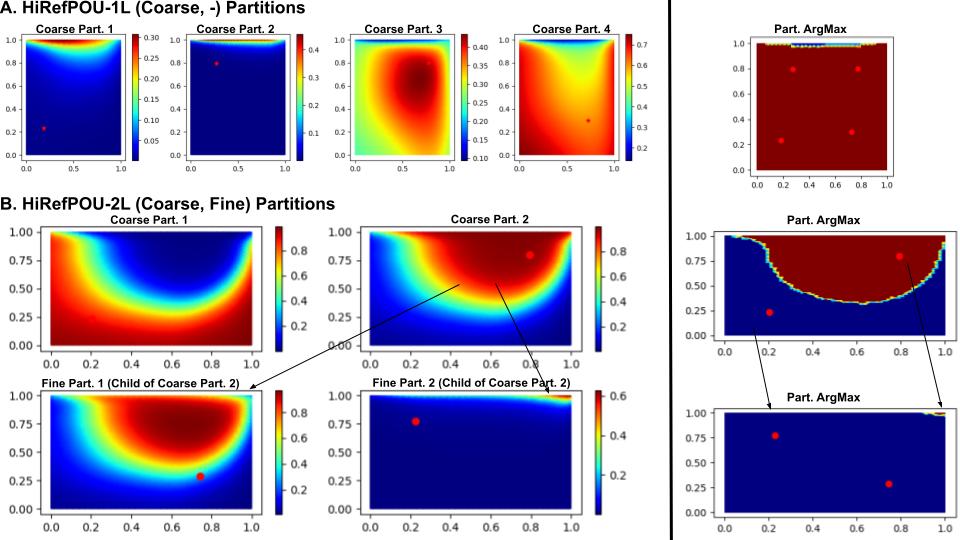}
    \vspace{-0.05cm}
\caption{\small 
Learned partition-of-unity (POU) gating functions for the cavity flow problem using the single-level (1L) and hierarchical two-level (2L) HiRefPOU-DeepONet models.
(A) Single-level (1L) model: spatially varying coarse gating functions associated with individual experts, together with the induced hard partition obtained via the pointwise ArgMax. \rthreeedits{The ArgMax maps are shown only to visualize the induced partition structure; model predictions use the continuous POU weights and dense evaluation of experts.}
(B) Hierarchical two-level (2L) model: learned coarse-level gating functions (top row) and their corresponding fine-level child partitions (bottom row). 
Rightmost panels show the induced hard partitions at each level.
Colors indicate gating weight magnitude, with warmer colors corresponding to 
\rthreeedits{larger POU gating weights.}
The hierarchical model \roneedits{subdivides the learned coarse partitions into localized fine-level child regions while preserving the nested parent--child structure.}
}\label{fig:cavity_results}
\end{figure}

Figure~\ref{fig:cavity_error_fields} \rtwoedits{provides a complementary spatial comparison between the 1L and 2L models. 
Although their aggregate errors are similar, the error-difference field shows that the 2L model reduces error in some localized regions while increasing it in others. 
Figure~\ref{fig:cavity_results} likewise shows that the hierarchical construction subdivides the learned coarse regions into finer child partitions. 
These results illustrate the additional spatial organization provided by the hierarchy, but do not imply that this organization produces uniformly lower prediction error.}

Finally, Table~\ref{tab:cavity_results} reports training and inference times.
The adaptive HiRefPOU models incur additional cost relative to the simplest vanilla DeepONet variants, reflecting the overhead of gating and expert evaluation.
However, this overhead is moderate within the DeepONet family and accompanies substantial improvements in both relative $\ell_2$ and $\ell_\infty$ error.
Thus, the cavity results support an accuracy--cost tradeoff: adaptive partitioning improves predictive accuracy and interpretability, while requiring additional computation relative to the simplest baselines.

\begin{table}[!htpb]
\centering
\renewcommand{\arraystretch}{1.15}
\setlength{\tabcolsep}{6pt}
\caption{\footnotesize
Testing $\ell_\infty$ errors, training times, and inference times for the Cavity Flow problem.
\rtwoedits{For the error and inference-time columns, the best value is shown in bold and
the second-best value is underlined. 
Training times are reported for
computational context but are not ranked across model families because the
reported unit differs between full-batch DeepONet and minibatch FNO models.}
Ties are bolded when reported values are equal to the displayed precision.
Rows corresponding to proposed adaptive POU-based models are highlighted.
}
\vspace{0.05cm}

\footnotesize
\begin{adjustbox}{width=\linewidth}
\begin{tabular}{c l c c c}
\toprule
 & Model & $\ell_\infty$ Error & \rtwoedits{Mean Training Time$^\dagger$ (s)} & Inference Time (s) \\
\midrule

\multirow{8}{*}{\rotatebox{90}{\textbf{DeepONets}}}
& Vanilla
 &  $\num{1.67e-01} \pm \num{2.7e-02}$
 &  $\num{1.01e-03} \pm \num{5.8e-04}$
 &  $\num{9.54e-04} \pm \num{6.0e-04}$
\\
& (P+1)-Vanilla
 &  $\num{7.02e-02} \pm \num{2.7e-02}$
 &  $\num{1.01e-03} \pm \num{7.4e-05}$
 &  \best{$\num{1.46e-04} \pm \num{1.3e-05}$}
\\
& (Vanilla, POU)
 &  $\num{6.33e-02} \pm \num{2.6e-02}$
 &  $\num{5.44e-03} \pm \num{3.8e-03}$
 &  $\num{5.26e-03} \pm \num{3.8e-03}$
\\
& (POD, POU)
 &  $\num{4.81e-03} \pm \num{3.7e-04}$
 &  $\num{3.33e-03} \pm \num{5.3e-05}$
 &  $\num{3.20e-03} \pm \num{7.1e-06}$
\\
& (Vanilla, POD, POU)
 &  $\num{6.95e-03} \pm \num{2.0e-03}$
 &  $\num{4.42e-03} \pm \num{6.6e-05}$
 &  $\num{4.29e-03} \pm \num{1.1e-04}$
\\
& \cellcolor{gray!10} HiRefPOU-1L (POD)
 & \cellcolor{gray!10} \best{$\num{3.03e-03} \pm \num{5.2e-04}$}
 & \cellcolor{gray!10} $\num{3.60e-03} \pm \num{7.4e-04}$
 & \cellcolor{gray!10} $\num{4.74e-04} \pm \num{1.1e-04}$
\\
& \cellcolor{gray!10} HiRefPOU-2L (POD)
 & \cellcolor{gray!10} \underline{$\num{3.61e-03} \pm \num{4.6e-04}$}
 & \cellcolor{gray!10} $\num{4.05e-03} \pm \num{1.1e-03}$
 & \cellcolor{gray!10} $\num{4.68e-04} \pm \num{8.5e-05}$
\\
& \cellcolor{gray!10} \METHODNAME-3L (POD)
 &  \cellcolor{gray!10} $\num{4.85e-03} \pm \num{5.6e-04}$
 &  \cellcolor{gray!10} $\num{4.97e-03} \pm \num{7.7e-07}$
 &  \cellcolor{gray!10} \underline{$\num{4.57e-04} \pm \num{1.9e-05}$}
\\

\midrule

\multirow{4}{*}{\rotatebox{90}{\textbf{FNOs}}}
& GeoFNO
 &  $\num{9.63e-02} \pm \num{2.0e-02}$
 &  $\num{1.92e-01} \pm \num{6.3e-02}$
 &  $\num{1.78e-01} \pm \num{1.4e-01}$
\\

& LocalFNO
 & $\num{1.98e-2} \pm \num{1.1e-3}$
 & $\num{4.01e-2} \pm \num{1.2e-3}$
 & $\num{2.37e+1} \pm \num{6.12e-1}$
\\

& \cellcolor{gray!10} POU-SharedFNO (GeoFNO)
 & \cellcolor{gray!10} $\num{9.71e-02} \pm \num{4.1e-02}$
 & \cellcolor{gray!10} $\num{5.42e-02} \pm \num{9.0e-04}$
 & \cellcolor{gray!10} $\num{1.88e-02} \pm \num{2.2e-04}$
\\

& \cellcolor{gray!10} POU-ExpertFNO (GeoFNO)
 & \cellcolor{gray!10} $\num{1.24e-02} \pm \num{8.0e-04}$
 & \cellcolor{gray!10} $\num{1.05e-01} \pm \num{2.8e-05}$
 & \cellcolor{gray!10} $\num{2.88e-02} \pm \num{1.1e-04}$
\\

\bottomrule
\end{tabular}
\end{adjustbox}
\footnotesize{
\rtwoedits{
$^\dagger$For full-batch DeepONet-family models, training time denotes one optimizer
update over the complete training set. For minibatch FNO-family models, it denotes
one complete training epoch.
}}
\label{tab:cavity_results}
\end{table}

\subsection{2D Reaction-Diffusion Problem}
\textit{Problem Setup.}
We next consider a two-dimensional reaction--diffusion equation describing the spatiotemporal concentration $c(y,t)$ of a chemical species undergoing binding and unbinding reactions.
The governing partial differential equation is
\begin{align}
    \frac{\partial c}{\partial t}
    &= k_{\mathrm{on}} (R - c)\, c_{\mathrm{amb}}
    - k_{\mathrm{off}}\, c
    + \nu \Delta c,
    \qquad y \in \Omega,\; t \in T,
    \label{eq:2d_reaction_diffusion}
\end{align}
with homogeneous Neumann boundary conditions
\begin{align}
    \nu \frac{\partial c}{\partial n} = 0,
    \qquad y \in \partial \Omega.
\end{align}
Here, $k_{\mathrm{on}}$ and $k_{\mathrm{off}}$ denote reaction on- and off-rates, $\nu = 0.1$ is the diffusion coefficient, and $R=2$ controls the available reactant concentration.
The ambient concentration
\[
c_{\mathrm{amb}}(y,t)
=
\bigl[1 + \cos(2\pi y_1)\cos(2\pi y_2)\bigr] e^{-\pi t}
\]
introduces spatial and temporal variability into the reaction term. 

\begin{figure}[!htpb]
    \centering
    \includegraphics[width=0.60\textwidth]{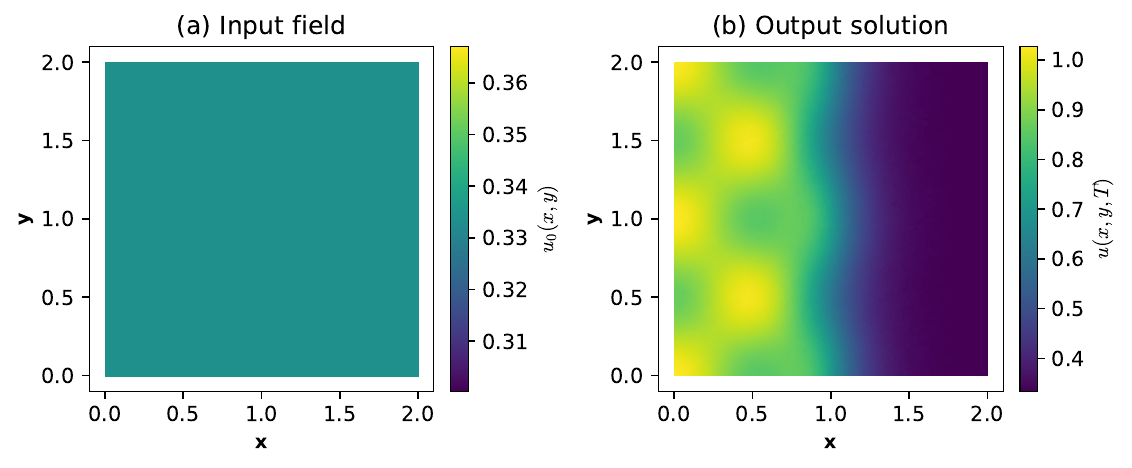}
\caption{
\small
Example input--output pair for the two-dimensional reaction-diffusion problem.
(a) Spatial input field $c(y,0)$ provided as input to the operator-learning model.
(b) Corresponding solution field $c(y,T)$ at the final time.
}
\label{fig:diffrec_examples}
\end{figure}

Both reaction coefficients are discontinuous in the horizontal direction $y_1$:
\begin{align}
    k_{\mathrm{on}}(y_1) &=
    \begin{cases}
        2, & y_1 \le 1, \\
        0, & \text{otherwise},
    \end{cases}
    &
    k_{\mathrm{off}}(y_1) &=
    \begin{cases}
        0.2, & y_1 \le 1, \\
        0, & \text{otherwise}.
    \end{cases}
    \label{eq:2d_reaction_coeffs}
\end{align}
This discontinuity induces a sharp transition across the interface $y_1=1$, producing localized gradients that are challenging for global operator-learning models.
Our objective is to learn the operator
\[
\mathcal{G} : c(y,0) \mapsto c(y,T),
\qquad T = 0.5,
\]
mapping the initial concentration field to the solution at the final time.
The problem is defined on $\Omega=[0,2]^2$ with initial condition $c(y,0)\sim\mathcal{U}(0,1)$.
An example input--output pair is shown in Figure~\ref{fig:diffrec_examples}.
Reference solutions are taken from~\cite{SharmaShankar2025ensembledeeponet}, computed using an RBF--FD discretization~\citep{SFJCP2018,SHANKAR2021highorder}.
The dataset consists of $2207$ spatial collocation points, with $1000$ training and $200$ test input--output pairs.

\textit{Results.}
\rtwoedits{We next compare the 1L, 2L, and 3L HiRefPOU DeepONet models on the 2D reaction--diffusion benchmark.}
Quantitative results are reported in Tables~\ref{tab:main_ol_results} and~\ref{tab:2ddiffrec_results}.
\rtwoedits{
All three HiRefPOU variants improve upon the vanilla and static-partition DeepONet baselines in both relative $\ell_2$ and $\ell_\infty$ error, again demonstrating the benefit of learned spatial localization in the presence of the fixed coefficient discontinuity at $y_1=1$.
}
The 1L and 2L models use the same total number of partitions and comparable parameter counts, so their performance differences isolate the effect of organizing the same partition budget hierarchically.
\rtwoedits{The 3L model introduces an additional nested partition level and a larger finest-level partition count, so its comparison with the shallower variants reflects both hierarchy depth and increased model capacity.}

\rtwoedits{
Within the HiRefPOU family, the mean errors exhibit a favorable trend with additional hierarchy: HiRefPOU-3L has the lowest mean relative $\ell_2$ and $\ell_\infty$  errors, followed by 2L and then 1L. 
However, these differences are modest relative to the reported variability across random seeds, and the corresponding mean ($\pm$) standard-deviation ranges overlap. 
We therefore interpret these results as evidence that hierarchical organization can be beneficial for this interface-dominated problem, rather than as evidence that increasing hierarchy depth is uniformly superior. 
}

\begin{figure}[!htpb]
    \centering
    \includegraphics[width=0.75\textwidth]{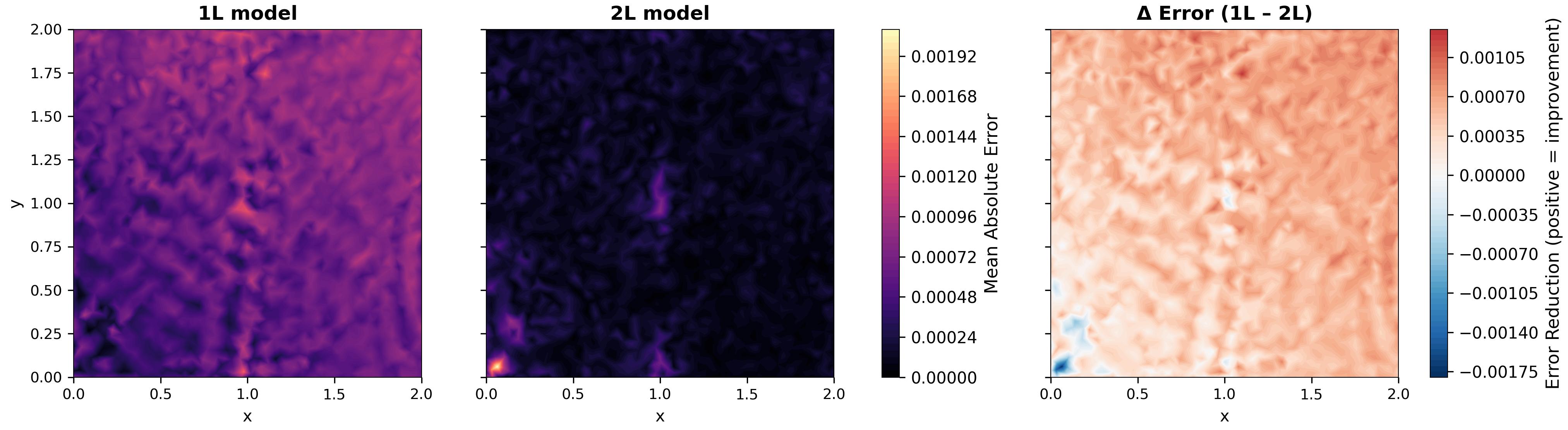} 
\caption{\small
Comparison of single-level (1L) and hierarchical two-level (2L) \METHODNAME DeepONet models.
Both models employ six total partitions with identical initial center locations.
The left and center panels show the mean absolute error (MAE) fields for the 1L ($e_{1L}$) and 2L ($e_{2L}$) architectures, respectively.
The rightmost panel displays the local error difference $\delta e = e_{1L} - e_{2L}$, where positive values indicate regions in which the hierarchical model improves upon the single-level model.
}\label{fig:diffrec_error_fields}
\end{figure}

\begin{figure}[!htpb]
    \centering
    \includegraphics[width=0.8\textwidth]{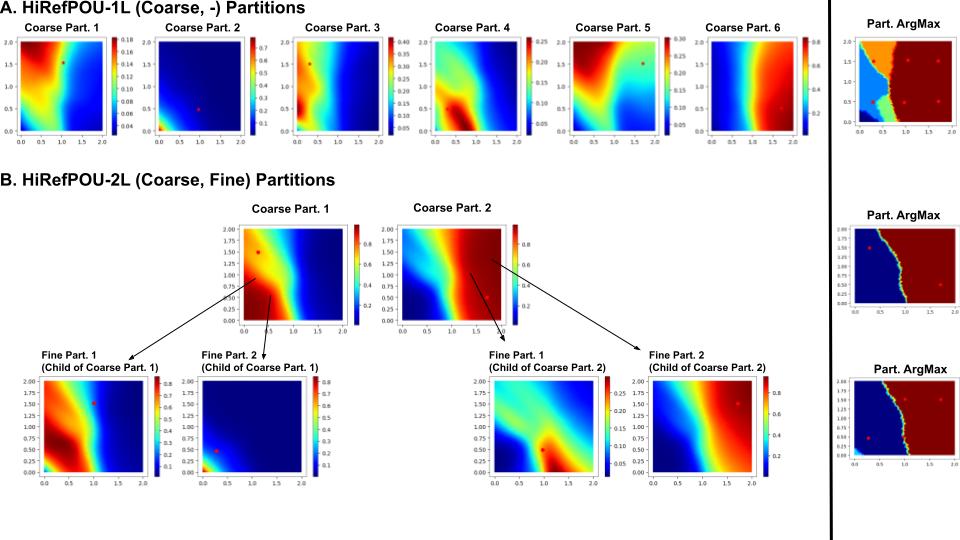}
\caption{\small 
    Comparison of single-level (1L) and hierarchical two-level (2L) \METHODNAME~ DeepONet learned partitions.
    Each panel shows the learned POU gating functions corresponding to different experts within the network.
    The top row illustrates the partitions produced by the single-level (1L) model,
    while the middle and bottom rows correspond to the hierarchical two-level (2L) model, where coarse partitions are further subdivided into locally refined subregions.
    The hierarchical model \roneedits{learns localized second-level child partitions near the coefficient discontinuity while preserving coherent coarse-level partitions across the domain.}
    }\label{fig:diffrec_partitions}
\end{figure}

Figure~\ref{fig:diffrec_error_fields} \rtwoedits{provides additional spatial context for the 1L--2L comparison. 
The error-difference field indicates regions near the coefficient interface in which the hierarchical model reduces the prediction error, together with regions where the single-level model remains competitive. }
Figure~\ref{fig:diffrec_partitions} shows the \rtwoedits{learned partitions, where the second level subdivides the coarse regions into more localized child partitions (shown in the bottom row of the figure).
Taken together, the quantitative and spatial results suggest that the hierarchy can provide useful additional localization around persistent heterogeneous structure, while the magnitude of the resulting accuracy improvement remains problem- and initialization-dependent.
}

\begin{table}[!htpb]
\centering
\renewcommand{\arraystretch}{1.15}
\setlength{\tabcolsep}{6pt}
\caption{\footnotesize
Testing $\ell_\infty$ errors, training times, and inference times for the 2D Reaction-Diffusion problem.
\rtwoedits{For the error and inference-time columns, the best value is shown in bold and
the second-best value is underlined. 
Training times are reported for
computational context but are not ranked across model families because the
reported unit differs between full-batch DeepONet and minibatch FNO models.}
Rows corresponding to proposed adaptive PoU-based models are highlighted.
}
\vspace{0.05cm}

\footnotesize
\begin{adjustbox}{width=\linewidth}
\begin{tabular}{c l c c c}
\toprule
 & Model & $\ell_\infty$ Error & \rtwoedits{Mean Training Time$^\dagger$ (s)} & Inference Time (s) \\
\midrule

\multirow{8}{*}{\rotatebox{90}{\textbf{DeepONets}}}
& Vanilla
 &  $\num{5.65e-03} \pm \num{1.3e-03}$
 &  $\num{3.66e-04} \pm \num{4.3e-06}$
 &  \best{$\num{8.74e-05} \pm \num{4.7e-06}$}
\\
& (P+1)-Vanilla
 &  $\num{1.81e-03} \pm \num{4.8e-04}$
 &  $\num{9.88e-04} \pm \num{3.2e-05}$
 &  \underline{$\num{1.41e-04} \pm \num{6.1e-06}$}
\\
& (Vanilla, POU)
 &  $\num{2.45e-03} \pm \num{5.5e-04}$
 &  $\num{2.84e-03} \pm \num{8.7e-04}$
 &  $\num{2.52e-03} \pm \num{1.2e-03}$
\\
& (POD, POU)
 &  $\num{2.15e-03} \pm \num{6.4e-04}$
 &  $\num{6.05e-03} \pm \num{2.7e-03}$
 &  $\num{5.75e-03} \pm \num{3.1e-03}$
\\
& (Vanilla, POD, POU)
 &  $\num{2.46e-03} \pm \num{2.2e-04}$
 &  $\num{7.38e-03} \pm \num{3.5e-04}$
 &  $\num{7.34e-03} \pm \num{3.2e-05}$
\\

& HiRefPOU-1L (POD)
 & \cellcolor{gray!10} $\num{1.22e-03} \pm \num{3.1e-04}$
 & \cellcolor{gray!10} $\num{3.42e-03} \pm \num{9.6e-04}$
 & \cellcolor{gray!10} $\num{3.76e-04} \pm \num{3.9e-05}$
\\
& HiRefPOU-2L (POD)
 & \cellcolor{gray!10} \underline{$\num{1.20e-03} \pm \num{2.5e-04}$}
 & \cellcolor{gray!10} $\num{4.72e-03} \pm \num{1.0e-03}$
 & \cellcolor{gray!10} $\num{5.62e-04} \pm \num{1.3e-04}$
\\
& HiRefPOU-3L (POD)
 & \cellcolor{gray!10} \best{$\num{1.14e-03} \pm \num{2.2e-04}$}
 & \cellcolor{gray!10} $\num{4.65e-03} \pm \num{4.0e-05}$
 & \cellcolor{gray!10} $\num{5.73e-04} \pm \num{2.9e-05}$
\\

\midrule

\multirow{4}{*}{\rotatebox{90}{\textbf{FNOs}}}
& GeoFNO
 &  $\num{1.57e-01} \pm \num{3.3e-01}$
 &  $\num{5.30e-01} \pm \num{1.3e-01}$
 &  $\num{5.44e-02} \pm \num{5.3e-02}$
\\

& LocalFNO
 & $\num{2.27e-2} \pm \num{6.8e-6}$
 & $\num{6.94e-2} \pm \num{1.2e-3}$
 & $\num{7.16e-2} \pm \num{2.7e-3}$
\\

& \cellcolor{gray!10} PoU-SharedFNO (GeoFNO)
& \cellcolor{gray!10} $\num{2.50e-02} \pm \num{3.5e-03}$
& \cellcolor{gray!10} $\num{5.35e-02} \pm \num{9.4e-05}$
& \cellcolor{gray!10} $\num{1.89e-02} \pm \num{9.2e-05}$ 
\\

& \cellcolor{gray!10} PoU-ExpertFNO (GeoFNO)
& \cellcolor{gray!10} $\num{7.00e-02} \pm \num{1.5e-03}$
& \cellcolor{gray!10} $\num{7.51e-01} \pm \num{1.2e-04}$
& \cellcolor{gray!10} $\num{4.19e-02} \pm \num{1.1e-05}$ 
\\

\bottomrule
\end{tabular}
\end{adjustbox}
\footnotesize{
\rtwoedits{
$^\dagger$For full-batch DeepONet-family models, training time denotes one optimizer
update over the complete training set. For minibatch FNO-family models, it denotes
one complete training epoch.
}}
\label{tab:2ddiffrec_results}
\end{table}

\subsection{3D Reaction-Diffusion with Variable Diffusion Coefficient}
\begin{figure}[!htpb]
    \centering
    \includegraphics[width=0.70\textwidth]{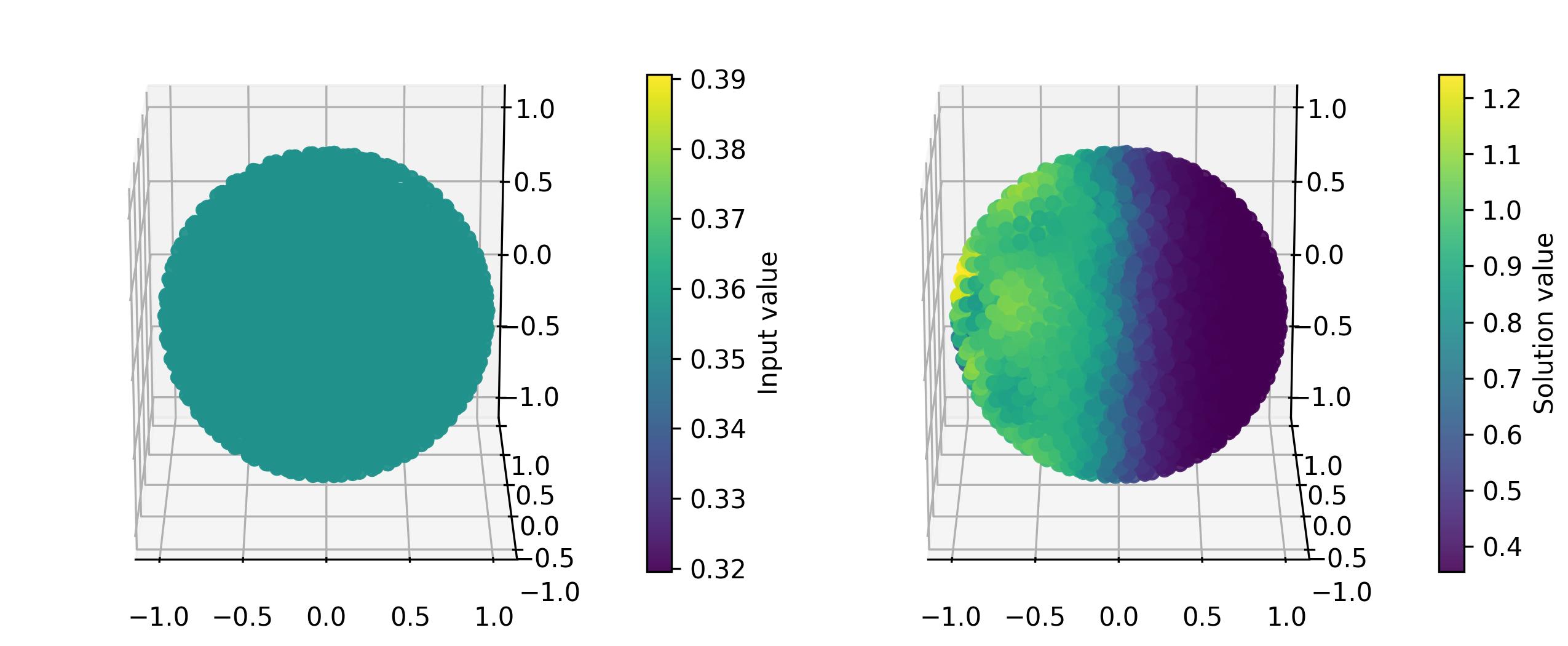}
\caption{
\small
Example input--output pair for the three-dimensional reaction-diffusion problem.
(Left) Spatial input field provided as input to the operator-learning model.
(Right) Corresponding solution field at the final time.
}\label{fig:3ddiffrec_examples}
\end{figure}

\textit{Problem Setup.}
We next consider a three-dimensional reaction-diffusion system with a spatially varying diffusion coefficient, posed on a curved domain.
Let $c(y,t)$ denote the concentration field with $y \in \mathbb{R}^3$.
The governing equation is
\begin{align}
    \frac{\partial c}{\partial t}
    = k_{\mathrm{on}} (R - c) c_{\mathrm{amb}}
    - k_{\mathrm{off}} c
    + \nabla \cdot \big( K(y)\nabla c \big),
    \qquad y \in \Omega,\; t \in T,
    \label{eq:3d_reaction_diffusion}
\end{align}
with homogeneous Neumann boundary conditions
\begin{align}
    K(y)\frac{\partial c}{\partial n} = 0,
    \qquad y \in \partial \Omega.
\end{align}
The computational domain is the unit ball
\[
\Omega = \{y \in \mathbb{R}^3 : \|y\|_2 \le 1\},
\]
with boundary $\partial\Omega=\mathbb{S}^2$, and the time interval is $T=[0,0.5]$.
An example input--output pair is shown in Figure~\ref{fig:3ddiffrec_examples}.
The reaction coefficients $k_{\mathrm{on}}$ and $k_{\mathrm{off}}$ follow the same configuration as in the two-dimensional case, taking nonzero values in the half-space $y_1 \le 0$ and vanishing elsewhere.
The ambient source term is defined as
\[
c_{\mathrm{amb}}(y,t)
=
\bigl[1 + \cos(2\pi y_1)\cos(2\pi y_2)\sin(2\pi y_3)\bigr] e^{-\pi t}.
\]
The spatially varying diffusion coefficient $K(y)$ is designed to introduce sharp gradients,
\begin{align}
    K(y_1) = B
    + \frac{C}{\tanh(A)}
    \Big[
    (A - 3)\tanh(8y_1 - 5)
    - (A - 15)\tanh(8y_1 + 5)
    + A\tanh(A)
    \Big],
    \label{eq:variable_diffusion}
\end{align}
with $A=9$, $B=0.0215$, and $C=0.005$.
This configuration yields strong spatial heterogeneity in the diffusion operator.
Our objective is to learn the solution operator
\[
\mathcal{G}: c(y,0) \mapsto c(y,0.5).
\]
Reference solutions are taken from~\cite{SharmaShankar2025ensembledeeponet} and computed using a high-order RBF-FD solver~\citep{SFJCP2018,SHANKAR2021highorder}.
The dataset consists of $1000$ training and $200$ test samples, with each function evaluated at $4325$ spatial collocation points.

\textit{Results.}
Quantitative results for the three-dimensional reaction--diffusion problem are reported in Tables~\ref{tab:main_ol_results} and~\ref{tab:3ddiffrec_results}.
\rtwoedits{The HiRefPOU variants achieve lower errors than the vanilla and static-partition DeepONet baselines, indicating that learned spatial localization remains effective in this three-dimensional heterogeneous setting. 
The mean results also favor the hierarchical variants: HiRefPOU-3L achieves the lowest relative $\ell_2$ error, while HiRefPOU-2L achieves the lowest mean $\ell_\infty$ error.}

\rtwoedits{These results provide the strongest indication among the tested problems that additional hierarchical organization can be useful when the operator contains persistent spatial heterogeneity at multiple spatial scales. Nevertheless, the differences among the 1L, 2L, and 3L variants remain comparable to the reported run-to-run variability, particularly for the 1L and 2L models. 
We therefore do not interpret the ordering of the mean errors as evidence of monotonic improvement with hierarchy depth. Rather, the results suggest that additional levels can provide useful representational flexibility for this problem, with the preferred depth depending on the error metric and training realization.}

\begin{figure}[!htpb]
    \centering
    \includegraphics[width=0.8\textwidth]{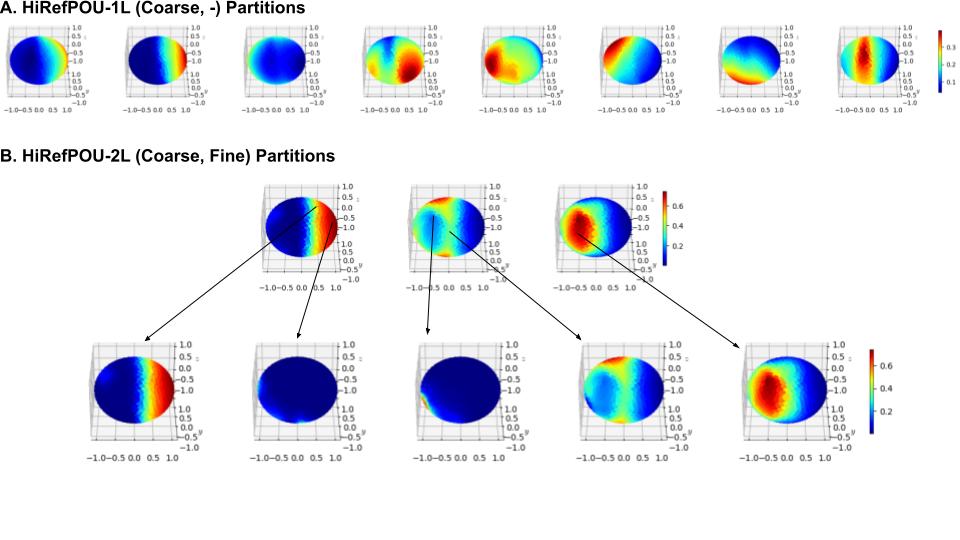}
    \caption{\small 
    Comparison of single-level (1L) and hierarchical two-level (2L) \METHODNAME~ DeepONet learned partitions.
    Each panel shows the learned POU gating functions corresponding to different experts within the network.
    The top row illustrates the partitions produced by the single-level (1L) model,
    while the middle and bottom rows correspond to the hierarchical two-level (2L) model, where coarse partitions are further subdivided into locally refined subregions.
    The hierarchical model \roneedits{learns localized second-level child partitions aligned with regions of strong diffusion variation while preserving coherent coarse-level partitions across the ball domain.}
    }\label{fig:3ddiffrec_partitions}
\end{figure}

Figure~\ref{fig:3ddiffrec_partitions} \rtwoedits{further illustrates the geometric effect of the hierarchy. 
The single-level model produces smooth learned partitions over the ball, whereas the hierarchical model subdivides these regions through nested child partitions. 
The learned subdivision is concentrated in regions associated with strong spatial variation in the diffusion coefficient, providing qualitative evidence that the additional partition levels can organize localized model capacity around persistent heterogeneous structure.}

\begin{table}[!htpb]
\centering
\renewcommand{\arraystretch}{1.15}
\setlength{\tabcolsep}{6pt}
\caption{\footnotesize
Testing $\ell_\infty$ errors, training times, and inference times for the 3D Reaction-Diffusion problem.
\rtwoedits{For the error and inference-time columns, the best value is shown in bold and
the second-best value is underlined. 
Training times are reported for
computational context but are not ranked across model families because the
reported unit differs between full-batch DeepONet and minibatch FNO models.}
Rows corresponding to proposed adaptive PoU-based models are highlighted.
}
\vspace{0.05cm}

\footnotesize
\begin{adjustbox}{width=\linewidth}
\begin{tabular}{c l c c c}
\toprule
 & Model & $\ell_\infty$ Error & \rtwoedits{Mean Training Time$^\dagger$ (s)} & Inference Time (s) \\
\midrule

\multirow{8}{*}{\rotatebox{90}{\textbf{DeepONets}}}
& Vanilla
 &  $\num{3.81e-03} \pm \num{1.5e-03}$
 &  $\num{4.61e-04} \pm \num{1.0e-04}$
 &  $\num{4.50e-04} \pm \num{1.2e-04}$
\\
& (P+1)-Vanilla
 &  $\num{9.77e-02} \pm \num{1.1e-01}$
 &  $\num{1.69e-03} \pm \num{5.7e-04}$
 &  $\num{1.64e-03} \pm \num{5.9e-04}$
\\
& (Vanilla, PoU)
 &  $\num{1.13e-03} \pm \num{5.1e-04}$
 &  $\num{1.40e-02} \pm \num{6.7e-03}$
 &  $\num{1.39e-02} \pm \num{6.7e-03}$
\\
& (POD, PoU)
 &  $\num{1.83e-03} \pm \num{5.2e-04}$
 &  $\num{1.11e-02} \pm \num{9.7e-05}$
 &  $\num{1.09e-02} \pm \num{4.8e-05}$
\\
& (Vanilla, POD, PoU)
 &  $\num{1.36e-03} \pm \num{5.0e-04}$
 &  $\num{1.13e-02} \pm \num{1.9e-04}$
 &  $\num{1.11e-02} \pm \num{1.1e-04}$
\\
& \cellcolor{gray!10} HiRefPOU-1L (POD)
 & \cellcolor{gray!10} $\num{6.87e-04} \pm \num{4.1e-04}$
 & \cellcolor{gray!10} $\num{2.63e-03} \pm \num{1.0e-04}$
 & \cellcolor{gray!10} \best{$\num{3.54e-04} \pm \num{2.7e-05}$}
\\
& \cellcolor{gray!10} HiRefPOU-2L (POD)
 & \cellcolor{gray!10} \best{$\num{4.75e-04} \pm \num{1.9e-04}$}
 & \cellcolor{gray!10} $\num{2.86e-03} \pm \num{1.3e-04}$
 & \cellcolor{gray!10} \underline{$\num{4.38e-04} \pm \num{3.3e-05}$}
\\

& \cellcolor{gray!10} \METHODNAME-3L (POD)
 & \cellcolor{gray!10} \underline{$\num{4.95e-04} \pm \num{5.2e-05}$}
 & \cellcolor{gray!10} $\num{7.13e-03} \pm \num{5.7e-05}$
 & \cellcolor{gray!10} $\num{6.16e-04} \pm \num{8.7e-06}$
\\
\midrule

\multirow{4}{*}{\rotatebox{90}{\textbf{FNOs}}}
& GeoFNO
 & $\num{1.80e+0} \pm \num{7.3e-01}$
 & $\num{5.61e+0} \pm \num{8.9e-03}$
 & $\num{2.52e-01} \pm \num{2.5e-02}$
\\

& LocalFNO
 & $\num{3.20e-2} \pm \num{1.1e-3}$
 & $\num{1.51e-1} \pm \num{8.4e-3}$
 & $\num{8.36e-2} \pm \num{2.5e-3}$
\\

& \cellcolor{gray!10} PoU-SharedFNO (GeoFNO)
& \cellcolor{gray!10} $\num{1.09e-0} \pm \num{5.6e-02}$
& \cellcolor{gray!10} $\num{7.35e-0} \pm \num{9.4e-02}$
& \cellcolor{gray!10} $\num{4.89e-01} \pm \num{9.2e-02}$ 
\\

& \cellcolor{gray!10} PoU-ExpertFNO (GeoFNO)
& \cellcolor{gray!10} $\num{7.09e-1} \pm \num{1.6e-02}$
& \cellcolor{gray!10} $\num{1.05e+01} \pm \num{5.2e-01}$
& \cellcolor{gray!10} $\num{1.09e-0} \pm \num{1.7e-01}$ 
\\

\bottomrule
\end{tabular}
\end{adjustbox}
\footnotesize{
\rtwoedits{
$^\dagger$For full-batch DeepONet-family models, training time denotes one optimizer
update over the complete training set. For minibatch FNO-family models, it denotes
one complete training epoch.
}}
\label{tab:3ddiffrec_results}
\end{table}

\subsection{1D Burgers' Equation} \label{sec:burgers_standard}
\done{
We next consider two one-dimensional Burgers benchmarks. 
The first is the standard periodic Gaussian-random-field benchmark used widely in operator-learning studies. 
The second is a modified multiscale periodic benchmark, constructed to contain more explicit mixed coarse- and fine-scale spatial structure. 
The purpose of the second benchmark is to more directly probe the regimes in which hierarchical adaptive partitioning is beneficial.
}
        
\begin{figure}[!htpb]
    \centering
    \includegraphics[width=0.70\textwidth]{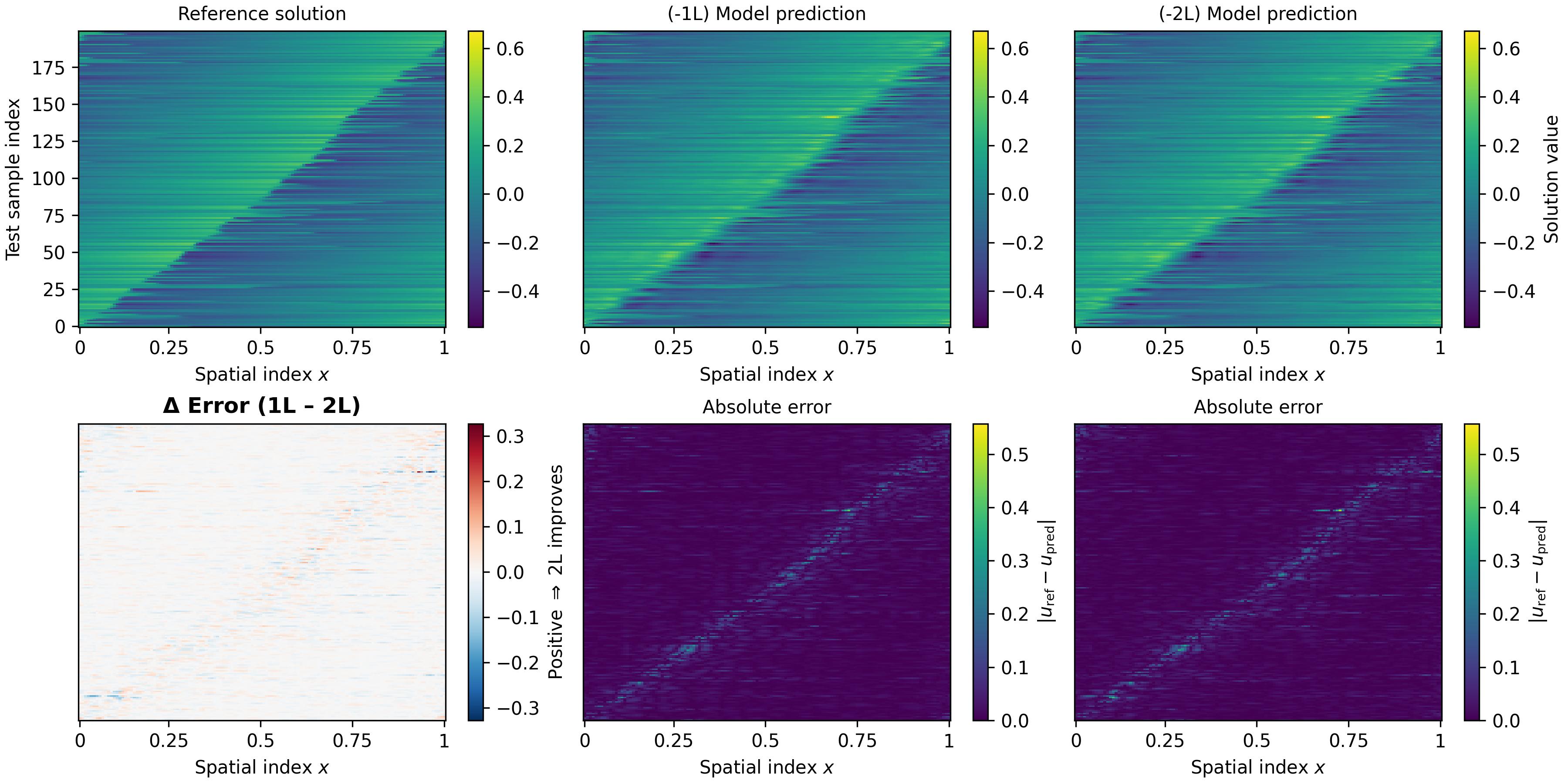}
    \vspace{-0.05cm}
\caption{\footnotesize \textit{(1D Burgers Problem.)}
Operator learning predictions at final time $t=1$.
(Top row) Reference solution $u(x,1)$, HiRefPOU-1L prediction, and HiRefPOU-2L prediction, evaluated over the full test set.
Each heatmap displays solution values as a function of spatial coordinate $x \in [0,1]$ (horizontal axis) and test sample index (vertical axis).
(Bottom row) Point-wise absolute error fields $\lvert u_{\mathrm{ref}} - u_{\mathrm{pred}} \rvert$ for the 1L and 2L models, along with their difference.
Positive values indicate locations where the 2L model achieves lower error than the 1L model.
}
\label{fig:burgers_results}
\end{figure}

\textit{Problem Setup.}
We next consider the one-dimensional Burgers' equation:
\begin{align}
    \frac{\partial u}{\partial t} + u \frac{\partial u}{\partial x} = \nu \frac{\partial^2 u}{\partial x^2}, \quad x \in (0,1), \ t \in (0,1]
\end{align}
with periodic boundary conditions and viscosity $\nu=0.001$.  
Examples of final-time solutions are shown in Figure~\ref{fig:burgers_results}.

\done{
In this problem, we learn the operator mapping from the initial condition $u(x,0) = u_0(x)$ to the solution $u(x,t)$ at $t=1$,
\begin{align}
    \mathcal{G} : u_0(x) \mapsto u(x,1).
\end{align}
Similar to~\citep{lu2022comprehensive}, we generate each dataset such that each initial condition is generated according to a random distribution:
\begin{align}
    \mu = \mathcal{N}(0, 625(-\Delta + 25I)^{-2}).
\end{align}
The initial condition $u_0$ is modeled as a sample from a Gaussian random field with mean zero and covariance operator
\begin{equation}
    C = \sigma^2 \left( -\Delta + \tau^2 I \right)^{-\gamma},
\end{equation}
where $\sigma^2 = 25$, $\tau = 5$, and $\gamma = 2.5$. 
The field is defined on the periodic domain $(0,1)$, and realizations are generated spectrally by drawing independent standard normal coefficients $\xi_k, \eta_k \sim \mathcal{N}(0,1)$ and scaling by the eigenvalues 
of the covariance operator:
\[
    \lambda_k = \sqrt{2}\,|\sigma|\,
    \left( (2\pi k)^2 + \tau^2 \right)^{-\gamma/2}.
\]
}

\done{
The resulting Fourier coefficients are combined using the \texttt{Chebfun} package in MATLAB to construct smooth periodic realizations of the form
  \[
  u_0(x)=\sum_{k=1}^{K}\alpha_k\cos(2\pi kx)+\beta_k\sin(2\pi kx),
  \]
  where $K$ denotes the number of retained Fourier modes and
  \[
  \alpha_k=\lambda_k\xi_k,\qquad
  \beta_k=\lambda_k\eta_k,\qquad
  \xi_k,\eta_k\overset{\text{i.i.d.}}{\sim}\mathcal N(0,1).
  \]
}

\done{
To generate the dataset, each realization of $u_0$ is evolved forward in time according to the Burgers' equation using a spectral solver implemented in \texttt{Chebfun}. 
This method employs adaptive Fourier expansions in space and high-order time-stepping schemes to ensure spectral accuracy. 
We discretize the spatial domain uniformly with $2^{13} = 8192$ collocation points and evolve the system over $500$ uniform time steps up to $t=1$.
For each realization, we record both the initial condition $u_0(x)$ and the final-time solution $u(x,1)$. 
This is a common benchmark problem for operator learning methods~\citep{lu2022comprehensive}.
}

\begin{figure}[!htpb]
    \centering
    \includegraphics[width=0.5\textwidth]{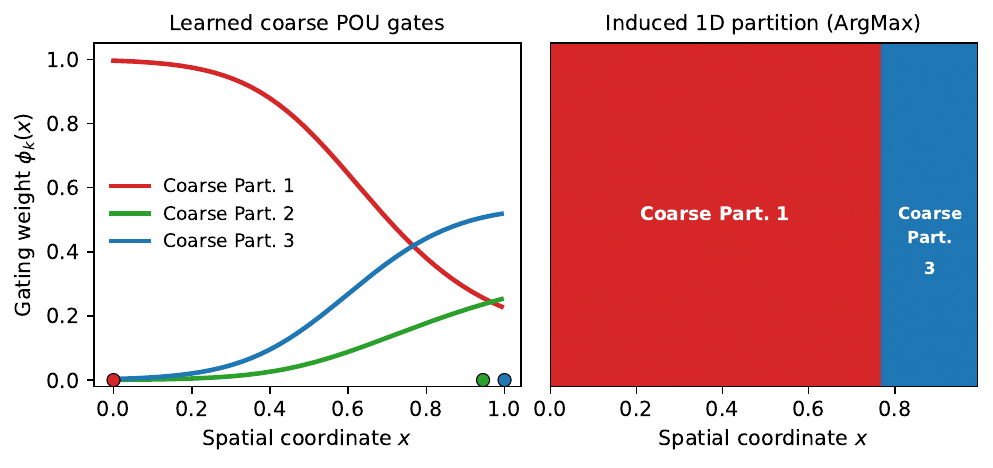}
    \includegraphics[width=0.5\textwidth]{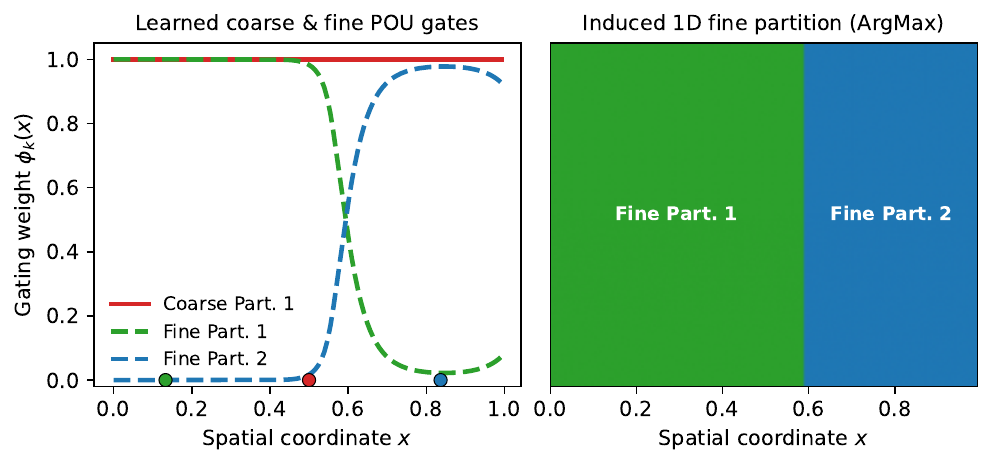}
    \vspace{-0.05cm}
\caption{\footnotesize \textit{(1D Burgers Problem.)}
Learned partition-of-unity (POU) gates for HiRefPOU-DeepONet models.
(Top left) Continuous coarse-level gating functions $\{\phi_j(x)\}_{j=1}^3$ learned by the HiRefPOU-1L model.
Colored markers indicate the learned center locations associated with each partition.
(Top right) Induced hard partition obtained via $\arg\max_j \phi_j(x)$, illustrating the discrete domain decomposition implied by the learned soft POU representation.
(Bottom left) Continuous coarse- and fine-level gating functions $\{\phi^c_j(x)\}_{j=1}^1$ and $\{\phi^f_k(x)\}_{k=1}^2$ learned by the HiRefPOU-2L model.
(Bottom right) Induced hard fine-level partition obtained via $\arg\max_k \phi^f_k(x)$.
}
\label{fig:burgers_learned_parts}
\end{figure}

\begin{table}[!htpb]
\centering
\renewcommand{\arraystretch}{1.15}
\setlength{\tabcolsep}{6pt}
\caption{\small
Testing $\ell_\infty$ errors, training times, and inference times for the 1D Burgers problem.
\rtwoedits{For the error and inference-time columns, the best value is shown in bold and
the second-best value is underlined. 
Training times are reported for
computational context but are not ranked across model families because the
reported unit differs between full-batch DeepONet and minibatch FNO models.}
For POU-FNO variants, the FNO backbone used by the model is indicated in parentheses.
Rows corresponding to proposed adaptive PoU-based models are highlighted.
}
\vspace{0.05cm}

\footnotesize
\begin{adjustbox}{width=\linewidth}
\begin{tabular}{c l c c c}
\toprule
 & Model & $\ell_\infty$ Error & \rtwoedits{Mean Training Time$^\dagger$ (s)} & Inference Time (s) \\
\midrule

\multirow{8}{*}{\rotatebox{90}{\textbf{DeepONets}}}
& Vanilla
 & $\num{5.17e-01} \pm \num{4.9e-03}$
 & $\num{2.61e-04} \pm \num{5.6e-06}$
 & \best{$\num{5.56e-05} \pm \num{5.8e-07}$}
\\
& (P+1)-Vanilla
 & $\num{5.35e-01} \pm \num{1.1e-02}$
 & $\num{5.59e-04} \pm \num{1.7e-05}$
 & $\num{1.30e-04} \pm \num{2.1e-05}$
\\
& (Vanilla, POU)
 & $\num{5.65e-01} \pm \num{2.1e-02}$
 & $\num{6.60e-04} \pm \num{2.5e-05}$
 & $\num{9.46e-05} \pm \num{6.5e-06}$
\\
& (POD, POU)
 & $\num{4.79e-01} \pm \num{5.2e-03}$
 & $\num{5.36e-04} \pm \num{1.8e-05}$
 & \underline{$\num{9.22e-05} \pm \num{2.6e-06}$}
\\
& (Vanilla, POD, POU)
 & $\num{4.83e-01} \pm \num{6.7e-03}$
 & $\num{6.60e-04} \pm \num{2.2e-05}$
 & $\num{9.27e-05} \pm \num{2.8e-06}$
\\

& \cellcolor{gray!10} HiRefPOU-1L (POD)
 & \cellcolor{gray!10} $\num{4.57e-01} \pm \num{4.6e-03}$
 & \cellcolor{gray!10} $\num{1.36e-03} \pm \num{7.8e-05}$
 & \cellcolor{gray!10} $\num{1.55e-04} \pm \num{8.0e-06}$
\\
& \cellcolor{gray!10} HiRefPOU-2L (POD)
 & \cellcolor{gray!10} $\num{4.56e-01} \pm \num{5.3e-03}$
 & \cellcolor{gray!10} $\num{2.46e-03} \pm \num{5.4e-04}$
 & \cellcolor{gray!10} $\num{2.29e-04} \pm \num{3.4e-05}$
\\
& \cellcolor{gray!10} \METHODNAME-3L (POD)
 & \cellcolor{gray!10} $\num{4.56e-01} \pm \num{7.9e-03}$
 & \cellcolor{gray!10} $\num{1.62e-03} \pm \num{8.6e-05}$
 & \cellcolor{gray!10} $\num{2.66e-04} \pm \num{1.3e-05}$
\\

\midrule

\multirow{6}{*}{\rotatebox{90}{\textbf{FNOs}}}
& GeoFNO
 & $\num{7.98e-01} \pm \num{4.2e-02}$
 & $\num{2.39e-02} \pm \num{2.7e-02}$
 & $\num{2.22e-03} \pm \num{7.2e-04}$
\\

& LocalFNO
 & \best{$\num{2.77e-01} \pm \num{1.0e-01}$} 
 & $\num{8.07e-2} \pm \num{2.5e-3}$
 & $\num{7.30e-2} \pm \num{4.3e-4}$
\\

& \cellcolor{gray!10} PoU-SharedFNO (GeoFNO)
& \cellcolor{gray!10} $\num{6.19e-01} \pm \num{1.7e-01}$
& \cellcolor{gray!10} $\num{4.19e-03} \pm \num{1.5e-04}$
& \cellcolor{gray!10} $\num{2.07e-03} \pm \num{1.3e-05}$ 
\\

& \cellcolor{gray!10} PoU-ExpertFNO (GeoFNO)
 & \cellcolor{gray!10} $\num{4.30e-01} \pm \num{9.3e-02}$
 & \cellcolor{gray!10} $\num{1.09e-02} \pm \num{3.6e-05}$
 & \cellcolor{gray!10} $\num{4.78e-03} \pm \num{2.2e-05}$
\\

& \cellcolor{gray!10} PoU-SharedFNO (LocalFNO)
 & \cellcolor{gray!10} $\num{4.01e-01} \pm \num{8.9e-02}$
 & \cellcolor{gray!10} $\num{5.96e-02} \pm \num{1.1e-03}$
 & \cellcolor{gray!10} $\num{3.70e-03} \pm \num{1.1e-04}$
\\

& \cellcolor{gray!10} PoU-ExpertFNO (LocalFNO)
 & \cellcolor{gray!10} \underline{$\num{3.07e-01} \pm \num{3.2e-02}$}
 & \cellcolor{gray!10} $\num{1.01e-01} \pm \num{6.8e-03}$
 & \cellcolor{gray!10} $\num{5.38e-03} \pm \num{9.8e-05}$
\\

\bottomrule
\end{tabular}
\end{adjustbox}
\footnotesize{
\rtwoedits{
$^\dagger$For full-batch DeepONet-family models, training time denotes one optimizer
update over the complete training set. For minibatch FNO-family models, it denotes
one complete training epoch.
}}
\label{tab:burgers_results}
\end{table}

\textit{Results.}
Table~\ref{tab:main_ol_results} reports testing relative $\ell_2$ errors, and Table~\ref{tab:burgers_results} reports testing $\ell_\infty$ errors, training times, and inference times for all evaluated operator-learning models on the 1D Burgers benchmark.
\rtwoedits{Within the DeepONet family, the HiRefPOU variants improve upon the vanilla and static-partition baselines, showing that learned spatial localization is useful for this problem. In contrast, there is little evidence that increasing hierarchy depth provides an additional benefit: the 1L, 2L, and 3L models obtain nearly identical relative $\ell_2$ errors, and their $\ell_\infty$ differences are small relative to the reported variability.}
\rtwoedits{These results suggest that much of the useful localization for the standard Burgers benchmark can already be captured by a single learned POU level. 
Additional nested subdivision therefore introduces model complexity without a clear corresponding improvement in predictive accuracy. 
This observation is consistent with the partition-count ablations in Appendix~\ref{app:burgers_additional_results}, which show that relatively few learned partitions are sufficient for this benchmark.}

\rtwoedits{The strongest overall results are obtained by LocalFNO. 
The regular periodic grid and translation-structured dynamics of this benchmark are particularly well matched to its local spectral inductive bias. 
The Burgers results therefore illustrate two complementary conclusions: learned POU localization improves the DeepONet representation, but neither deeper HiRefPOU hierarchy nor an additional POU decomposition on top of an already locality-aware Fourier backbone is uniformly advantageous.}

\subsection{Full-Field Burgers Benchmark} \label{sec:burgers_hierarchical3l}
To more directly probe the value of deeper hierarchical partitioning, we introduce a synthetic multiscale Burgers benchmark on the periodic domain $\Omega=[-1,1]$.
Unlike the standard Burgers benchmark, which maps an initial condition to a single final-time solution profile, here the learning task is to map a one-dimensional initial condition to the full solution field over space-time.
The dataset is constructed so that each realization contains nested coarse-, medium-, and fine-scale localized structure, thereby providing a setting in which additional hierarchy in the model can be meaningfully tested.

\textit{Problem Setup.}
We consider the viscous Burgers equation
\begin{align}
  \frac{\partial u}{\partial t} + u \frac{\partial u}{\partial x}
  = \nu \frac{\partial^2 u}{\partial x^2},
  \qquad x \in [-1,1], \quad t \in [0,1],
\end{align}
with periodic boundary conditions and viscosity $\nu=10^{-3}$.
For each sample, the operator-learning objective is to map the initial condition $u_0(x)$ to the full solution field $u(x,t)$.

Each initial condition is generated as the superposition of a smooth low-frequency periodic function and a nested three-level hierarchy of localized oscillatory packets,
\begin{align}
  u_0(x)
  =
  u_{\mathrm{bg}}(x)
  +
  \sum_{\ell=1}^{3}
  b_\ell B(x;c_\ell,w_\ell)\sin(2\pi k_\ell x + \phi_\ell),
\end{align}
where the background term is given by
\begin{align}
  u_{\mathrm{bg}}(x)
  =
  a_1\sin(2\pi x + p_1)
  + a_2\cos(4\pi x + p_2)
  + a_3\sin(6\pi x + p_3),
\end{align}
and the periodic envelope function is
\begin{align}
  B(x;c,w)
  =
  \exp\!\left(-\frac{\sin^2(\pi(x-c))}{w^2}\right).
\end{align}
Here $a_1,a_2,a_3$, $b_1,b_2,b_3$, and the phases $p_1,p_2,p_3,\phi_1,\phi_2,\phi_3$ are randomly sampled.
The packet centers are chosen hierarchically, with $c_2$ sampled near $c_1$ and $c_3$ sampled near $c_2$, so that the medium-scale feature is spatially nested inside the broad-scale feature and the fine-scale feature is nested inside the medium-scale feature.
The widths and frequencies satisfy
\begin{align}
  w_1 > w_2 > w_3,
  \qquad
  k_1 < k_2 < k_3,
\end{align}
so that the three localized packets represent broad, medium, and fine oscillatory structure, respectively.

This construction yields a family of periodic initial conditions with nested multiscale organization.
The broad packet determines a coarse active region, the medium packet refines that region, and the fine packet introduces highly localized oscillatory detail inside it.
Because the locations, widths, amplitudes, and phases are randomized from sample to sample, the dataset does not reduce to a fixed template, while still consistently exhibiting hierarchical localized structure.
We use this benchmark to assess whether deeper adaptive partitioning provides measurable benefits over shallower adaptive models and non-adaptive models.

\textit{Results.}
Table~\ref{tab:main_ol_results} reports the relative $\ell_2$ errors, while Table~\ref{tab:burgersv2_results} reports full-field $\ell_\infty$ errors and timing results.
\rtwoedits{ 
Within the DeepONet family, the HiRefPOU variants are competitive with or improve upon the vanilla and static-partition baselines, indicating that learned localization remains useful when predicting the complete space--time solution field.
}

\rtwoedits{
Despite the explicitly nested structure used to construct this benchmark, the differences among hierarchy depths are small. 
HiRefPOU-2L and HiRefPOU-3L obtain slightly lower mean errors than HiRefPOU-1L on some metrics, but these differences are small relative to the reported variability across seeds. 
Thus, this experiment does not establish a clear advantage from increasing hierarchy depth. 
Instead, it shows that the hierarchical architecture can represent the nested localized structure without degrading accuracy substantially, while a single adaptive level remains highly competitive.
}

\rtwoedits{
The strongest overall results are obtained by the POU-FNO variants with a LocalFNO backbone. 
This indicates that, for a periodic and grid-aligned space--time problem, local Fourier processing provides a particularly strong inductive bias. 
The results therefore reinforce that the benefit of hierarchy and the appropriate localization mechanism depend on both the structure of the target operator and the underlying neural-operator backbone.
}

\begin{table}[!htpb]
\centering
\renewcommand{\arraystretch}{1.15}
\setlength{\tabcolsep}{6pt}
\caption{\small
Testing $\ell_\infty$ errors, training times, and inference times for the Full-Field Burgers problem.
\rtwoedits{For the error and inference-time columns, the best value is shown in bold and
the second-best value is underlined. 
Training times are reported for
computational context but are not ranked across model families because the
reported unit differs between full-batch DeepONet and minibatch FNO models.}
For POU-FNO variants, the FNO backbone used by the model is indicated in parentheses.
Rows corresponding to proposed adaptive PoU-based models are highlighted.
}
\vspace{0.05cm}

\footnotesize
\begin{adjustbox}{width=\linewidth}
\begin{tabular}{c l c c c}
\toprule
 & Model & Full-Field $\ell_\infty$ Error & \rtwoedits{Mean Training Time$^\dagger$ (s)} & Inference Time (s)  \\
\midrule

\multirow{8}{*}{\rotatebox{90}{\textbf{DeepONets}}}
& Vanilla
 &  $\num{3.02e-01} \pm \num{5.7e-03}$
 &  $\num{2.79e-04} \pm \num{7.9e-06}$
 &  \best{$\num{1.31e-04} \pm \num{1.1e-04}$}
\\
& (P+1)-Vanilla
 &  $\num{2.90e-01} \pm \num{5.0e-03}$
 &  \underline{$\num{1.08e-03} \pm \num{2.6e-05}$}
 &  $\num{9.11e-04} \pm \num{3.8e-04}$
\\
& (Vanilla, POU)
 &  $\num{2.72e-01} \pm \num{3.4e-03}$
 &  $\num{1.47e-02} \pm \num{1.6e-05}$
 &  $\num{1.47e-02} \pm \num{1.6e-05}$
\\
& (POD, POU)
 &  $\num{2.69e-01} \pm \num{4.2e-03}$
 &  $\num{1.45e-02} \pm \num{7.4e-05}$
 &  $\num{1.46e-02} \pm \num{3.2e-05}$
\\
& (Vanilla, POD, POU)
 &  $\num{2.84e-01} \pm \num{4.2e-03}$
 &  $\num{4.87e-02} \pm \num{4.2e-03}$
 &  $\num{4.90e-02} \pm \num{1.2e-03}$
\\
& HiRefPOU-1L (POD)
 & \cellcolor{gray!10} $\num{2.72e-01} \pm \num{6.7e-03}$
 & \cellcolor{gray!10} $\num{1.50e-03} \pm \num{2.6e-05}$
 & \cellcolor{gray!10} \underline{$\num{3.57e-04} \pm \num{6.0e-05}$}
\\
& HiRefPOU-2L (POD)
 & \cellcolor{gray!10} $\num{2.69e-01} \pm \num{5.1e-03}$
 & \cellcolor{gray!10} $\num{3.59e-03} \pm \num{6.0e-04}$
 & \cellcolor{gray!10} $\num{5.50e-04} \pm \num{7.1e-05}$
\\
& HiRefPOU-3L (POD)
 & \cellcolor{gray!10} $\num{2.66e-01} \pm \num{5.5e-03}$
 & \cellcolor{gray!10} $\num{1.49e-02} \pm \num{1.6e-03}$
 & \cellcolor{gray!10} $\num{2.12e-03} \pm \num{3.6e-04}$
\\

\midrule

\multirow{4}{*}{\rotatebox{90}{\textbf{FNOs}}}
& GeoFNO
 &  $\num{8.26e-02} \pm \num{8.0e-3}$
 &  $\num{3.97e+00} \pm \num{1.6e-03}$
 &  $\num{4.36e-01} \pm \num{1.6e-04}$
\\

& LocalFNO
 &  $\num{6.47e-02} \pm \num{7.4e-03}$ 
 &  $\num{1.46e-01} \pm \num{1.3e-04}$
 &  $\num{7.33e-02} \pm \num{4.7e-04}$
\\

& \cellcolor{gray!10} POU-SharedFNO (LocalFNO)
 & \cellcolor{gray!10} \best{$\num{5.28e-02} \pm \num{5.0e-03}$}
 & \cellcolor{gray!10} $\num{1.59e-01} \pm \num{1.6e-03}$
 & \cellcolor{gray!10} $\num{7.59e-02} \pm \num{1.1e-04}$
\\

& \cellcolor{gray!10} POU-ExpertFNO (LocalFNO)
 & \cellcolor{gray!10} \underline{$\num{5.82e-02} \pm \num{6.8e-03}$}
 & \cellcolor{gray!10} $\num{2.38e-01} \pm \num{1.6e-03}$
 & \cellcolor{gray!10} $\num{7.94e-02} \pm \num{1.2e-04}$
\\

\bottomrule
\end{tabular}
\end{adjustbox}
\footnotesize{
\rtwoedits{
$^\dagger$For full-batch DeepONet-family models, training time denotes one optimizer
update over the complete training set. For minibatch FNO-family models, it denotes
one complete training epoch.
}}
\label{tab:burgersv2_results}
\end{table}

\subsection{Hierarchical Multiscale Kuramoto--Sivashinsky Benchmark} \label{sec:ks_hierarchical3l}
To complement the Burgers experiments with a more dynamically unstable benchmark, we consider the one-dimensional Kuramoto--Sivashinsky (KS) equation on a periodic domain.
The KS equation exhibits mixed-scale spatiotemporal behavior, including instability growth, nonlinear interaction, and localized fine-scale structure, making it a natural testbed for hierarchical adaptive operator learning.
In contrast to the standard Burgers benchmark, where the output is often a single final-time solution profile, here the learning task is to map an initial condition to the full solution field over space-time.

\textit{Problem Setup.}
We study the periodic KS equation
\begin{align}
  \frac{\partial u}{\partial t}
  + u \frac{\partial u}{\partial x}
  + \frac{\partial^2 u}{\partial x^2}
  + \frac{\partial^4 u}{\partial x^4}
  = 0,
  \qquad x \in [0,L], \quad t \in [0,T],
\end{align}
with periodic boundary conditions.
For each sample, the operator-learning task is to map a one-dimensional initial condition $u_0(x)$ to the full solution field $u(x,t)$.

We construct the initial conditions from a randomized three-level hierarchy of localized oscillatory structure superimposed on a smooth low-frequency periodic background.
Specifically, each initial condition is of the form
\begin{align}
  u_0(x)
  =
  u_{\mathrm{bg}}(x)
  +
  \sum_{\ell=1}^{3}
  b_\ell B(\xi;c_\ell,w_\ell)\sin(2\pi k_\ell \xi + \phi_\ell),
  \qquad \xi = x/L,
\end{align}
where the background component is given by
\begin{align}
  u_{\mathrm{bg}}(x)
  =
  a_1\sin(2\pi \xi + p_1)
  + a_2\cos(4\pi \xi + p_2)
  + a_3\sin(6\pi \xi + p_3),
\end{align}
and the periodic bump envelope is
\begin{align}
  B(\xi;c,w)
  =
  \exp\!\left(-\frac{\sin^2(\pi(\xi-c))}{w^2}\right).
\end{align}
The three packet centers are sampled hierarchically, with $c_2$ chosen near $c_1$ and $c_3$ chosen near $c_2$, so that the medium-scale packet is spatially nested inside the broad-scale packet and the fine-scale packet is
nested inside the medium-scale packet.
The widths and frequencies are sampled so that
\begin{align}
  w_1 > w_2 > w_3,
  \qquad
  k_1 < k_2 < k_3,
\end{align}
thereby yielding broad, medium, and fine localized oscillatory features, respectively.
The amplitudes, phases, and relative signs are randomized across samples.

This benchmark is intended to test whether deeper adaptive partitioning provides measurable gains when the target operator exhibits genuinely nested structure across both space and time.
A shallow partition hierarchy may \roneedits{capture shared spatial localization patterns in the solution field, while additional levels provide further nested spatial subdivision within regions emphasized by the parent gates.}

\textit{Results.}
Table~\ref{tab:main_ol_results} reports the testing relative $\ell_2$ errors, while Table~\ref{tab:ks_results} reports full-field $\ell_\infty$ errors, training times, and inference times for the KS benchmark.
\rtwoedits{
Within the DeepONet family, all HiRefPOU variants improve over the vanilla and static-partition baselines, demonstrating that learned spatial localization is beneficial for this dynamically unstable space--time prediction problem.
}

\rtwoedits{
In contrast to the reaction--diffusion benchmarks, additional hierarchy does not improve the HiRefPOU results. HiRefPOU-1L achieves the lowest mean relative $\ell_2$ and $\ell_\infty$ errors among the DeepONet variants, with performance degrading as additional levels are introduced. 
This provides a clear example in which a shallower adaptive representation is preferable despite the presence of multiscale dynamics. 
One possible interpretation is that the spatial organization learned by a single POU level already captures the useful dataset-level localization, while additional coordinate-based subdivision does not correspond to the dominant variation of this strongly evolving solution operator.
}

\rtwoedits{
The FNO-based models achieve the lowest errors overall, with the LocalFNO-based variants particularly strong. Together with the standard and full-field Burgers results, this reinforces that deeper geometric hierarchy should not be expected to improve performance simply because a problem contains multiscale dynamics; the usefulness of additional partition levels depends on whether those dynamics admit persistent spatial organization that can be exploited by the coordinate-dependent hierarchy.
}

\begin{table}[!htpb]
\centering
\renewcommand{\arraystretch}{1.15}
\setlength{\tabcolsep}{6pt}
\caption{\footnotesize
Testing $\ell_\infty$ errors, training times, and inference times for the KS problem.
\rtwoedits{For the error and inference-time columns, the best value is shown in bold and
the second-best value is underlined. 
Training times are reported for
computational context but are not ranked across model families because the
reported unit differs between full-batch DeepONet and minibatch FNO models.}
For POU-FNO variants, the FNO backbone used by the model is indicated in parentheses.
Rows corresponding to proposed adaptive PoU-based models are highlighted.
}
\vspace{0.05cm}

\footnotesize
\begin{adjustbox}{width=\linewidth}
\begin{tabular}{c l c c c}
\toprule
 & Model & Full-Field $\ell_\infty$ Error & \rtwoedits{Mean Training Time$^\dagger$ (s)} & Inference Time (s)  \\
\midrule

\multirow{8}{*}{\rotatebox{90}{\textbf{DeepONets}}}
& (Vanilla)
 &  $\num{2.45e+00} \pm \num{3.2e-02}$
 &  $\num{7.49e-04} \pm \num{7.8e-05}$
 &  $\num{7.58e-04} \pm \num{2.3e-04}$
\\
& (P+1)-Vanilla
 &  $\num{2.09e+00} \pm \num{3.2e-02}$
 &  $\num{2.22e-03} \pm \num{6.6e-05}$
 &  $\num{2.26e-03} \pm \num{2.3e-05}$
\\
& (Vanilla, POU)
 &  $\num{1.96e+00} \pm \num{3.5e-02}$
 &  $\num{2.09e-02} \pm \num{2.9e-04}$
 &  $\num{2.10e-02} \pm \num{1.7e-04}$
\\
& (POD, POU)
 &  $\num{1.87e+00} \pm \num{2.5e-02}$
 &  $\num{1.31e-02} \pm \num{4.6e-05}$
 &  $\num{9.91e-03} \pm \num{2.9e-03}$
\\
& (Vanilla, POD, POU)
 &  $\num{1.91e+00} \pm \num{2.4e-02}$
 &  $\num{2.03e-02} \pm \num{4.6e-05}$
 &  $\num{2.06e-02} \pm \num{2.3e-05}$
\\

& HiRefPOU-1L (POD)
 & \cellcolor{gray!10} $\num{1.50e+00} \pm \num{1.4e-02}$
 & \cellcolor{gray!10} $\num{1.56e-03} \pm \num{1.4e-05}$
 & \cellcolor{gray!10} \underline{$\num{4.13e-04} \pm \num{2.7e-05}$}
\\
& HiRefPOU-2L (POD)
 & \cellcolor{gray!10} $\num{1.69e+00} \pm \num{1.4e-02}$
 & \cellcolor{gray!10} $\num{4.57e-03} \pm \num{1.3e-06}$
 & \cellcolor{gray!10} \best{$\num{3.38e-04} \pm \num{1.3e-06}$}
\\
& HiRefPOU-3L (POD)
 & \cellcolor{gray!10} $\num{1.80e+00} \pm \num{4.7e-04}$
 & \cellcolor{gray!10} $\num{1.07e-02} \pm \num{1.1e-05}$
 & \cellcolor{gray!10} $\num{6.84e-04} \pm \num{3.9e-06}$
\\

\midrule

\multirow{4}{*}{\rotatebox{90}{\textbf{FNOs}}}
& GeoFNO
 &  $\num{9.09e-01} \pm \num{1.5e-02}$
 &  $\num{8.65e-01} \pm \num{1.3e-02}$
 &  $\num{1.51e-01} \pm \num{3.4e-04}$
\\

& LocalFNO
 &  $\num{8.37e-01} \pm \num{1.3e-02}$
 &  $\num{3.08e-01} \pm \num{1.1e-02}$
 &  $\num{2.21e-02} \pm \num{3.8e-04}$
\\

& \cellcolor{gray!10} POU-SharedFNO (LocalFNO)
 & \cellcolor{gray!10} \best{$\num{8.23e-01} \pm \num{1.1e-02}$}
 & \cellcolor{gray!10} $\num{3.22e-01} \pm \num{7.5e-03}$
 & \cellcolor{gray!10} $\num{2.24e-02} \pm \num{4.5e-04}$
\\

& \cellcolor{gray!10} POU-ExpertFNO (LocalFNO)
 & \cellcolor{gray!10} \underline{$\num{8.36e-01} \pm \num{2.4e-02}$}
 & \cellcolor{gray!10} $\num{4.43e-01} \pm \num{1.6e-02}$
 & \cellcolor{gray!10} $\num{2.85e-02} \pm \num{1.5e-03}$
\\

\bottomrule
\end{tabular}
\end{adjustbox}
\footnotesize{
\rtwoedits{
$^\dagger$For full-batch DeepONet-family models, training time denotes one optimizer
update over the complete training set. For minibatch FNO-family models, it denotes
one complete training epoch.
}}
\label{tab:ks_results}
\end{table}

\subsection{Discussion}
\label{sec:overall_results_discussion}

The results reveal a problem-dependent relationship between adaptive spatial partitioning, hierarchical refinement, and the choice of operator-learning backbone.
Across the DeepONet-family models, the proposed HiRefPOU architectures consistently improve over the corresponding vanilla and static-partition DeepONet baselines.
This trend is especially clear on the cavity flow and reaction--diffusion benchmarks, where the learned partitions provide a mechanism for 
\rthreeedits{assigning greater weight to local experts}
near coefficient interfaces, localized features, or regions of strong spatial heterogeneity.

The comparison with FNO-based models is more nuanced.
On the one-dimensional periodic Burgers and Kuramoto--Sivashinsky benchmarks, LocalFNO and PoU-FNO variants are highly competitive and sometimes achieve the lowest errors overall.
This is not unexpected: these benchmarks are defined on regular periodic grids, and their solution fields contain coherent space--time structures that are well matched to Fourier-based convolutional representations.
In such settings, LocalFNO provides a strong inductive bias by combining local spatial processing with spectral information, and the advantage of adaptive DeepONet partitioning is reduced.
At the same time, the PoU-FNO results show that the proposed partition-of-unity mechanism can also improve Fourier-based backbones in some regimes, most notably on the Full-Field Burgers and KS benchmarks.

For the PoU-FNO variants, the backbone is selected to match the geometry and data representation of each benchmark.
For the one-dimensional periodic Burgers and KS problems, we use LocalFNO as the backbone because these problems are naturally represented on regular space--time grids and local Fourier structure is a strong modeling prior.
For cavity flow and the reaction--diffusion problems, we use GeoFNO as the backbone because these datasets involve geometry-dependent or point-cloud-based spatial discretizations for which GeoFNO is the more appropriate Fourier baseline.
Thus, the parenthetical labels in the PoU-SharedFNO and PoU-ExpertFNO rows identify the underlying FNO architecture being augmented by the PoU mechanism.
This design choice allows the PoU extension to be evaluated on top of a suitable FNO backbone rather than against a mismatched baseline.

The reaction--diffusion results also illustrate the setting in which the HiRefPOU-DeepONet architecture is most effective.
In the 2D and 3D reaction--diffusion problems, the solution operators are shaped by discontinuous reaction coefficients, spatially varying diffusion, and localized gradients.
Here, the hierarchical DeepONet variants achieve the best relative $\ell_2$ errors, while the FNO-based models are substantially less accurate.
This suggests that adaptive spatial partitioning is particularly beneficial when the solution operator contains localized heterogeneity that is not naturally aligned with a global spectral representation.

Finally, the hierarchy does not uniformly improve performance simply by adding levels.
On some benchmarks, the single-level or two-level variant is sufficient, while the three-level model gives the best results on others.
This behavior is consistent with the ablation studies: the benefit of additional hierarchy depends on whether the underlying problem contains \roneedits{persistent localized structure that can benefit from additional nested spatial subdivision.}
Thus, the main conclusion is not that deeper hierarchies are always better, but that hierarchical POU \roneedits{organization} provides a flexible mechanism for allocating local capacity \roneedits{through nested spatial partitions when the structure of the PDE solution operator benefits from such localization.}
\rtwoedits{For new problems, we therefore recommend using HiRefPOU-1L as the default starting point and introducing additional hierarchy only when performance on held-out data justifies the added model capacity and dense evaluation cost.}

\subsubsection{\rtwoedits{Limitations}}
\rtwoedits{
The POU gates used in this work depend only on the evaluation coordinates and
therefore do not adapt to individual input instances.
Instead, the learned partitions capture dataset-level geometric structure that
recurs at consistent spatial locations across the training distribution.
Because the gates are shared across input samples, they cannot move or reshape
their partition boundaries in response to sample-dependent features.
For benchmarks such as Burgers and Kuramoto--Sivashinsky, where localized
structures may shift across samples, any improvement should therefore be
interpreted as a benefit of a shared coordinate-dependent localization
representation rather than as evidence that the gates track the features of
each individual input.
Conditioning the gating network on the input function or its branch-network
representation is a natural extension that could enable instance-dependent
partitioning in such settings.
}

\subsubsection{Broader Impacts}

\rtwoedits{
The methods developed in this work are intended as surrogate modeling tools for scientific computing and should not be interpreted as certified numerical solvers. As with other learned operator models, predictive accuracy may degrade outside the training distribution or in regimes that are insufficiently represented by the available data. Consequently, deployment in safety-critical or otherwise consequential applications should include independent validation against trusted numerical or experimental references, appropriate uncertainty or error assessment, and problem-specific verification of physically relevant quantities. The present study focuses on methodological development and benchmark PDE problems and does not involve human participants, sensitive data, or automated decision-making about individuals.
}
\section{Conclusion} \label{sec:conclusion}
This work introduced \METHODNAME, a hierarchical partition-of-unity (POU) framework for learning PDE solution operators with localized and multiscale structure.
The method combines adaptive POU gating with residual\roneedits{-style additive level-wise representations, yielding a structured mixture-of-experts architecture in which a global representation is combined with one or more nested levels of spatially weighted local expert representations.} 
Unlike purely global operator models, the proposed hierarchy provides spatially localized expressivity while preserving a smooth global prediction through the differentiable POU formulation.
\rthreeedits{
We view HiRefPOU as a localized neural-operator architecture built from established
POU, mixture-of-experts, global--local representation, and hierarchical-partitioning
ideas rather than as a fundamentally new operator-learning paradigm.
}
\rtwoedits{
Its architectural contribution is the particular integration of learned hybrid
geometric gating, nested parent--child POU weights, and additive level-wise
operator representations, together with their study in DeepONet- and FNO-type
models.
}

Across a diverse set of benchmark problems, \METHODNAME\ consistently improves over the corresponding vanilla and static-partition DeepONet baselines.
The gains are most pronounced on problems with localized heterogeneity, sharp coefficient variation, or spatially concentrated error structures, such as cavity flow and the reaction--diffusion benchmarks.
\rtwoedits{The learned partitions capture dataset-shared spatial structure associated with problem-specific features} and provide an interpretable mechanism for allocating representational capacity to regions where the solution operator is more difficult to approximate.

The comparison with FNO-based models shows that the best-performing backbone depends on the structure of the problem.
For periodic, grid-aligned space--time benchmarks such as Burgers and Kuramoto--Sivashinsky, LocalFNO and POU-FNO variants are highly competitive and sometimes achieve the lowest errors overall.
For heterogeneous reaction--diffusion problems, the HiRefPOU-DeepONet variants achieve the strongest accuracy, suggesting that adaptive spatial partitioning is especially valuable when the operator contains localized features that are not naturally captured by global spectral representations.
These results indicate that the proposed POU mechanism is not limited to a single neural-operator family: it can improve DeepONet-style models directly and can also be used as a localization module for Fourier-based backbones.

Overall, the results support a nuanced conclusion.
Hierarchical POU \roneedits{organization} is not uniformly superior to every Fourier-based architecture on every benchmark, nor does increasing the number of levels always improve accuracy.
Rather, the proposed framework provides a flexible and interpretable way to introduce adaptive locality into neural operator learning.
When the PDE solution operator contains localized, heterogeneous, or multiscale structure, this localization can substantially improve accuracy without relying solely on increased parameter count.
These findings suggest that structured, geometry-aware localization is an important ingredient for building reliable operator-learning models for complex scientific computing problems.


\appendix
\section{\rtwoedits{Full Experimental Details}} \label{sec:apx_full_exp_details}

  \begin{table}[!htbp]
  \centering
  \color{black}
  \caption{\rtwoedits{
  HiRefPOU hierarchy configurations used for each benchmark.
  $N_\ell$ denotes the number of experts at level $\ell$.
  Parent--child allocations list the number of children assigned to each
  parent at successive levels. The notation $[2]_{\times k}$ denotes a
  length-$k$ list in which every parent has two children.
  }}
  \label{tab:hierarchy_config}
  \small
  \renewcommand{\arraystretch}{1.18}

  \begin{tabularx}{\linewidth}{@{}l c l Y c }
  \toprule
  \textbf{Benchmark} &
  \textbf{Variant} &
  \makecell{\textbf{Experts}\\\textbf{by level}} &
  \makecell{\textbf{Parent--child}\\\textbf{allocation}} &
  \makecell{\textbf{Leaf}\\\textbf{partitions}} \\
  \midrule

  \multirow{3}{*}{Cavity}
  & 1L
  & $N_1=4$
  & --
  & 4
  \\

  & 2L
  & $(N_1,N_2)=(2,2)$
  & \makecell[l]{$L_1\!\to\!L_2:$ $[0,2]$}
  & 3
  \\

  & 3L
  & $(N_1,N_2,N_3)=(2,2,4)$
  & \makecell[l]{
      $L_1\!\to\!L_2:\ [0,2]$, \;
      $L_2\!\to\!L_3:\ [2,2]$
    }
  & 4
  \\

  \midrule

  \multirow{3}{*}{2D RD}
  & 1L
  & $N_1=6$
  & --
  & 6
  \\

  & 2L
  & $(N_1,N_2)=(2,4)$
  & \makecell[l]{$L_1\!\to\!L_2:$ $[2,2]$}
  & 4
  \\

  & 3L
  & $(N_1,N_2,N_3)=(2,4,8)$
  & \makecell[l]{
      $L_1\!\to\!L_2:\ [2,2]$, \;
      $L_2\!\to\!L_3:\ [2]_{\times4}$
    }
  & 8
  \\

  \midrule

  \multirow{3}{*}{3D RD}
  & 1L
  & $N_1=8$
  & --
  & 8
  \\

  & 2L
  & $(N_1,N_2)=(3,5)$
  & \makecell[l]{$L_1\!\to\!L_2:$ $[2,2,1]$}
  & 5
  \\

  & 3L
  & $(N_1,N_2,N_3)=(3,5,10)$
  & \makecell[l]{
      $L_1\!\to\!L_2:\ [2,2,1]$, \;
      $L_2\!\to\!L_3:\ [2]_{\times5}$
    }
  & 10
  \\

  \midrule

  \multirow{3}{*}{1D Burgers}
  & 1L
  & $N_1=3$
  & --
  & 3
  \\

  & 2L
  & $(N_1,N_2)=(1,2)$
  & \makecell[l]{$L_1\!\to\!L_2:$ $[2]$}
  & 2
  \\

  & 3L
  & $(N_1,N_2,N_3)=(2,4,8)$
  & \makecell[l]{
      $L_1\!\to\!L_2:\ [2,2]$, \;
      $L_2\!\to\!L_3:\ [2]_{\times4}$
    }
  & 8
  \\

  \midrule

  \multirow{3}{*}{\makecell[l]{Full-field\\Burgers}}
  & 1L
  & $N_1=6$
  & --
  & 6
  \\

  & 2L
  & $(N_1,N_2)=(3,6)$
  & \makecell[l]{$L_1\!\to\!L_2:$ $[2]_{\times3}$}
  & 6
  \\

  & 3L
  & $(N_1,N_2,N_3)=(6,12,24)$
  & \makecell[l]{
      $L_1\!\to\!L_2:\ [2]_{\times6}$, \;
      $L_2\!\to\!L_3:\ [2]_{\times12}$
    }
  & 24
  \\

  \midrule

  \multirow{3}{*}{\makecell[l]{Kuramoto--\\Sivashinsky}}
  & 1L
  & $N_1=6$
  & --
  & 6
  \\

  & 2L
  & $(N_1,N_2)=(3,6)$
  & \makecell[l]{$L_1\!\to\!L_2:$ $[2]_{\times3}$}
  & 6
  \\

  & 3L
  & $(N_1,N_2,N_3)=(6,12,24)$
  & \makecell[l]{
      $L_1\!\to\!L_2:\ [2]_{\times6}$, \;
      $L_2\!\to\!L_3:\ [2]_{\times12}$
    }
  & 24
  \\

  \bottomrule
  \end{tabularx}
  \end{table}

\begin{table*}[!htbp]
\color{black}
\centering
\caption{\rtwoedits{
Network architecture specifications used for the HiRefPOU experiments.
Depth denotes the number of hidden trainable layers and excludes the final
output projection. For convolutional branch networks, the hidden-width entry
lists the output channels of the convolutional layers followed by the width
of the dense hidden layer. The branch output contains the $p$ learned
DeepONet coefficients and the $r_{\mathrm{POD}}$ POD coefficients.
Where architectures differ among hierarchy variants, the corresponding
1L, 2L, and 3L values are listed explicitly. Hierarchy configurations are
reported in Table~\ref{tab:hierarchy_config}.
The POU center locations are initialized using the same prescribed center
configurations as the corresponding prior-work benchmarks where applicable;
for the benchmarks introduced in this work, the initial center locations are
prescribed as part of the benchmark-specific model configuration.
In all cases, the center locations remain trainable and are optimized jointly
with the remaining model parameters during training.
Unless otherwise specified, neural-network weights use the default Haiku initialization; the POU centers use the prescribed initial configurations described above and remain trainable thereafter.
}}
\label{tab:model_architectures}
\renewcommand{\arraystretch}{1.15}
\scriptsize

\begin{adjustbox}{max width=\textwidth}
\begin{tabular}{llccccl}
\toprule
\textbf{Benchmark}
&
\textbf{Network component}
&
\makecell{\textbf{Input}\\\textbf{dim.}}
&
\makecell{\textbf{Hidden}\\\textbf{width(s)}}
&
\textbf{Depth}
&
\makecell{\textbf{Output}\\\textbf{dim.}}
&
\textbf{Activation}
\\
\midrule

\multirow{4}{*}{Cavity flow}
& Branch CNN
& $65\times65\times2$
& $64,128,128$
& 3
& 106
& $\tanh$
\\

& Global trunk
& 2
& 128
& 3
& 100
& $\tanh$
\\

& Local trunk expert
& 2
& 128
& 3
& 100
& $\tanh$
\\

& POU neural gate
& 2
& 128
& 2
& $N_{\mathrm{part}}$
& $\tanh$
\\

\midrule

\multirow{4}{*}{\makecell[l]{2D reaction--\\diffusion}}
& Branch CNN
& $20\times20\times1$
& $64,128,128$
& 3
& 120
& ReLU
\\

& Global trunk
& 2
& 128
& 3
& 100
& ReLU
\\

& Local trunk expert
& 2
& 128
& 3
& 100
& ReLU
\\

& POU neural gate
& 2
& 128
& 2
& $N_{\mathrm{part}}$
& $\tanh$
\\

\midrule

\multirow{4}{*}{\makecell[l]{3D reaction--\\diffusion}}
& Branch MLP
& 4325
& 128
& 3
& 120
& ReLU
\\

& Global trunk
& 3
& 128
& 3
& 100
& ReLU
\\

& Local trunk expert
& 3
& 128
& 3
& 100
& ReLU
\\

& POU neural gate
& 3
& 64
& 2
& $N_{\mathrm{part}}$
& $\tanh$
\\

\midrule

\multirow{4}{*}{Standard Burgers}
& Branch MLP
& 128
& 128
& 3
& 120
& ReLU
\\

& Global trunk
& 1
& 128
& 2
& 100
& ReLU
\\

& Local trunk expert
& 1
& 128
& 2
& 100
& ReLU
\\

& POU neural gate
& 1
& 16
& 1
& $N_{\mathrm{part}}$
& $\tanh$
\\

\midrule

\multirow{4}{*}{\makecell[l]{Full-field\\Burgers}}
& Branch MLP
& 128
& 128
& 3
& 120
& ReLU
\\

& Global trunk
& 2
& 128
& 3
& 100
& ReLU
\\

& Local trunk expert
& 2
& 128
& 3
& 100
& ReLU
\\

& POU neural gate
& 2
& 128
& 3
& $N_{\mathrm{part}}$
& $\tanh$
\\

\midrule

\multirow{4}{*}{\makecell[l]{Kuramoto--\\Sivashinsky}}
& Branch MLP
& 128
& 128
& 3
& 120
& ReLU
\\

& Global trunk
& 2
& 128
& 3
& 100
& ReLU
\\

& Local trunk expert
& 2
& 128
& 3
& 100
& ReLU
\\

& POU neural gate
& 2
& 128
& 3
& $N_{\mathrm{part}}$
& $\tanh$
\\

\bottomrule
\end{tabular}
\end{adjustbox}

\vspace{1mm}
\begin{minipage}{0.98\textwidth}
\scriptsize
For the cavity and 2D reaction--diffusion branch CNNs, the convolutional
layers use 64 and 128 output channels, respectively, with kernel size 5,
stride 2, and valid padding, followed by flattening and a 128-unit dense
hidden layer. The cavity branch output dimension is $100+6=106$; all other
branch outputs have dimension $100+20=120$.
\end{minipage}
\end{table*}

  \begin{table*}[!htbp]
  \color{black}
  \centering
  \caption{
  \rtwoedits{
  Architecture specifications for all Fourier neural operator (FNO)-based
  models used in the experiments. Fourier modes are reported per spatial or
  space--time dimension. GeoFNO and the POU--FNO variants use five spectral
  layers. For LocalFNO, the number of local neural-operator layers is reported.
  Input-channel counts include coordinate channels for GeoFNO and POU--FNO,
  but not for LocalFNO, which constructs its coordinate representation
  internally. 
  GeoFNO and GeoFNO-based POU-FNO variants use five spectral layers; LocalFNO and LocalFNO-based POU variants use the number of local operator layers reported below.
  }}
  \label{tab:fno_architectures}
  \renewcommand{\arraystretch}{1.15}
  \scriptsize

  \begin{adjustbox}{max width=\textwidth}
  \begin{tabular}{
  ll
  c
  c
  c
  c
  c
  c
  c
  r
  }
  \toprule
  \textbf{Benchmark}
  &
  \textbf{Model}
  &
  \makecell{\textbf{Input}\\\textbf{channels}}
  &
  \makecell{\textbf{FNO}\\\textbf{width}}
  &
  \makecell{\textbf{Fourier}\\\textbf{modes}}
  &
  \makecell{\textbf{FNO}\\\textbf{layers}}
  &
  \makecell{\textbf{Output}\\\textbf{channels}}
  &
  \makecell{\textbf{Local / expert}\\\textbf{structure}}
  &
  \makecell{\textbf{POU / gate}\\\textbf{configuration}}
  \\
  \midrule

  \multirow{4}{*}{Cavity flow}
  & GeoFNO
  & 4 & 64 & $(20,20)$ & 5 & 2
  & Single global GeoFNO
  & --
  \\

  & LocalFNO
  & 2 & 50 & $(20,20)$ & 3 & 2
  & 3 local operator layers
  & --
  \\

  & POU-SharedFNO
  & 4 & 64 & $(20,20)$ & 5 & 2
  & \makecell{Shared GeoFNO backbone;\\expert heads $(64,64)$}
  & \makecell{4 partitions;\\gate width 128, depth 2}
  \\

  & POU-ExpertFNO
  & 4 & 32 & $(18,18)$ & 5 & 2
  & \makecell{4 independent GeoFNO experts;\\expert heads $(64,64)$}
  & \makecell{4 partitions;\\gate width 128, depth 2}
  \\

  \midrule


  \multirow{4}{*}{\makecell[l]{2D reaction--\\diffusion}}
  & GeoFNO
  & 3 & 32 & $(12,12)$ & 5 & 1
  & Single global GeoFNO
  & --
  \\

  & LocalFNO
  & 1 & 30 & $(12,12)$ & 2 & 1
  & 2 local operator layers
  & --
  \\

  & POU-SharedFNO 
  & 3 & 32 & $(10,10)$ & 5 & 1
  & \makecell{Shared GeoFNO backbone;\\expert heads $(32,32)$}
  & \makecell{6 partitions;\\gate width 128, depth 2}
  \\

  & POU-ExpertFNO  
  & 3 & 24 & $(6,6)$ & 5 & 1
  & \makecell{4 independent GeoFNO experts;\\expert heads $(32,32)$}
  & \makecell{4 partitions;\\gate width 128, depth 2}
  \\

  \midrule


  \multirow{4}{*}{\makecell[l]{3D reaction--\\diffusion}}
  & GeoFNO
  & 4 & 16 & $(7,7,7)$ & 5 & 1
  & Single global GeoFNO
  & --
  \\

  & LocalFNO
  & 1 & 12 & $(8,8,8)$ & 2 & 1
  & 2 local operator layers
  & --
  \\

  & POU-SharedFNO 
  & 4 & 16 & $(6,6,6)$ & 5 & 1
  & \makecell{Shared GeoFNO backbone;\\expert heads $(32,32)$}
  & \makecell{4 partitions;\\gate width 128, depth 2}
  \\

  & POU-ExpertFNO
  & 4 & 16 & $(5,5,5)$ & 5 & 1
  & \makecell{2 independent GeoFNO experts;\\expert heads $(32,32)$}
  & \makecell{2 partitions;\\gate width 128, depth 2}
  \\

  \midrule


  \multirow{4}{*}{1D Burgers}
  & GeoFNO
  & 129 & 50 & 12 & 5 & 1
  & Single global GeoFNO
  & --
  \\

  & LocalFNO
  & 1 & 32 & 24 & 6 & 1
  & 6 local operator layers
  & --
  \\

  & POU-SharedFNO
  & 129 & 50 & 12 & 5 & 1
  & \makecell{Shared GeoFNO backbone;\\expert heads $(32,32)$}
  & \makecell{3 partitions;\\gate width 16, depth 1}
  \\

  & POU-ExpertFNO
  & 129 & 16 & 6 & 5 & 1
  & \makecell{3 independent GeoFNO experts;\\expert heads $(32,32)$}
  & \makecell{3 partitions;\\gate width 16, depth 1}
  \\

  \midrule


  \multirow{4}{*}{\makecell[l]{Full-field\\Burgers}}
  & GeoFNO
  & 3 & 12 & $(18,18)$ & 5 & 1
  & Single global GeoFNO
  & --
  \\

  & LocalFNO
  & 1 & 12 & $(18,18)$ & 2 & 1
  & 2 local operator layers
  & --
  \\

  & POU-SharedFNO
  & 1 & 12 & $(18,18)$ & 2 & 1
  & \makecell{Shared LocalFNO backbone;\\expert heads $(12,12)$}
  & \makecell{4 partitions;\\gate width 12, depth 1}
  \\

  & POU-ExpertFNO
  & 1 & 12 & $(8,8)$ & 2 & 1
  & 2 independent LocalFNO experts
  & \makecell{4 partitions;\\gate width 12, depth 1}
  \\

  \midrule


  \multirow{4}{*}{\makecell[l]{Kuramoto--\\Sivashinsky}}
  & GeoFNO
  & 3 & 12 & $(18,18)$ & 5 & 1
  & Single global GeoFNO
  & --
  \\

  & LocalFNO
  & 1 & 12 & $(18,18)$ & 2 & 1
  & 2 local operator layers
  & --
  \\

  & POU-SharedFNO
  & 1 & 12 & $(18,18)$ & 2 & 1
  & \makecell{Shared LocalFNO backbone;\\expert heads $(12,12)$}
  & \makecell{4 partitions;\\gate width 12, depth 1}
  \\

  & POU-ExpertFNO
  & 1 & 12 & $(8,8)$ & 2 & 1
  & 2 independent LocalFNO experts
  & \makecell{4 partitions;\\gate width 12, depth 1}
  \\
  
  \bottomrule
  \end{tabular}
  \end{adjustbox}
  \end{table*}

  \begin{table}[!htbp]
  \color{black}
  \centering
  \caption{
  \rtwoedits{
  POD basis construction used in the DeepONet experiments. All POD statistics
  and basis vectors are computed exclusively from the training outputs. For
  cavity flow, separate rank-3 bases are constructed for the two velocity
  components, giving a total POD rank of 6.
  The POD ranks were fixed a priori and were not selected using test-set performance.
  }}
  \label{tab:pod_details}
  \renewcommand{\arraystretch}{1.15}
  \small
  \begin{tabular}{lccc}
  \toprule
  \textbf{Benchmark}
  &
  \textbf{POD rank}
  &
  \makecell{\textbf{Basis}\\\textbf{samples}}
  &
  \makecell{\textbf{Centering /}\\\textbf{normalization}}
  \\
  \midrule

  Cavity
  & $6$ $(3+3)$
  & 100 training outputs
  & \makecell{Pointwise training mean/std;\\separate basis per component}
  \\

  2D reaction--diffusion
  & 20
  & 1,000 training outputs
  & Pointwise training mean/std
  \\

  3D reaction--diffusion
  & 20
  & 1,000 training outputs
  & Pointwise training mean/std
  \\

  1D Burgers
  & 20
  & 1,000 training outputs
  & Pointwise training mean/std
  \\

  Full-Field Burgers
  & 20
  & 1,000 training outputs
  & Pointwise training mean/std
  \\

  Kuramoto--Sivashinsky
  & 20
  & 1,000 training outputs
  & \makecell{Global training mean/std, followed by\\pointwise training mean/std}
  \\

  \bottomrule
  \end{tabular}
  \end{table} 
\rtwoedits{
  For an output snapshot matrix $X\in\mathbb{R}^{N_{\mathrm{train}}\times n}$,
  we compute the pointwise training mean $\mu_j$ and standard deviation
  $\sigma_j$ and form
  \[
      \widetilde{X}_{ij}
      =
      \frac{X_{ij}-\mu_j}{\sigma_j+10^{-9}}.
  \]
  The POD modes are obtained from the eigendecomposition of
  $\widetilde{X}^{\mathsf T}\widetilde{X}/(N_{\mathrm{train}}-1)$.
  The training mean is retained as the first basis vector, followed by the
  leading covariance eigenvectors, with the total number of retained vectors
  equal to the rank reported in Table~\ref{tab:pod_details}. No test outputs
  are used to compute the mean, standard deviation, or basis vectors.
  For cavity flow, this procedure is applied independently to the two velocity
  components. The Kuramoto--Sivashinsky outputs additionally undergo the
  scalar training-set normalization used by the dataset preprocessing pipeline
  before the pointwise POD standardization.
}

  \begin{table*}[!htbp]
  \color{black}
  \centering
  \caption{
  \rtwoedits{
  Training and optimization settings for DeepONet-based experiments. 
  Adam and AdamW use the Optax default parameters
  $\beta_1=0.9$, $\beta_2=0.999$, and $\epsilon=10^{-8}$.
  Cavity models use Adam. All other rows use AdamW with weight decay
  $10^{-4}$. The inverse-time schedule is
  $\eta_t=\eta_0/(1+\gamma t)$; the implementation uses
  \texttt{decay\_steps=1} without staircase decay.
  Each HiRefPOU model uses $\lambda_{cov}=\lambda_{cont}=10^{-4}$ and are linearly annealed to zero over the 150,000 gradient updates.
  }}
  \label{tab:training_protocol}
  \renewcommand{\arraystretch}{1.15}
  \scriptsize
  \begin{adjustbox}{max width=\textwidth}
  \begin{tabular}{lclcc}
  \toprule
  \textbf{Benchmark}
  &
  \makecell{\textbf{Initial}\\\textbf{LR}}
  &
  \textbf{LR schedule}
  &
  \makecell{\textbf{Batch}\\\textbf{size}}
  &
  \makecell{\textbf{Gradient}\\\textbf{updates}}
  \\
  \midrule

  Cavity
  & $3\times10^{-4}$
  & \makecell[l]{Inverse-time decay\\$(\gamma=10^{-4})$}
  & Full batch (100)
  & 150,000
  \\

  2D reaction--diffusion
  & $3\times10^{-4}$
  & \makecell[l]{Inverse-time decay\\$(\gamma=10^{-4})$}
  & Full batch (1,000)
  & 150,000
  \\

  3D reaction--diffusion
  & $6\times10^{-3}$
  & \makecell[l]{Inverse-time decay\\$(\gamma=10^{-4})$}
  & Full batch (1,000)
  & 150,000
  \\

  Standard Burgers
  & $3\times10^{-4}$
  & \makecell[l]{Inverse-time decay\\$(\gamma=10^{-4})$}
  & Full batch (1,000)
  & 150,000
  \\

  Full-field Burgers
  & $3\times10^{-4}$
  & \makecell[l]{Inverse-time decay\\$(\gamma=5\times10^{-5})$}
  & Full batch (1,000)
  & 150,000
  \\

  Kuramoto--Sivashinsky
  & $3\times10^{-4}$
  & \makecell[l]{Inverse-time decay\\$(\gamma=10^{-4})$}
  & Full batch (1,000)
  & 150,000
  \\


  \bottomrule
  \end{tabular}
  \end{adjustbox}

  \vspace{2pt}
  \begin{minipage}{0.98\textwidth}
  \end{minipage}
  \end{table*}

  \begin{table*}[!htbp]
  \color{black}
  \centering
  \caption{
  \rtwoedits{
  Standardized training and optimization settings for the FNO-based
  experiments. All FNO models use mini-batches of 32 input functions and
  minimize the mean squared output-field error. AdamW uses the Optax
  default parameters $\beta_1=0.9$, $\beta_2=0.999$, and
  $\epsilon=10^{-8}$, with weight decay $10^{-4}$. The inverse-time
  schedule is $\eta_t=\eta_0/(1+\gamma t)$, using
  \texttt{decay\_steps=1} without staircase decay. The POU--FNO models
  jointly optimize the backbone or experts, output heads, and POU gate.
  No containment or coverage penalties are applied:
  $\lambda_{\mathrm{cont}}=\lambda_{\mathrm{cov}}=0$.
  The number of epochs is selected to give approximately 150,000 gradient
  updates, computed as
  $N_{\mathrm{epoch}}\lceil N_{\mathrm{train}}/B\rceil$.
  }}
  \label{tab:fno_training_protocol}
  \renewcommand{\arraystretch}{1.15}
  \scriptsize

  \begin{adjustbox}{max width=\textwidth}
  \begin{tabular}{lclccc}
  \toprule
  \textbf{Benchmark}
  &
  \makecell{\textbf{Initial}\\\textbf{LR}}
  &
  \textbf{LR schedule}
  &
  \makecell{\textbf{Batch}\\\textbf{size}}
  &
  \makecell{\textbf{Training}\\\textbf{epochs}}
  &
  \makecell{\textbf{Gradient}\\\textbf{updates}}
  \\
  \midrule

  Cavity
  & $3\times10^{-4}$
  & \makecell[l]{Inverse-time decay\\$(\gamma=10^{-4})$}
  & 32
  & 37,500
  & 150,000
  \\

  2D reaction--diffusion
  & $3\times10^{-4}$
  & \makecell[l]{Inverse-time decay\\$(\gamma=10^{-4})$}
  & 32
  & 4,688
  & 150,016
  \\

  3D reaction--diffusion
  & $3\times10^{-4}$
  & \makecell[l]{Inverse-time decay\\$(\gamma=10^{-4})$}
  & 32
  & 4,688
  & 150,016
  \\

  Standard Burgers
  & $3\times10^{-4}$
  & \makecell[l]{Inverse-time decay\\$(\gamma=10^{-4})$}
  & 32
  & 4,688
  & 150,016
  \\

  Full-field Burgers
  & $3\times10^{-4}$
  & \makecell[l]{Inverse-time decay\\$(\gamma=5\times10^{-5})$}
  & 32
  & 4,688
  & 150,016
  \\

  Kuramoto--Sivashinsky
  & $3\times10^{-4}$
  & \makecell[l]{Inverse-time decay\\$(\gamma=10^{-4})$}
  & 32
  & 4,688
  & 150,016
  \\

  \bottomrule
  \end{tabular}
  \end{adjustbox}
  \end{table*}

  \begin{table*}[!htbp]
  \color{black}
  \centering
  \caption{
  \rtwoedits{
  Dataset provenance, train/test splits, and reference-solution
  details for each benchmark. 
  Reference-grid resolutions are distinguished from
  the subsampled grids supplied to the networks.
  }}
  \label{tab:dataset_details}
  \renewcommand{\arraystretch}{1.15}
  \scriptsize
  \begin{adjustbox}{max width=\textwidth}
  \begin{tabular}{
  l
  l
  l
  c
  c
  l
  l
  l
  }
  \toprule
  \textbf{Benchmark}
  &
  \textbf{Operator / task}
  &
  \makecell{\textbf{Domain /}\\\textbf{network grid}}
  &
  $N_{\mathrm{train}}$
  &
  $N_{\mathrm{test}}$
  &
  \textbf{Dataset source}
  &
  \textbf{Reference solver}
  &
  \textbf{Preprocessing}
  \\
  \midrule

  Cavity flow
  &
  $u_b(x)\mapsto (u,v)(x,y)$
  &
  \makecell[l]{$[0,1]^2$; $65\times65$ grid;\\
  65 lid-velocity sensors}
  &
  100
  &
  10
  &
  \makecell[l]{External benchmark;\\
  Lu et al.\ / Sharma--Shankar}
  &
  LBM
  &
  \makecell[l]{Separate train/test files;\\
  no coordinate or output scaling}
  \\

  2D reaction--diffusion
  &
  $c(\cdot,0)\mapsto c(\cdot,0.5)$
  &
  \makecell[l]{$[0,2]^2$;\\
  2,207 scattered points}
  &
  1,000
  &
  200
  &
  Sharma--Shankar
  &
  High-order RBF--FD
  &
  \makecell[l]{Fixed contiguous split;\\
  no coordinate or output scaling}
  \\

  3D reaction--diffusion
  &
  $c(\cdot,0)\mapsto c(\cdot,0.5)$
  &
  \makecell[l]{Unit ball in $\mathbb{R}^3$;\\
  4,325 scattered points}
  &
  1,000
  &
  200
  &
  Sharma--Shankar
  &
  High-order RBF--FD
  &
  \makecell[l]{Fixed contiguous split;\\
  no coordinate or output scaling}
  \\

  Standard Burgers
  &
  $u_0\mapsto u(\cdot,1)$
  &
  \makecell[l]{$[0,1]$, periodic;\\
  8,192 reference points;\\
  128 network points}
  &
  1,000
  &
  200
  &
  Generated in this work
  &
  \makecell[l]{Chebfun spectral solver;\\
  $\nu=10^{-3}$, 500 time points}
  &
  \makecell[l]{Spatial stride 64;\\
  no coordinate or output scaling}
  \\

  Full-field Burgers
  &
  $u_0\mapsto u(x,t)$
  &
  \makecell[l]{$[-1,1]\times[0,1]$, periodic;\\
  $256\times101$ stored;\\
  $128\times51$ network grid}
  &
  1,000
  &
  200
  &
  Generated in this work
  &
  \makecell[l]{Chebfun spectral solver;\\
  $\nu=10^{-3}$}
  &
  \makecell[l]{Spatial/temporal stride 2;\\
  no coordinate or output scaling}
  \\

  Kuramoto--Sivashinsky
  &
  $u_0\mapsto u(x,t)$
  &
  \makecell[l]{$[0,20)\times[0,30]$, periodic;\\
  $128\times101$ stored;\\
  $64\times51$ network grid}
  &
  1,000
  &
  200
  &
  Generated in this work
  &
  \makecell[l]{Fourier ETDRK4; $\Delta t=10^{-3}$;\\
  burn-in time 23}
  &
  \makecell[l]{Spatial/temporal stride 2;\\
  coordinates mapped to $[0,1]^2$;\\
  training-statistic input/output z-score}
  \\


  \bottomrule
  \end{tabular}
  \end{adjustbox}

  \vspace{2pt}
  \begin{minipage}{0.98\textwidth}
  \end{minipage}
  \end{table*}

\rtwoedits{
Sampling distributions used to generate the Burgers and
  Kuramoto--Sivashinsky datasets. For 1D Burgers, initial conditions
  were sampled from the periodic Gaussian random field
  $\mathcal{N}(0,C)$ with
  $C=\sigma^2(-\Delta+\tau^2I)^{-\gamma}$,
  $\gamma=2.5$, $\tau=5$, and $\sigma=25$. The construction retained
  $K=4096$ positive Fourier modes, with independent standard-normal sine
  and cosine coefficients scaled by
  $\sqrt{2}\sigma((2\pi k)^2+\tau^2)^{-\gamma/2}$.
}

\rtwoedits{For Full-field Burgers, the global-background amplitudes were sampled as
  $a_1\sim\mathcal{U}(0.05,0.12)$,
  $a_2\sim\mathcal{U}(0.02,0.07)$, and
  $a_3\sim\mathcal{U}(0.01,0.04)$, with independent phases
  $p_i\sim\mathcal{U}(0,2\pi)$. The three localized-packet amplitudes were
  $b_1\sim\mathcal{U}(0.08,0.14)$,
  $b_2=s_2\widetilde b_2$ with
  $s_2\in\{-1,1\}$ equiprobable and
  $\widetilde b_2\sim\mathcal{U}(0.04,0.09)$, and
  $b_3=s_3\widetilde b_3$ with
  $\widetilde b_3\sim\mathcal{U}(0.02,0.06)$.
  Their phases were independent
  $\phi_i\sim\mathcal{U}(0,2\pi)$.
  The broad center was $c_1\sim\mathcal{U}(0,1)$, while
  $c_2=(c_1+\epsilon_2)\bmod 1$ and
  $c_3=(c_2+\epsilon_3)\bmod 1$, where
  $\epsilon_2\sim\mathcal{N}(0,0.03^2)$ and
  $\epsilon_3\sim\mathcal{N}(0,0.01^2)$.
  Widths were sampled from
  $w_1\sim\mathcal{U}(0.14,0.22)$,
  $w_2\sim\mathcal{U}(0.045,0.070)$, and
  $w_3\sim\mathcal{U}(0.010,0.025)$.
  Frequencies were discrete uniform on
  $\{1,2,3\}$, $\{4,\ldots,8\}$, and
  $\{10,\ldots,20\}$, respectively.}

  \rtwoedits{For Kuramoto--Sivashinsky, the pre-burn-in initial condition was
  $u_0(x)=\sum_{k=1}^{12}
  [a_k\cos(2\pi kx/L)+b_k\sin(2\pi kx/L)]$, where
  $a_k,b_k\overset{\mathrm{iid}}{\sim}\mathcal{N}(0,k^{-4})$.
  Each profile was rescaled to have maximum absolute amplitude two and
  evolved for a burn-in time of $23$ before recording the subsequent
  trajectory.}


  \begin{table*}[!htbp]
  \centering
  \color{black}
  \caption{
  \rtwoedits{
  Trainable parameter counts for the principal model variants and benchmarks.
  Counts include all trainable network, expert, gating, and projection
  parameters.
  }}
  \label{tab:parameter_counts}
  \renewcommand{\arraystretch}{1.15}
  \scriptsize

  \begin{adjustbox}{max width=\textwidth}
  \begin{tabular}{lrrrrrr}
  \toprule
  \textbf{Model}
  &
  \textbf{Cavity}
  &
  \makecell{\textbf{2D}\\\textbf{RD}}
  &
  \makecell{\textbf{3D}\\\textbf{RD}}
  &
  \makecell{\textbf{1D}\\\textbf{Burgers}}
  &
  \makecell{\textbf{Full-field}\\\textbf{Burgers}}
  &
  \makecell{\textbf{Kuramoto--}\\\textbf{Sivashinsky}}
  \\
  \midrule

  Vanilla
  & 3,488,893 
  & 331,465   
  & 646,089   
  & 108,617
  & 108,745
  & 108,745   
  \\

  $(P+1)$-Vanilla
  & 3,766,125   
  & 686,713     
  & 1,120,777   
  & 285,857
  & 463,993
  & 458,617     
  \\

  (POD, POU)
  & 3,618,491  
  & 565,585    
  & 973,721    
  & 203,557
  & 342,865
  & 334,673
  \\

  (Vanilla, POU)
  & 3,687,025  
  & 622,213    
  & 1,030,477  
  & 260,057
  & 399,493
  & 391,301
  \\

  (Vanilla, POD, POU)
  & 3,728,199  
  & 624,793    
  & 1,033,057  
  & 262,637
  & 402,073
  & 393,881
  \\

  \addlinespace
  HiRefPOU-1L (POD)
  & 3,780,529  
  & 760,143    
  & 1,174,823  
  & 266,356
  & 454,863
  & 454,863
  \\

  HiRefPOU-2L (POD)
  & 3,813,940  
  & 793,554    
  & 1,183,400  
  & 266,645
  & 791,563
  & 791,563
  \\

  HiRefPOU-3L (POD)
  & 4,097,623  
  & 1,327,511  
  & 1,893,067  
  & 772,339
  & 2,053,841
  & 2,053,841
  \\

  \addlinespace
  GeoFNO
  & 3,390,854  
  & 394,757    
  & 419,719    
  & 109,741    
  & 109,077    
  & 109,077    
  \\

  LocalFNO
  & 3,438,652  
  & 337,111    
  & 377,485    
  & 109,089    
  & 109,357    
  & 109,357    
  \\

  POU-SharedFNO
  & 3,701,599 
  & 340,164   
  & 324,112   
  & 111,322   
  & 109,846   
  & 109,846   
  \\

  POU-ExpertFNO
  & 3,088,169  
  & 429,733    
  & 426,777    
  & 196,355    
  & 115,125    
  & 115,125    
  \\

  \bottomrule
  \end{tabular}
  \end{adjustbox}
  \end{table*}

\FloatBarrier
\clearpage

\section{Additional PINN Experiments: Darcy Flow with Discontinuous Coefficients}
\label{sec:appendix_pinn_darcy}

\begin{figure}[h]
    \centering
    \includegraphics[width=0.7\textwidth]{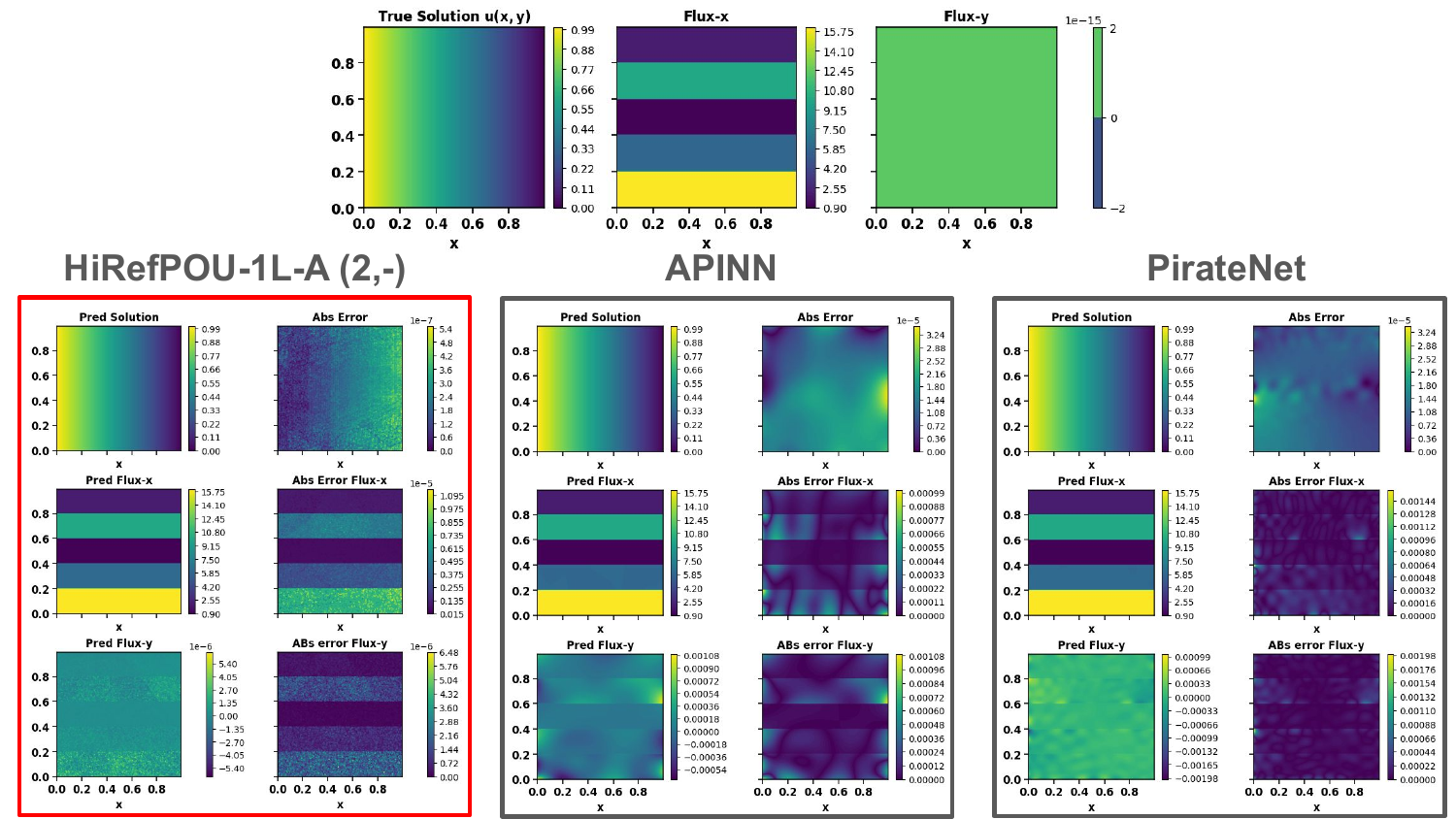}
    \vspace{-0.05cm}
    \caption{\small (Top) the true solution and fluxes. 
    (Bottom) Predicted solutions, fluxes, and absolute errors for the Darcy flow benchmark using HiRefPOU-1L, APINN, and PirateNet. 
    Left bottom figures show the predictions while the right bottom figures show the absolute errors. 
    }\label{fig:df_rough_preds}
\end{figure}

This appendix provides additional technical details for the Darcy flow experiment presented in Section~\ref{results:darcyflow_fiveslab}.
These results are included for completeness and are not intended as an exhaustive comparison with domain-decomposition PINN methods such as XPINNs.
\rtwoedits{All Darcy models were trained for 100,000 Adam updates with
  $\beta_1=0.9$, $\beta_2=0.999$, $\epsilon=10^{-8}$, an initial learning
  rate of $10^{-3}$, and cosine learning-rate decay over the complete
  training run. At each update, 4,096 residual collocation points were
  sampled without replacement from a fixed pool of 30,760 interior
  points. The complete set of 288 Neumann-boundary points was used in the
  boundary loss. The base PINN objective contained the PDE-residual,
  Neumann-boundary, and uniqueness-pin terms, initialized with unit
  weights. 
  }

  \begin{table}[!htbp]
  \color{black}
  \centering
  \caption{
  \rtwoedits{
  Architecture specifications for the Darcy PINN experiments.
  Width and depth refer to the primary network for PirateNet and to the
  expert networks for the POU-based models. Gate specifications are reported
  as hidden width/depth. The HiRefPOU-2L entry lists coarse and fine experts,
  respectively. Parameter counts include all trainable network, embedding,
  expert, and gating parameters.
  }}
  \label{tab:darcy_architectures}
  \renewcommand{\arraystretch}{1.15}
  \small

  \begin{tabular}{lccccc}
  \toprule
  \textbf{Model}
  &
  \makecell{\textbf{Main / expert}\\\textbf{width}}
  &
  \makecell{\textbf{Main / expert}\\\textbf{depth}}
  &
  \makecell{\textbf{Gate}\\\textbf{width/depth}}
  &
  \makecell{\textbf{POU experts}\\\textbf{by level}}
  &
  \makecell{\textbf{Trainable}\\\textbf{parameters}}
  \\
  \midrule

  APINN
  & 32
  & 4
  & $32/3$
  & 2
  & 11,940
  \\

  PirateNet
  & 32
  & 3
  & --
  & --
  & 11,684
  \\

  HiRefPOU-1L
  & 32
  & 3
  & $16/1$
  & 2
  & 11,935
  \\

  HiRefPOU-2L
  & 32
  & 3
  & $16/1$
  & $2+3$
  & 22,162
  \\

  \bottomrule
  \end{tabular}

  \vspace{1mm}
  \begin{minipage}{0.96\linewidth}
  \footnotesize
  All models use the $\tanh$ activation. PirateNet uses a trainable
  32-dimensional Fourier embedding. APINN and HiRefPOU-1L use two
  POU components. HiRefPOU-2L uses two coarse experts and three fine
  experts, with parent--child allocation $(1,2)$.
  \end{minipage}
  \end{table}

\paragraph{Problem formulation and interface conditions}

We consider the steady Darcy problem
\begin{align}
  -\nabla\!\cdot\!\big(\mu(x)\,\nabla \phi(x)\big) &= f(x),
  \qquad x \in \Omega=[0,1]^2,
\end{align}
with permeability field $\mu(x)$ that is piecewise constant across subdomains.
The flux is defined as $\mathbf{q} = -\mu\nabla\phi$.

At material interfaces $\Gamma_{ij}$ between subdomains $\Omega_i$ and $\Omega_j$, the governing equations impose the physical continuity conditions
\begin{align}
  \phi_i(x) &= \phi_j(x), \\
  \mathbf{n}\!\cdot\!\mu_i\nabla\phi_i(x)
  &= \mathbf{n}\!\cdot\!\mu_j\nabla\phi_j(x),
  \qquad x\in\Gamma_{ij},
\end{align}
while tangential flux components may be discontinuous.

\paragraph{Five-strip manufactured solution}

The domain is partitioned into five horizontal layers
\[
  \Omega_i = \{(x,y)\in[0,1]^2 : 0.2(i-1)\le y \le 0.2i\},
  \quad i=1,\dots,5,
\]
with permeability values
\[
\mu|_{\Omega_1}=16,\;
\mu|_{\Omega_2}=6,\;
\mu|_{\Omega_3}=1,\;
\mu|_{\Omega_4}=10,\;
\mu|_{\Omega_5}=2.
\]
The manufactured solution $\phi_{\mathrm{ex}}(x,y)=1-x$ yields exact satisfaction of the interface conditions.

\paragraph{Limitations of globally smooth PINNs}

A central difficulty in applying standard PINNs to interface problems arises from the use of automatic differentiation over globally smooth neural networks.
When $\mu(x)$ is discontinuous, the product $\mu(x)\nabla\phi_\theta(x)$ is not differentiable across interfaces.
As a result, the PDE residual computed via automatic differentiation is inconsistent near interfaces, leading to spurious residual terms and degraded accuracy~\citep{yao2023Deep}.
Because the neural approximation $\phi_\theta(x)$ is globally smooth, it cannot represent solutions whose gradients are discontinuous, which explains the oscillatory behavior and reduced flux accuracy observed for standard PINN formulations.

\paragraph{Furthest Point Sampling for Center Point Initialization.}
\done{
We initialize the centers and associated radii using the farthest point sampling (FPS) and greedy covering strategy, a technique often used in meshfree approximation and RBF-based partitioning.
Given a set of data points $X = \{x_1, \dots, x_N\}$ and a desired number of centers $N_c$, the algorithm proceeds as follows.
Starting from a randomly chosen initial point (usefully one on the domain boundary) $c_1 \in X$, each subsequent center $c_i$ is selected as the data point in $X$ that maximizes its minimum distance to all previously chosen centers $\{c_1, \dots, c_{i-1}\}$. 
This greedy criterion ensures that each new center lies in the least represented region of the domain, resulting in a near-uniform covering with respect to the underlying data geometry.
}

\roneedits{
An appealing property of the FPS initialization is that it produces an ordered
set of centers that progressively improves geometric coverage of the domain.
Early-selected centers occupy widely separated regions, while later-selected
centers fill regions that remain far from the previously selected centers.
In the present framework, this ordering provides a convenient geometry-aware
initialization for the POU centers and should not be interpreted as imposing a
functional coarse-to-fine decomposition among the corresponding expert outputs.
The hierarchy of the HiRefPOU model is instead defined explicitly through the
nested parent--child POU weights described in Section~\ref{sec:methods}.
After initialization, the center locations and associated gating parameters are
optimized jointly with the remaining model parameters.
Thus, FPS provides well-distributed initial spatial anchors without imposing
frequency separation, stage-wise residual correction, or a multiresolution
decomposition of the learned representation.
Finally, this initialization strategy is domain-agnostic and
dimension-independent, since it only requires access to the input point cloud
$X$.
}

\FloatBarrier
\clearpage

\section{Additional Burgers Results}
\label{app:burgers_additional_results}

\noindent
This section reports additional ablations for the standard 1D Burgers benchmark, focusing on the gate parameterization and the sensitivity of the 1L HiRefPOU-DeepONet model to the number of adaptive partitions.

\begin{table}[!htpb]
  \centering
  \caption{\small \textit{
  Standard 1D Burgers benchmark: gate ablation for the HiRefPOU-1L DeepONet with $M=3$ adaptive partitions.
  We compare RBF, box, and hybrid gate formulations.
  Reported values are mean $\pm$ standard deviation over random seeds.
  Epoch time denotes the average wall-clock time per training epoch.
  }}
  \resizebox{0.9\textwidth}{!}{%
  \begin{tabular}{lccccc}
  \toprule
  \textbf{Gate} & \textbf{Test rel. $\ell_2$} & \textbf{Std.} & \textbf{Test rel. $\ell_\infty$} & \textbf{Std.} & \textbf{Epoch time} \\
  \midrule
  RBF gate    & $0.177134$ & $0.004654$ & $0.457744$ & $0.009059$ & $0.000542$ \\
  Box gate    & $0.176758$ & $0.004868$ & $0.457380$ & $0.007711$ & $0.000606$ \\
  Hybrid gate & $\mathbf{0.172893}$ & $0.001394$ & $\mathbf{0.454045}$ & $0.005357$ & $0.000629$ \\
  \bottomrule
  \end{tabular}
  }
  \label{tab:burgers_standard_1l_gate_ablation}
\end{table}

\done{
\paragraph{Gate ablation.}
Table~\ref{tab:burgers_standard_1l_gate_ablation} compares three gate parameterizations for the HiRefPOU-1L DeepONet on the standard 1D Burgers benchmark with $M=3$ adaptive partitions.
Among the three variants, the hybrid gate achieves the lowest mean test relative $\ell_2$ and $\ell_\infty$ errors, while also exhibiting the smallest standard deviation across seeds.
The box and RBF gates perform similarly to one another, but both are slightly less accurate and less stable than the hybrid formulation.
Although the hybrid gate incurs a modest increase in per-epoch cost, the overall trend suggests that combining center-based and box-based structure yields a more robust gate than either component alone on this benchmark.
}

\begin{table}[!htpb]
  \centering
  \caption{\small \textit{
  Standard 1D Burgers benchmark: sensitivity of the HiRefPOU-1L DeepONet to the number of adaptive partitions $M$ when using the hybrid gate.
  Reported values are mean $\pm$ standard deviation over random seeds.
  Epoch time denotes the average wall-clock time per training epoch.
  }}
  \resizebox{0.9\textwidth}{!}{%
  \begin{tabular}{lccccc}
  \toprule
  \textbf{Configuration} & \textbf{Test rel. $\ell_2$} & \textbf{Std.} & \textbf{Test rel. $\ell_\infty$} & \textbf{Std.} & \textbf{Epoch time} \\
  \midrule
  Hybrid gate, $M=2$ & $\mathbf{0.172622}$ & $0.002183$ & $\mathbf{0.452593}$ & $0.002469$ & $0.000547$ \\
  Hybrid gate, $M=3$ & $0.176576$ & $0.001287$ & $0.459151$ & $0.006000$ & $0.000629$ \\
  Hybrid gate, $M=4$ & $0.174892$ & $0.002286$ & $0.458236$ & $0.001768$ & $0.000692$ \\
  Hybrid gate, $M=6$ & $0.173398$ & $0.002347$ & $0.453612$ & $0.006004$ & $0.000833$ \\
  \bottomrule
  \end{tabular}
  }
  \label{tab:burgers_standard_1l_sensitivity}
\end{table}

\done{
\paragraph{Partition-count sensitivity.}
Table~\ref{tab:burgers_standard_1l_sensitivity} studies the sensitivity of the HiRefPOU-1L DeepONet to the number of adaptive partitions when using the hybrid gate.
The results do not show a monotone improvement as $M$ increases.
Instead, relatively small partition counts already perform well, with $M=2$ giving the best mean test relative $\ell_2$ and $\ell_\infty$ errors in this study, and $M=6$ achieving comparable accuracy at higher computational cost.
This suggests that for the standard 1D Burgers benchmark, the main benefit comes from introducing adaptive localization rather than from aggressively increasing the number of partitions.}

\FloatBarrier
\clearpage
\section{Additional Cavity Ablations}
\label{app:cavity_additional_ablations}

\noindent
This section reports additional ablations for the cavity benchmark, including the sensitivity of the 1L model to the partition count and residual strength, the effect of gate parameterization, and the effect of child allocation in the 2L hierarchy.

\begin{table}[!htpb]
\centering
\caption{\small \textit{
Cavity benchmark: partition-count and residual-strength ablation for the HiRefPOU-1L model.
The parameter $\alpha$ denotes the residual scaling used in the POU correction term.
Reported values are mean $\pm$ standard deviation over random seeds.
}}
\small
\begin{adjustbox}{width=0.7\linewidth}
\begin{tabular}{lllll}
\toprule
\textbf{Gate} & \textbf{$M$} & \textbf{$\alpha$} & \textbf{Final test rel. $\ell_2$} & \textbf{Std.} \\
\midrule
hybrid & 6 & 1    & $\num{1.298e-03}$ & $\num{1.858e-04}$ \\
hybrid & 6 & 1.25 & $\num{1.385e-03}$ & $\num{8.321e-05}$ \\
hybrid & 6 & 0.75 & $\num{1.418e-03}$ & $\num{1.589e-04}$ \\
hybrid & 2 & 1.25 & $\num{1.996e-03}$ & $\num{3.662e-04}$ \\
hybrid & 2 & 1    & $\num{1.868e-03}$ & $\num{7.134e-05}$ \\
hybrid & 2 & 0.75 & $\num{1.984e-03}$ & $\num{1.467e-04}$ \\
\bottomrule
\end{tabular}
\end{adjustbox}
\label{tab:cavity_hirefpou_1l_alpha_ablation}
\end{table}

\done{
\paragraph{Partition-count and residual-strength ablation.}
Table~\ref{tab:cavity_hirefpou_1l_alpha_ablation} examines the effect of the partition count $M$ and the residual scaling parameter $\alpha$ for the HiRefPOU-1L model on the cavity benchmark.
Across both values of $M$, the choice $\alpha=1$ gives the best mean test relative $\ell_2$ error, while either increasing or decreasing the residual strength leads to a mild degradation in performance.
The impact of the partition count is more pronounced: increasing from $M=2$ to $M=6$ substantially improves accuracy for all three $\alpha$ values.
These results suggest that the cavity problem benefits more strongly from increased spatial specialization than the Burgers benchmark, while also indicating that the unscaled residual contribution ($\alpha=1$) provides the most reliable 1L setting among the values tested here.
}

\begin{table}[!htpb]
\centering
\caption{\small \textit{
Cavity benchmark: gate-type ablation for the HiRefPOU-1L model.
Reported values are mean $\pm$ standard deviation over random seeds.
}}
\small
\begin{adjustbox}{width=0.52\linewidth}
\begin{tabular}{lll}
\toprule
\textbf{Gate} & \textbf{Final test rel. $\ell_2$} & \textbf{Std.} \\
\midrule
box    & $\num{1.559e-03}$ & $\num{3.561e-04}$ \\
hybrid & $\num{1.536e-03}$ & $\num{1.588e-04}$ \\
rbf    & $\num{1.821e-03}$ & $\num{2.109e-04}$ \\
\bottomrule
\end{tabular}
\end{adjustbox}
\label{tab:cavity_hirefpou_1l_gate_ablation}
\end{table}

\done{
\paragraph{Gate-type ablation.}
Table~\ref{tab:cavity_hirefpou_1l_gate_ablation} compares the box, hybrid, and RBF gate parameterizations for the HiRefPOU-1L model on the cavity benchmark.
The hybrid gate achieves the best mean test relative $\ell_2$ error, though its advantage over the box gate is modest.
In contrast, the RBF gate performs noticeably worse than both the box and hybrid variants.
Taken together with the Burgers results, this supports the use of the hybrid gate as a consistent default choice: it is never worse in these experiments and is generally either the best or tied for best, while also offering lower variability than the purely RBF formulation.
}

\begin{table}[!htpb]
\centering
\caption{\small \textit{
Cavity benchmark: child-allocation ablation for the HiRefPOU-2L model.
Here, $M_c$ and $M_f$ denote the numbers of coarse and fine experts, respectively, while $\alpha_c$ and $\alpha_f$ denote the corresponding residual strengths.
The allocation label indicates how fine-level children are distributed across the coarse-level parents.
Reported values are mean $\pm$ standard deviation over random seeds.
}}
\small
\begin{adjustbox}{width=0.80\linewidth}
\begin{tabular}{lllllll}
\toprule
\textbf{Allocation} & \textbf{$M_c$} & \textbf{$M_f$} & \textbf{$\alpha_c$} & \textbf{$\alpha_f$} & \textbf{Final test rel. $\ell_2$} & \textbf{Std.} \\
\midrule
0\_2 & 2 & 2 & 1 & 1.25 & $\num{1.955e-03}$ & $\num{1.827e-04}$ \\
0\_2 & 2 & 2 & 1 & 0.75 & $\num{1.920e-03}$ & $\num{1.812e-04}$ \\
2\_0 & 2 & 2 & 1 & 1    & $\num{2.013e-03}$ & $\num{3.089e-04}$ \\
2\_2 & 2 & 4 & 1 & 1    & $\num{1.944e-03}$ & $\num{1.421e-04}$ \\
2\_0 & 2 & 2 & 1 & 0.75 & $\num{2.012e-03}$ & $\num{1.736e-04}$ \\
2\_0 & 2 & 2 & 1 & 1.25 & $\num{2.123e-03}$ & $\num{2.716e-04}$ \\
0\_2 & 2 & 2 & 1 & 1    & $\num{1.934e-03}$ & $\num{1.780e-04}$ \\
2\_2 & 2 & 4 & 1 & 1.25 & $\num{1.966e-03}$ & $\num{1.728e-04}$ \\
2\_2 & 2 & 4 & 1 & 0.75 & $\num{1.971e-03}$ & $\num{1.695e-04}$ \\
\bottomrule
\end{tabular}
\end{adjustbox}
\label{tab:cavity_hirefpou_2l_child_allocation_ablation}
\end{table}

\paragraph{Child-allocation ablation.}
Table~\ref{tab:cavity_hirefpou_2l_child_allocation_ablation} studies how the distribution of fine-level children across the coarse parents affects performance in the HiRefPOU-2L cavity model.
The results show that performance is sensitive to how refinement capacity is allocated.
Among the tested configurations, the \texttt{0\_2} allocation yields the best results, while the \texttt{2\_0} allocation is consistently the weakest.
The balanced \texttt{2\_2} configuration performs competitively, but does not outperform the best asymmetric allocation in this experiment.
This indicates that, for the cavity benchmark, refinement is most effective when it is concentrated in the parent region associated with the more challenging local structure rather than distributed uniformly by default.

\vskip 0.2in
\bibliographystyle{tmlr}
\bibliography{references}

\end{document}